\documentclass[journal,10pt,onecolumn,draftclsnofoot,]{IEEEtran}

\usepackage{hyperref}
\usepackage{url}
\usepackage{cite}
\usepackage{amsthm}
\newtheoremstyle{mystyle}
  {}
  {}
  {\itshape}
  {}
  {\bfseries}
  {.}
  { }
  {\thmname{#1}\thmnumber{ #2}\thmnote{ (#3)}}

\theoremstyle{mystyle}

\newtheorem*{lemma*}{Lemma}
\newtheorem{corollary}{Corollary}
\newtheorem{lemma}{Lemma}
\newtheorem{definition}{Definition}
\newtheorem{prop}{Proposition}
\usepackage{times,amsmath,epsfig}
\usepackage{epstopdf}
\usepackage{amssymb}
\usepackage{float}
\usepackage[framemethod=1]{mdframed}
\usepackage{xcolor}
\usepackage{scalerel}
\usepackage{bbding}
\usepackage{mathtools}
\usepackage{graphicx}
\usepackage{url}
\usepackage{multirow}
\usepackage{multicol}
\usepackage{subfigure}
\usepackage{threeparttable}
\usepackage{booktabs}
\usepackage{amsfonts}
\usepackage{textcomp}
\usepackage{enumitem}
\usepackage[ruled,vlined]{algorithm2e}
\usepackage{color}
\usepackage{algorithmic}
 
\usepackage[hang,flushmargin]{footmisc}
\usepackage{lipsum}
\makeatletter
\newcommand{\algorithmfootnote}[2][\footnotesize]{%
  \let\old@algocf@finish\@algocf@finish
  \def\@algocf@finish{\old@algocf@finish
    \leavevmode\rlap{\begin{minipage}{\linewidth}
    #1#2
    \end{minipage}}%
  }%
}
\ifodd 0
    \newcommand\rev[1]{{\color{blue}#1}}
    \newcommand{\com}[1]{\textbf{\color{red} (COMMENT: #1)}} 
\else
    \newcommand\rev[1]{{#1}}
    \newcommand{\com}[1]{}
\fi

\newcommand{\tabincell}[2]{\begin{tabular}{@{}#1@{}}#2\end{tabular}}

\begin{document}
\title{Ensuring DNN Solution Feasibility for Optimization Problems with Convex Constraints and Its Application to DC Optimal Power Flow Problems}


 \author{Tianyu Zhao, Xiang Pan, Minghua Chen, and Steven H. Low \thanks{Tianyu Zhao and Xiang Pan are with Information Engineering, The Chinese University of Hong Kong. Minghua Chen is with School of Data Science, City University of Hong Kong. Steven H. Low is with Department of Computing and Mathematical Sciences, California Institute of Technology.}}

%



\maketitle
\IEEEpeerreviewmaketitle

\begin{abstract}

Ensuring solution feasibility is a key challenge in developing Deep Neural Network (DNN) schemes for solving constrained optimization problems, due to inherent DNN prediction errors. In this paper, we propose a ``preventive learning'' framework to guarantee DNN solution feasibility for problems with convex constraints and general objective functions without post-processing, upon satisfying a mild condition on constraint calibration. 
{Without loss of generality, we focus on problems with only inequality constraints.} We systematically calibrate inequality constraints used in DNN training, thereby anticipating prediction errors and ensuring the resulting solutions remain feasible. We characterize the calibration magnitudes and the DNN size sufficient for ensuring universal feasibility. We propose a new \textit{Adversarial-Sample Aware} training algorithm to improve DNN's optimality performance without sacrificing feasibility guarantee. Overall, the framework provides two DNNs. The first one from characterizing the sufficient DNN size can guarantee universal feasibility while the other from the proposed training algorithm further improves optimality and maintains DNN's universal feasibility simultaneously. We apply the framework to develop \textsf{DeepOPF+} for solving essential DC optimal power flow problems in grid operation. Simulation results over IEEE test cases show that it outperforms {existing strong DNN baselines} in ensuring 100\% feasibility and attaining consistent optimality loss ($<$0.19\%) and speedup (up to $\times$228) in both light-load and heavy-load regimes, as compared to a state-of-the-art solver. We also apply our framework to a non-convex problem and show its performance advantage over existing schemes.

\end{abstract}

\section{Introduction}\label{sec:intro}
Recently, there have been increasing interests in employing neural networks, including deep neural networks (DNN), to solve constrained optimization problems in various problem domains, especially those needed to be solved repeatedly in real-time. The idea behind these machine learning approaches is to leverage the universal approximation capability of DNNs~\cite{hornik1989multilayer, leshno1993multilayer, hanin2019universal} to learn the mapping between the input parameters to the solution of an optimization problem. Then one can directly pass the input parameters through the trained neural network to obtain a quality solution much faster than iterative solvers. For example, researchers have developed DNN schemes to solve the essential optimal power flow problems in grid operation with sub-percentage optimality loss and several order of magnitude speedup as compared to conventional solvers, for power networks with more than two thousand buses~\cite{deepopf1,deepopf2,pan2020deepopf,donti2021dc3,chatzos2020high,lei2020data,huang2021deepopf}. Similarly, DNN-based schemes also obtain desirable results for real-time power control and beam-forming designs~\cite{sun2018learning,xia2019deep} problems in wireless communication systems in a fraction of the time used by existing solvers. 

Despite these promising results, however, a major criticism of DNN and machine learning schemes is that they usually cannot guarantee the solution feasibility with respect to all the inequality and equality constraints of the optimization problem~\cite{deepopf2}. This is due to the inherent prediction errors of neural network models. Failing to respect the system physical and operational constraints can be fatal and lead to system instability or incur higher operating cost for the system operator~\cite{zhao2020deepopf}. Existing works address the feasibility concern mainly by incorporating the constraints violation (e.g., a Lagrangian relaxation to compute constraint violation with Lagrangian multipliers) into the loss function to guide the DNN training. These endeavors, while generating great insights to the DNN design and working to some extent in case studies, can not guarantee the solution feasibility without resorting to expensive post-processing procedures, e.g., feeding the DNN solution as a warm start point into an iterative solver to obtain a feasible solution. {See Sec.~\ref{sec:related.work} for more discussions.} To date, it remains a largely open issue of ensuring DNN solution \rev{(output of DNN)} feasibility for constrained optimization problems.

In this paper, we address this issue for Optimization Problems with Convex (Inequality) Constraints (OPCC) and general objective functions \rev{with varying problem inputs and fixed objective/constraints parameters}. Since linear equality constraints can be exploited  to reduce the number of decision variables without losing optimality (and removed), it suffices to focus on problems with inequality constraints. Our idea is to train DNN in a preventive manner to ensure the resulting solutions remain feasible even with prediction errors, thus avoiding the need of post-processing. We make the following contributions: We make the following contributions:

\begin{itemize}

    \item  After formulating the OPCC problem in Sec.~\ref{sec:OPCC_formulation}, we propose a ``preventive learning'' framework to ensure the DNN solution feasibility  for OPCC in Sec.~\ref{sec:preventive.learning.framework}. We first remove the non-critical inequality constraints without loss of generality. We then exploit (and remove) the linear equality constraints and reduce the number of decision
variables without losing optimality by adopting the predict-and-reconstruct design~\cite{deepopf2}. Then we systematically calibrate inequality constraints used in DNN training, thereby anticipating prediction errors and ensuring the resulting DNN solutions \rev{(outputs of the DNN)} remain feasible.

    \item  Then {in Sec.~\ref{ssec:calibrationrange}, we characterize the allowed calibration rate necessary for ensuring universal feasibility with respect to the entire parameter input region by solving a bi-level problem with a heuristic method, i.e., the rate of adjusting (reducing) constraints limits that represents the room for (prediction) errors without violating constraints.} We then derive the sufficient DNN size for ensuring DNN solution feasibility in Sec.~\ref{ssec:feasibilityDNN}, by adapting an integer linear formulation of DNN from~\cite{venzke2020learning,tjeng2018evaluating}. Note that a universal feasibility guaranteed DNN can be directly constructed without training. 
    \item Observing the feasibility-guaranteed DNN may not achieve strong optimality performance, in Sec.~\ref{ssec:ASAA}, based on the ideas of active learning and adversarial training, we propose a new \textit{Adversarial-Sample Aware} training algorithm to improve DNN's optimality performance without sacrificing feasibility guarantee. {Overall, the framework provides two DNNs. The first one constructed from the step of determining the sufficient DNN size can guarantee universal feasibility, while the other DNN obtained from the proposed \textit{Adversarial-Sample Aware} training algorithm further improves optimality and maintains DNN's universal feasibility simultaneously.}

    \item In Sec.~\ref{sec:simulations}, we apply the framework to design a DNN scheme, \textsf{DeepOPF+}, for solving DC optimal power flow (DC-OPF) problems in grid operation. It improves over existing DNN schemes in ensuring feasibility and attaining consistent desirable speedup performance in both light-load and heavy-load regimes. {Note that under the heavy-load regime, the system constraints are highly binding and the existing DNN schemes may not achieve high speedups due to the need of an expensive post-processing procedure to recover feasibility of infeasible DNN solutions.} Simulation results over IEEE 30/118/300-bus test cases show that \textsf{DeepOPF+} outperforms existing DNN schemes in ensuring $100\%$ feasibility and attaining consistent optimality loss ($<$0.19\%) and computational speedup (up to two orders of magnitude $\times$228) in both light-load and heavy-load regimes, as compared to a state-of-the-art iteration-based solver.
\end{itemize}

\section{Related Work}\label{sec:related.work}
There have been active studies in employing machine learning models, including DNNs, \rev{to solve constrained optimizations directly~\cite{KotaryFHW21,deepopf1,deepopf2,zhou2022deepopf,guha2019machine,zamzam2020learning,fioretto2020predicting,dobbe2019towards,sanseverino2016multi,elmachtoub2022smart,huang2021deepopf,deepopfngt}, obtaining close-to-optimal solution much faster than conventional iterative solvers.} For brevity, we focus on applying learning-based methods to solve constrained optimization problems, divided into two categories.

The first category is the hybrid approach. It integrates learning techniques to facilitate conventional algorithms solving challenging constrained optimization problems~\cite{gutierrez2010neural,vaccaro2016knowledge,halilbavsic2018data,biagioni2020learning,pineda2020data,jamei2019meta,deka2019learning,karagiannopoulos2019data,baker2019joint,ng2018statistical}. For example, some works use DNN to identify the active/inactive constraints of LP/QP to reduce problem size~\cite{misra2018learning,zhai2010fast,roald2019implied,9091534,chen2020learning} or predict warm-start initial points or gradients to accelerate the solving process~\cite{baker2019learning,dong2020smart} and speed up the branch-and-bound algorithm~\cite{balcan2018learning,he2014learning}. Nevertheless, the core of these methods is still conventional solver that may incur high computational costs for large-scale programs due to the inevitable iteration process.

The second category is the stand-alone approach, which leverages machine earning models to predict constrained optimization problems {solutions} without resorting to the conventional solver~\cite{KotaryFHW21,deepopf1,deepopf2,guha2019machine,zamzam2020learning,fioretto2020predicting,dobbe2019towards,sanseverino2016multi}. For example, 
existing works belong to the ``learn to optimize'' field, using RNN to mimic the gradient descent-wise iteration and achieve faster convergence speed empirically~\cite{li2016learning,chen2017learning}. Other works like~\cite{donti2021dc3,pan2020deepopf,zhao2020deepopf} directly used the DNN model to predict the final solution (regarded as end-to-end method), which can further reduce the computing time compared to the iteration-based approaches.
These approaches, in general, can have better speedup performance compared with the hybrid approaches.

Though end-to-end methods have been actively studied for constrained optimizations with promising speedups, the lack of feasibility guarantees presents a fundamental barrier for practical application, e.g., infeasibility due to inaccurate active/inactive limits identification. Infeasible solutions from the end-to-end approach are also observed~\cite{zhao2020deepopf,deepopf2}, especially considering the DNN worst-case performance under Adversarial input {with serious constraints violations~\cite{venzke2020learning,nellikkath2021physics,nellikkath2021physics2}.} 
This echoes the critical challenge of ensuring the DNN solutions feasibility w.r.t. constraints due to inherent prediction errors.

Some efforts have been put to improve DNN feasibility, e.g., considering solution generalization~\cite{zhang2020convex} or appealing to post-processing schemes~\cite{deepopf2}.
Some existing works tackle the feasibility concern by incorporating the constraints violation in DNN training~\cite{pan2020deepopf,donti2021dc3}. In~\cite{nellikkath2021physics,nellikkath2021physics2}, physics-informed neural networks are applied to predict solutions while incorporating the KKT conditions of optimizations during training.  Though the PINNs present better worst-case performance, the constraints satisfaction is not guaranteed by the obtained predicted solution. These approaches, while attaining insightful performance in case studies, {do not provide solution feasibility guarantee and may resort to expensive projection procedure~\cite{deepopf2}. There is an emerging line of works focusing on developing structured neural network layers that specify the implicit relationships between inputs and outputs~\cite{wang2019satnet,chen2018neural,ling2018game,de2018end,bai2019deep,donti2017task,djolonga2017differentiable,tschiatschek2018differentiable,wilder2019melding,gould2019deep}.  Such approaches can directly enforce constraints, e.g., by projecting neural network outputs onto the feasible region described by linear constraints using quadratic programming layers~\cite{amos2017optnet}, or convex optimization layers~\cite{agrawal2019differentiable} for general convex constraints. While the projection based post-processing step can retrieve a feasible solution in the face of infeasibility, the scheme turns to be computationally expensive and inefficient. A gradient-based violation correction is proposed in~\cite{donti2021dc3}. Though a feasible solution can be recovered for linear constraints, it can be computationally inefficient and may not converge for general optimizations. A DNN scheme applying gauge function that maps a point in an $l_1$-norm unit ball to the (sub)-optimal solution is proposed in~\cite{li2022learning}. However, its feasibility enforcement is achieved from a computationally expensive interior-point finder program.} There is also a line of work~\cite{ferrari2009multiobjective,yin2022learning,qin2019verification,limanond1998neural} focusing on verifying whether the output of a given DNN satisfies a set of
requirements/constraints. However, these approaches are only used for evaluation and not capable of obtaining a DNN with feasibility-guarantee and strong optimality. To our best knowledge, this work is the \textit{first} to guarantee DNN solution feasibility without post-processing.

In addition to constructing new DNN layers, several techniques that try to repair the wrong behaviors of DNN by adjusting the DNN weights are proposed~\cite{sotoudeh2021provable,yang2021neural,sohn2019search}. However, such modifications may lead to unanticipated performance degradation of DNNs due to the lack of performance guarantee. In~\cite{sotoudeh2021provable}, a decoupled DNN architecture is introduced. The idea is to decouple the activations of the DNN from values of the DNN by augmenting the original neural network. With such construction, a LP based approach is proposed for single-layer weight repair. However, the considered feasible region of the DNN output are are fixed polytopes and hence can not handle the interested problems with input-varying output feasible regions. In addition, since only a single layer repair is considered, there is no guarantee to always find a practicable adjustment and hence fails the approach.

Our work also fundamentally relates to the field of DNN robustness. {Several methods have been proposed to verify DNN robustness against input adversarial perturbations unconstrained for classification tasks~\cite{sheikholeslami2020provably,dvijotham2018dual,wong2018provable,li2020sok}.} These approaches generally depends on the network relaxation or the Lipschitz bound of DNN with accuracy as the metric. Our work differs significantly from ~\cite{zhao2020deepopf} in that we can provably guarantee DNN solution feasibility for optimization with convex/linear constraints and develop a new learning algorithm to improve solution optimality, including determining both the inequality constraint calibration rate and DNN size necessary for ensuring universal feasibility and deriving the active training scheme considering both optimality and feasibility. 

To our best knowledge, our work is the first to provide systematical understanding whether it is possible to achieve DNN solution's universal feasibility for all the inputs within an interested region, and if so, how to design and train a DNN to achieve decent optimality performance while ensuring solution feasibility. 



\section{Optimization Problems with Convex Constraints}\label{sec:OPCC_formulation}

We focus on the OPCC formulated as follows~\cite{boyd2004convex,faisca2007multiparametric}:
\begin{align}
    \min_{\boldsymbol{x}\in \mathcal{R}^N} \ &f(\boldsymbol{x},\boldsymbol{\theta})\label{equ:OPCC.formulation-1} \\
     \mathrm{s.t.} \quad &g_j(\boldsymbol{x,\theta})\leq e_j,\ j\in\mathcal{E}.\label{equ:OPCC.formulation-3}
     \\
     &\underline{x}_k\leq x_k\leq \bar{x}_k,\ k=1,...,N.\label{equ:OPCC.formulation-4}
\end{align}
In the formulation, $\boldsymbol{x}\in\mathcal{R}^N$ are the decision variables, $\mathcal{E}$ is the set of inequality constraints, $\boldsymbol{\theta}\in\mathcal{D}$ are the input parameters. The objective function $f(\boldsymbol{x},\boldsymbol{\theta})$ is general and can be either convex or non-convex. 

We assume the input domain $\mathcal{D}=\{\boldsymbol{\theta}\in \mathcal{R}^M|\mathbf{A}_{\boldsymbol{\theta}}\boldsymbol{\theta}\leq \boldsymbol{b_{\theta}}\}$ is a convex polytope specified by matrix $\mathbf{A}_{\boldsymbol{\theta}}$ and vector $\boldsymbol{b_{\theta}}$ such that for each $\boldsymbol{\theta}\in\mathcal{D}$, the OPCC in \eqref{equ:OPCC.formulation-1}--\eqref{equ:OPCC.formulation-4} admits a unique optimal solution.\footnote{Here $\mathbf{A}_{\boldsymbol{\theta}}$ and $\boldsymbol{b_{\theta}}$ are constant matrix and vector and are not changing w.r.t. $\boldsymbol{\theta}$ and hence $\mathcal{D}$ is a constant polytope. \rev{Our approach is also applicable to non-unique solution and unbounded $\boldsymbol{x}$. See Appendix~\ref{appendix.D} for a discussion.}}. $g_j: \mathcal{R}^N\times\mathcal{R}^M\rightarrow \mathcal{R}, j\in\mathcal{E}$ are convex functions w.r.t. $\boldsymbol{x}$.
{We also model each $x_k$ to be restricted by an upper bound $\bar{x}_k$ and lower bound $\underline{x}_k$ (box constraints).
{Here we focus on the setting that all the inequality constraints $g_j$ are critical. Formally, the critical inequality constraint is defined as}
\begin{definition}\label{def:critical}
An inequality constraints $g_j(\boldsymbol{x},\boldsymbol{\theta})\leq e_j$ is critical if there exists a $\boldsymbol{\theta}\in\mathcal{D}$ and $\boldsymbol{x}$ satisfying \eqref{equ:OPCC.formulation-4} such that $g_j(\boldsymbol{x},\boldsymbol{\theta})\leq e_j$ is active.
\end{definition}

{Non-critical constraints are always respected for any combination of input $\boldsymbol{\theta}\in\mathcal{D}$ and $\boldsymbol{x}$ satisfying the box constraints \eqref{equ:OPCC.formulation-4}. Thus, removing them will not change the optimal solution of OPCC for any input parameter in the input domain. Without loss of generality, we assume that all the inequality constraints $g_j$ are critical. We refer to Appendix~\ref{appen:removing.noncritical} for the problem formulations with potential non-critical inequality constraints and a method to identify and remove these non-critical inequality constraints as well as the corresponding discussions.  We note that linear equality constraints can be exploited (and removed) to reduce the number of decision variables without losing optimality as discussed in Appendix~\ref{appendix.equality}, it suffices to focus on OPCC with inequality constraints as formulated in (\ref{equ:OPCC.formulation-1})-(\ref{equ:OPCC.formulation-4}).}

The OPCC in \eqref{equ:OPCC.formulation-1}--\eqref{equ:OPCC.formulation-4} has wide applications in various engineering domains, e.g., DC-OPF problems in power systems~\cite{deepopf1} and model-predictive control problems in control systems~\cite{bemporad2000explicit}. While many numerical solvers based on, e.g., those based on interior-point methods~\cite{ye1989extension}, can be applied to obtain its solution, the time complexity can be significant and limits their practical applications {especially considering the problem input uncertainty under various scenarios} As a concrete example, a critical problem in power system operation, the security-constrained DC-OPF (SC-DCOPF) problem incurs a complexity of $\mathcal{O} \left( K^{12}\right)$ to solve it optimally, where $K$ is number of buses, limiting its practicability.


The observation that opens the door for DNN scheme development lies in that solving OPCC is equivalent to learning the mapping between input $\boldsymbol{\theta}$ to the optimal solution $\boldsymbol{x}^*(\boldsymbol{\theta})$, which is continuous w.r.t. $\boldsymbol{\theta}$ if OPCC admits a unique optimal solution for every $\boldsymbol{\theta}\in\mathcal{D}$~\cite{pan2020deepopf,bemporad2006algorithm}. For multiparametric quadratic programs (mp-QP), i.e., $f$ is quadratic w.r.t. $\boldsymbol{x}$ and $g_i$ are linear functions,  $\boldsymbol{x}^*(\boldsymbol{\theta})$ can be further characterize to be piece-wise linear~\cite{faisca2007multiparametric}. 
As such, it is conceivable to leverage the universal approximation capability of deep feed-forward neural networks~\cite{hornik1989multilayer, leshno1993multilayer,goodfellow2016deepma}, to learn the input-solution mapping $\boldsymbol{x}^*(\boldsymbol{\theta})$ for a given OPCC formulation, and then apply the DNN to obtain optimal solutions for any $\boldsymbol{\theta}\in\mathcal{D}$ with significantly lower time complexity. {For example, DNN schemes have been proposed to solve the above-mentioned SC-DCOPF problems with a complexity as low as $\mathcal{O} \left( K^{5}\right)$ and minor optimality loss~\cite{deepopf2,pan2020deepopf}.} See Sec.~\ref{sec:related.work} for more discussions on developing DNN schemes for solving optimization problems.

While DNN schemes achieve promising speedup and optimality performance, a fundamental challenge lies in ensuring solution feasibility, which is nontrivial due to inherent DNN prediction errors. {For example, in the previous work~\cite{pan2020deepopf,deepopf2}, the obtained DNN solutions may violate the inequality constraints especially when the constraints are binding.} In the following, we propose a preventive learning framework to tackle this issue for designing DNN schemes to solve OPCC in~(\ref{equ:OPCC.formulation-1})-(\ref{equ:OPCC.formulation-4}).}

\section{Preventive Learning Framework for OPCC}\label{sec:preventive.learning.framework}
\subsection{Overview of the Framework}\label{ssec:overview}
\begin{figure}[!t]
	\centering
	\includegraphics[width = 1\textwidth]{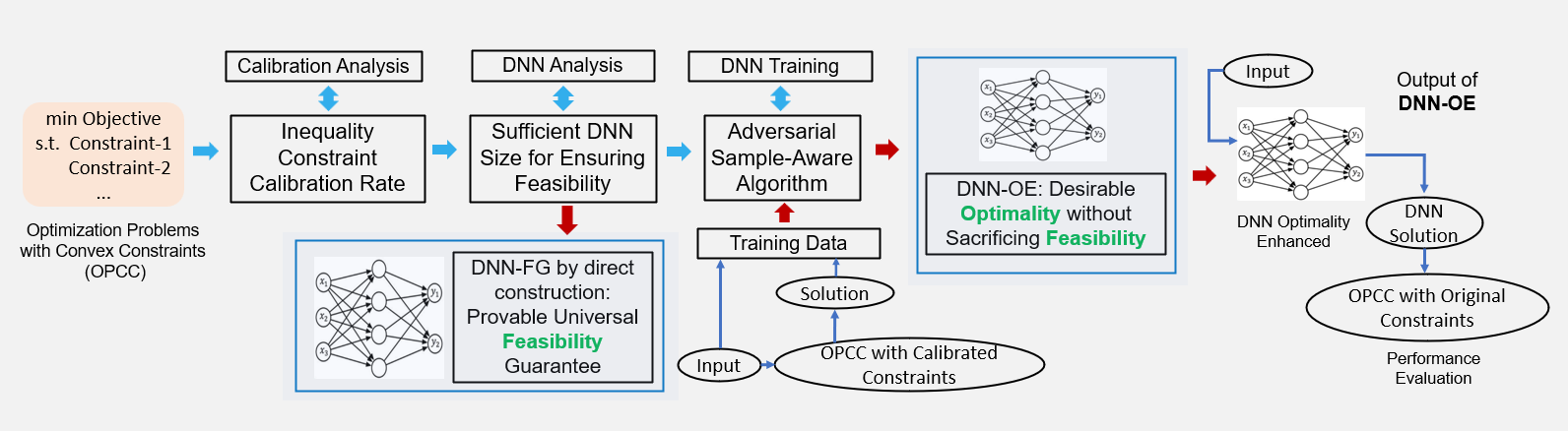}
	\caption{Overview of preventive learning framework for solving OPCC.
	The maximum calibration rate is first characterized to preserve the input domain. The sufficient DNN size in guaranteeing universal feasibility is then determined, and a DNN model can be constructed directly with universal feasibility guarantee in this step. With the determined calibration rate and sufficient DNN size, a DNN model with enhanced optimality without sacrificing feasibility is obtained using the \textit{Adversarial Sample-Aware} algorithm for performance evaluation.
	}
	\label{fig:framework}
\end{figure}

\begin{figure} [!t]
  \centering
    \includegraphics[width = 0.8\textwidth]{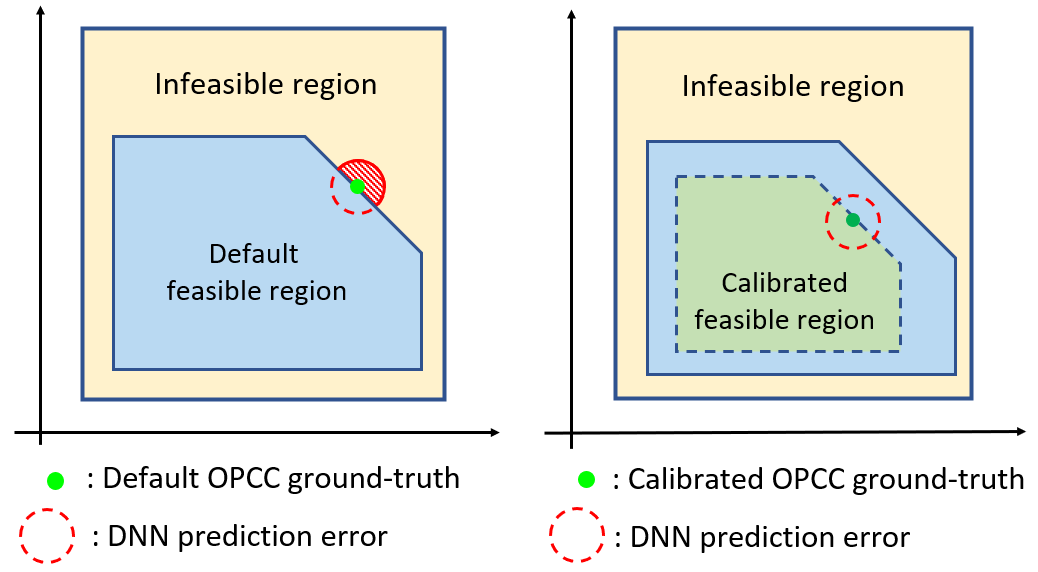}
  \caption{Left: the solution of DNN trained with default OPCC ground-truth can be infeasible due to inevitable prediction errors. \rev{Right: the solution of DNN trained with calibrated OPCC ground-truth can ensure universal feasibility even with prediction errors.}}
  \label{fig.illustration}
\end{figure}

We propose a preventive learning framework to develop DNN schemes for solving OPCC in \eqref{equ:OPCC.formulation-1}--\eqref{equ:OPCC.formulation-4} by learning input-solution mapping, as depicted in Fig.~\ref{fig:framework}. As a key component of the proposed framework, we \textit{calibrate} the inequality constraints used in DNN training such that for any interested input parameter, the trained DNN can provide a feasible and close-to-optimal solution even with the approximation error. See Fig.~\ref{fig.illustration} for illustrations. 
Then, we train the DNN on a (algorithmic designed) dataset created with calibrated limits to learn the corresponding input-solution mapping (${\Omega}: \boldsymbol{\theta} \mapsto \mathcal{S}$) and evaluate its performance on a test data-set with the original limits. Thus, even with the inherent prediction error of DNN, the obtained solution can still remain feasible.  We remark that during the training stage, the inequality limits calibration does not reduce the feasibility region of inputs $\boldsymbol{\theta}$ is in consideration. Also, we note that the constraints calibration could lead to the (sub)optimal solutions that are interior points within the original feasible region (the inequality constraints are expected to be not binding) when approximating the input-solution mapping for the OPCC with calibrated constraints. Thus, determine a proper calibration rate is important. As the approximation capability depends on the size of DNN, another critical problem is to design DNN with sufficient size for ensuring universal feasibility on the entire parameter input region. In the following subsections, we discuss how to address these problems with a proposed systematic scheme, which consists of three steps:\footnote{We note that the proposed \textit{preventive leaning} framework is also applicable to non-linear inequality constraints, e.g., AC-OPF problems with several thousand buses, but with additional computational challenge in solving the related programs corresponding to the required steps. We leave the application to optimization problem with non-linear constraints for future study.}
\begin{itemize}
    \item First, in Sec.~\ref{ssec:calibrationrange}, we determine the maximum calibration rate for inequality constraints, so that solutions from a preventively-trained DNN using the calibrated constraints respect the original constraints for all possible inputs. \rev{Here we refer the output of the DNN as the DNN solution.}
     
    \item Second, in Sec.~\ref{ssec:feasibilityDNN}, we determine a sufficient DNN size so that with preventive learning there exists a DNN whose worst-case violation on calibrated constraints is smaller than the maximum calibration rate, thus ensuring DNN solution feasibility, \rev{i.e., DNN's output always satisfies~(\ref{equ:OPCC.formulation-3})-(\ref{equ:OPCC.formulation-4}) for any input.} We construct a {provable feasibility-guaranteed DNN model, namely DNN-FG,} as shown in Fig.~\ref{fig:framework}.
    
    \item Third, observing DNN-FG may not achieve strong optimality performance, in Sec.~\ref{ssec:ASAA}, we propose an adversarial \textit{Adversarial Sample-Aware} training algorithm. It aims to further improve DNN's optimality performance without sacrificing feasibility guarantee, resulting in a optimality-enhanced DNN as shown in Fig.~\ref{fig:framework}.
\end{itemize}
{Overall, the framework provides two DNNs. The first one constructed from the step of determining the sufficient DNN size can guarantee universal feasibility (DNN-FG), while the other DNN obtained from the proposed \textit{Adversarial-Sample Aware} training algorithm further improves optimality without sacrificing DNN's universal feasibility (DNN Optimality Enhanced).}
To better deliver the results in the framework, we briefly summarize the relationship between the settings and the applied methodologies in Table~\ref{tablesummary2}. We further discuss the results that can be obtained in polynomial time by solving the proposed programs. 

\begin{table}[H]
	\centering
	\fontsize{9}{12}\selectfont
	\caption{Methodologies under different settings.}\label{tablesummary2}
	\begin{threeparttable}
		\scalebox{1}{
	\begin{tabular}{c|c|c|c}
	\toprule
		\hline
		{\tabincell{c}{Problem setting}}& 
		{\tabincell{c}{Determine calibration \\rate}}&
		{\tabincell{c}{Determine DNN size for\\ ensuring universal feasibility}}&
		{\tabincell{c}{ASA training algorithm}} \cr
		\hline
		\multirow{1}{*}{\tabincell{c}{General OPCC}}
		& \tabincell{c}{$^*$Non-convex\\ optimization}&\tabincell{c}{$^*$Bi-level non-convex \\mixed-integer optimization}&\tabincell{c}{$^*$Non-convex \\mixed-integer optimization}\cr\hline
		{\tabincell{c}{OPLC}}
		& \tabincell{c}{$\hat{}$  Mixed-integer \\linear programming  }&\tabincell{c}{$\hat{}$  Bi-level mixed-integer \\linear programming}&\tabincell{c}{$\hat{}$ Mixed-integer \\linear programming}\cr\hline
		\bottomrule
	\end{tabular}}
	\begin{tablenotes}
			\footnotesize
            \item[*] {Symbol $^*$ represents that we can obtain a valid bound in polynomial time by solving the corresponding program, which is still be useful for analysis.}
            \item[*] {Symbol \ $\hat{}$ \ represents that we may not obtain a valid and useful bound in polynomial time by solving the corresponding program for further analysis.}
		\end{tablenotes}
	\end{threeparttable}
\end{table}

We remark that the involved problems for each step are indeed non-convex programs. The existing solvers, e.g., Gurobi, CPLEX, or APOPT, may not provide the global optimum. However, we may still be able to obtain the useful bounds from the solver under the specific setting. We briefly present the results in the following, in which the upper/lower bounds denote the (sub-optimal) objective values of the programs that
can be obtained in polynomial time, e.g., when the solvers terminate at any time but not returning an optimal solution.

For general OPCC:
\begin{itemize}
    \item Determine calibration rate: We can get a upper bound on the global optimum of the maximum calibration rate (with the feasible (sub-optimal) solution), which may not be valid and useful for further analysis. Such an upper bound may lead some input to be infeasible and hence universal feasibility may not be guaranteed.
    \item Determine DNN size for ensuring universal feasibility: We can get a lower bound (with the feasible (sub-optimal) solution) on the worst-case violation with the obtained DNN parameters, while we may not get the valid and useful result of the global optimum of the bi-level program for further analysis. Such determined DNN size may not guarantee universal feasibility with the obtained objective value under the specified DNN parameters.
    \item Adversarial sample-aware training algorithm: We can get a lower bound on the global optimum of the worst-case violation (with the feasible (sub-optimal) solution), which may not be valid for further analysis. Such a lower bound may not guarantee universal feasibility under the trained DNN.
\end{itemize}

For Optimization Problems with Linear Constraints (OPLC), i.e., $g_j(\boldsymbol{x},\boldsymbol{\theta})\triangleq \boldsymbol{a^T_j}\boldsymbol{x}+\boldsymbol{b^T_j}\boldsymbol{\theta}\leq e_j,\ j\in \mathcal{E}$ are all linear:
\begin{itemize}
    \item Determine calibration rate: We can get a \textit{valid lower bound} on the global optimum of the maximum calibration rate (may or may not not with the feasible solution). Such a lower bound ensures that we will not calibrate the constraints too much and hence preserves the input parameter region, which is still useful for further analysis in the following steps.
    \item Determine DNN size for ensuring universal feasibility: We can get a \textit{valid upper bound} on the worst-case violation with the obtained DNN parameters (may or may not not with the feasible solution) and the global optimum of the bi-level MILP program. If such a upper bound is no greater than the calibration rate, the determined DNN size is assured to be sufficient and universal feasibility of DNN is guaranteed.
    
    \item Adversarial sample-aware training algorithm: We can get a \textit{valid upper bound} on the global optimum of the worst-case violation (may or may not with the feasible solution), which is still useful for analysis. If such a upper bound is no greater than the calibration rate, the universal feasibility guarantee of the obtained DNN is assured.
\end{itemize}

We remark that the bounds obtained from the setting of OPLC can still be useful and valid for further analysis. Therefore, we can construct the DNNs with provable universal solution feasibility guarantee. However, the results under the general OPCC can be loose and may not be utilized with desired performance guarantee. In addition, we state that for each involved program, we can always obtain a feasible (sub-optimal) solution for further use. These results are discussed in the corresponding parts in the paper.

\subsection{{Inequality Constraint Calibration Rate} }\label{ssec:calibrationrange}

We calibrate each inequality limit $g_j(\boldsymbol{x},\boldsymbol{\theta})\leq e_j,$\footnote{For $g_j$ with $e_j=0$, one can add an auxiliary constant $\tilde{e}_j\neq 0$ such that $g_j(\boldsymbol{x},\boldsymbol{\theta})+\tilde{e}_j\leq \tilde{e}_j$ for the design and formulation consistency. The choice of $\tilde{e}_j$ can be problem dependent. For example, in our simulation, $\tilde{e}_j$ is set as the maximum slack bus generation for its lower bound limit in OPF discussed in Appendix~\ref{appendix.deepopf+}.} $ j\in{\mathcal{E}}$ by a calibration rate $\eta_j\geq0$:\footnote{Another intuitive calibration method is to keep the mean of the (calibrated) upper bound and lower bound of the constraint unchanged. That is, $(\hat{e}^u_j+\hat{e}^l_j)/2$ is the same as the one before calibration, where $\hat{e}^u_j$ and $\hat{e}^l_j$ denote the upper/lower bounds after calibration. We remark that such method 1) may not be applicable to the constraints with only one single unilateral bound, and 2) it introduces additional calibration requirements on the constraints limits compared with the one in \eqref{equ:inequality.calibration}, which could cause some constraints to have small and conservative calibration rate, while it can indeed be further calibrated. We refer to Sec.~\ref{ssec:calibrationrange} and Appendix~\ref{appen:sec.mscr} for the discussion on determining the calibration rate.}
\begin{equation}
\label{equ:inequality.calibration}
    g_j(\boldsymbol{x},\boldsymbol{\theta})\leq \hat{e}_j=  \begin{cases}
          e_{j}\left(1-\eta_{j}\right), & \mbox{if }e_{j}\geq0;\\
       e_{j}\left(1+\eta_{j}\right), & \mbox{otherwise}.
            \end{cases}
\end{equation}
 Recall that in the framework, the DNN is trained on the samples from OPCC with calibrated critical constraints as discussed in Sec.~\ref{ssec:overview}.\footnote{The non-critical constraints are always respected for any input $\boldsymbol{\theta}\in\mathcal{D}$ and $\boldsymbol{x}$ in the solution space. Hence, calibrating those constraints may only cause higher optimality loss since the reference sub-optimal solutions of OPCC with such calibrations could have larger deviations from the ones under the original setting.} However, an inappropriate calibration rate could lead to poor performance of DNN. If one adjusts the limits too much, some input $\boldsymbol{\theta}\in\mathcal{D}$ will become infeasible under the calibrated constraints and hence lead to poor generalization of the preventatively-trained DNN, though they are feasible for the original limits. Therefore, it is essential to determine the appropriate calibration range without shrinking the parameter input region $\mathcal{D}$. To this end, we solve the following bi-level optimization problem to obtain the maximum calibration rate, such that the calibrated feasibility set of $\boldsymbol{x}$ can still support the input region, i.e., the OPCC in~\eqref{equ:OPCC.formulation-1}--\eqref{equ:OPCC.formulation-4} with a reduced feasible set has a solution for any $\boldsymbol{\theta}\in \mathcal{D}$.
\begin{align}
    \min_{\boldsymbol{\theta}\in \mathcal{D}}  \max_{{\boldsymbol{x}},\nu^{c}} \;\;\;  & \nu^{c}\label{equ:calibration.rate-1} \\
    \mathrm{s.t.}\;\;\; &\eqref{equ:OPCC.formulation-3},\eqref{equ:OPCC.formulation-4} \nonumber \\ 
    & \nu^c\leq (e_j-g_j(\boldsymbol{\theta},\boldsymbol{x}))/|e_j|,\; \forall j\in\mathcal{{E}}. \label{equ:calibration.rate-2}
\end{align}
{Constraints \eqref{equ:OPCC.formulation-3}--\eqref{equ:OPCC.formulation-4} enforce the feasibility of ${\boldsymbol{x}}$ with respect to the associated input $\boldsymbol{\theta}\in\mathcal{D}$. (\ref{equ:calibration.rate-2}) represents the maximum element-wise least redundancy among all constraints, i.e., the maximum constraint calibration rate.}
 Consider the inner maximization problem, the objective finds the maximum of the element-wise least redundancy among all inequality constraints, which is the largest possible constraints calibration rate at each given $\boldsymbol{\theta}$.   Therefore, solving \eqref{equ:calibration.rate-1}--\eqref{equ:calibration.rate-2} gives the maximum allowed calibration rate among all inequality constraints for all $\boldsymbol{\theta}\in\mathcal{D}$, and correspondingly, the supported input feasible region is not reduced. We remark that though the inner maximization problem is a convex optimization problem (convex constrained with linear objective), the bi-level program \eqref{equ:calibration.rate-1}--\eqref{equ:calibration.rate-2} is challenging to solve~\cite{ben1990computational,jeroslow1985polynomial}. In the following Sec.~\ref{sssec:bilevel.calibrate} and Sec.~\ref{sssec:milp.calibrate}, we propose an applicable technique to reformulate the bi-level program utilizing the problem characteristic and discuss the optimality and the complexity of the problem. 


\subsubsection{Techniques for the Bi-Level Program and Maximum Calibration Rate}\label{sssec:bilevel.calibrate} Here several techniques can be applied to such bi-level problems. {In the following, we adopt the standard approach to reformulated bi-level program to single level by replacing the inner-level problem by its Karush-Kuhn-Tucker (KKT) conditions.\footnote{We always assume Slater's condition hold. Otherwise for some $\boldsymbol{\theta}$, the calibration rate turns to be zero.} This yields a single-level mathematical program with complementarity constraints (MPCCs). In particular, the approach contains the following two steps:}


\begin{mdframed}[backgroundcolor=gray!10, skipabove=6mm, skipbelow=3mm]
\begin{itemize}
    \item Step 1. Reformulate the bi-level program to an equivalent single-level one, by replacing the inner problem with its sufficient and necessary KKT conditions~\cite{boyd2004convex}. 
    \item  Step 1. Incorporate the KKT conditions into the upper-level program as constraints, representing the optimality of the inner maximization in $\boldsymbol{x}$.
\end{itemize}
\end{mdframed}
After solving \eqref{equ:calibration.rate-1}--\eqref{equ:calibration.rate-2}, we derive the maximum calibration rate, denoted as $\Delta$. We have the following lemma highlighting the appropriate constraints calibration rate without shrinking the original input feasible region $\mathcal{D}$, considering $\eta_{j}=\eta, \forall j\in\mathcal{E}$ in \eqref{equ:inequality.calibration}.

\begin{lemma}
If the calibration rate of the inequality constraints calibration satisfies $\eta\leq\Delta$, then any input $\boldsymbol{\theta}\in\mathcal{D}$ is feasible, i.e., for any $\boldsymbol{\theta}\in\mathcal{D}$ there exist a feasible $\boldsymbol{x}$ such that \eqref{equ:OPCC.formulation-4}, \eqref{equ:inequality.calibration} hold.
\end{lemma}

We remark that the obtained uniform calibration rate on each critical constraints forms the \textit{outer bound} of the minimum supporting calibration region defined as follows:
\begin{definition}\label{def:mimial.region}
The minimal supporting calibration region is defined as the set of calibration rate $\{\eta_j\}_{j\in\mathcal{E}}$ and for each $\boldsymbol{\theta}\in\mathcal{D}$, there exist an $\boldsymbol{x}$ such that \eqref{equ:OPCC.formulation-4}, and \eqref{equ:inequality.calibration} hold. Meanwhile, there exist a $\boldsymbol{\theta}\in\mathcal{D}$ and there does not exist an $\boldsymbol{x}$ such that \eqref{equ:OPCC.formulation-4}, \eqref{equ:inequality.calibration} hold under $\{\eta_j+\delta_j\}_{j\in\mathcal{E}}$ for any $\delta_j\geq0$ and at least one $\delta_j>0$.
\end{definition}
The minimal supporting calibration region describes the set of maximum calibration rate such that 1) the input parameter region is maintained, and 2) any further calibration on the constraints will lead some input to be infeasible. We remark that such minimal supporting calibration region is not unique; see Appendix~\ref{appen:sec.mscr} {for an example} and the approach to obtain (one of) such minimal supporting calibration region. In this work, we consider the uniform calibration rate $\Delta$ for further analysis.\footnote{{We remark that the uniform calibration method may introduce the asymmetry on the calibration magnitude as large limit would have large calibration magnitude. An alternative approach is to set the individual calibration rate $\eta_j$ for each constraint while maintain the supported input region. However, the choice of such individual calibration rates is not unique due to the non-uniqueness of the minimum supporting calibration region. We leave the analysis of such individual constraints calibration for future study. We refer to Appendix~\ref{appen:sec.mscr} for a discussion and leave it for future study.}}

Note that the reformulated problem \eqref{equ:calibration.rate-1}--\eqref{equ:calibration.rate-2} is indeed in general, still a non-convex optimization problem after such KKT replacement. Existing solvers, e.g., Gurobi, CPLEX, or APOPT,  may not generate the global optimal solution for the problem \eqref{equ:calibration.rate-1}--\eqref{equ:calibration.rate-2} {due to the challenging nature} of itself.
In the following, we present that for the special class of OPLC, e.g., mp-QP, which is also common in practice, we can improve the results by obtaining a useful lower bound.

\subsubsection{Special Case: OPLC}\label{sssec:milp.calibrate}  We remark that for the OPLC, i.e., $g_j$ are all linear, $\forall j\in\mathcal{E}$, the reformulated bi-level problem is indeed in the form of quadratically constrained program due to the complementary slackness requirements in the KKT conditions. As the input domain $\mathcal{D}$ is a convex polytope, such quadratically constrained program can be cast as the mixed-integer linear programming (MILP). See Appendix~\ref{appen:sec.bilevel.lp.reformulation} for the reformulation. In particular, we apply the following procedure to obtain a lower bound of the optimal objective in polynomial time.
\begin{mdframed}[hidealllines=true, roundcorner=10pt, backgroundcolor=gray!10, skipabove=6mm, skipbelow=3mm]
\begin{itemize}
    \item Step 1. Reformulate the bi-level program to an equivalent single-level one, by replacing the inner problem with its sufficirnt and necessary KKT conditions~\cite{boyd2004convex}. 
    \item Step 2. Transform the single-level optimization problem into a MILP by replacing the bi-linear equality constraints (comes from the complementary slackness in KKT conditions) with equivalent mixed-integer linear inequality constraints.
    \item Step 3. Solve the MILP using the branch-and-bound algorithm~\cite{lawler1966branch}. Let the obtained objective value be $\Delta\geq 0$ from \rev{the primal constraint~(\ref{equ:OPCC.formulation-3}) and constraint~(\ref{equ:calibration.rate-2})}.
\end{itemize}
\end{mdframed}
\textbf{Remark:} (i) the branch-and-bound algorithm returns $\Delta$ \rev{(lower bound of the maximum calibration rate $\nu^{c*}$)} with a polynomial time complexity of  $\mathcal{O}((M+4|\mathcal{E}|+5N)^{2.5})$~\cite{vaidya1989speeding}, where $M$ and $N$ are the dimensions of the input and decision variables, and $|\mathcal{E}|$ is the number of constraints. (ii) $\Delta$ is a lower bound to the maximum calibration rate as the algorithm may not solve the MILP problem exactly (with a non-zero optimality gap by relaxing (some of) the integer variables). Such a lower bound still guarantees that the input region is supported. If the MILP is solved to zero optimality gap, i.e., exact bound with global optimality, then we obtain the provable maximum calibration rate. (iii) If $\Delta=0$, then reducing the feasibility set may lead to no feasible solution for some input. (iv) If $\Delta >0$, then we can use it to determine the sufficient DNN size and obtain a DNN with provably universal solution feasibility guarantee as shown in Sec.~\ref{ssec:feasibilityDNN} and design the \textit{Adversarial Sample-Aware} training algorithm for desirable optimality performance without sacrificing feasibility guarantee in Sec.~\ref{ssec:ASAA}. (v) After solving \eqref{equ:calibration.rate-1}--\eqref{equ:calibration.rate-2}, we set each $\eta_j$ in \eqref{equ:inequality.calibration} to be $\Delta$, such uniform constraints calibration forms the \textit{outer bound} of the minimum supporting calibration region as defined in Definition~\ref{def:mimial.region}. See Appendix~\ref{appen:sec.mscr} for more discussion; (vi) we observe that the branch-and-bound algorithm can actually return the exact optimal objective $\nu^{c*}$ of all the reformulated MILP calibration rate programs \eqref{equ:calibration.rate-1}--\eqref{equ:calibration.rate-2} in  less than 20 mins for the numerical examples studied in Sec.~\ref{sec:simulations},

Note that such a lower bound $\Delta$ guarantees that we will not calibrate the constraints over the allowable limits such that the OPLC with calibrated constraints admits a feasible optimal solution for each input $\boldsymbol{\theta}$ in the interested parameter input region $\mathcal{D}$.\footnote{For general OPCC, we may only obtain an \textit{upper bound} on the maximum calibration rate if the proposed program is not solved global optimally which can not preserve the input region $\mathcal{D}$. Such a larger calibration rate could cause some input parameter $\boldsymbol{\theta}$ to be infeasible and hence lead the target mapping to learn (from input $\boldsymbol{\theta}$ to (sub)optimal solution of OPCC with calibrated constraints) to be illegitimate and not valid within the entire input domain $\mathcal{D}$.} In practice, one may use a smaller calibration compared with the obtain $\Delta$. We summarize the result in the following proposition.
\begin{prop}
Consider the OPLC, i.e., $g_j$ are all linear, $\forall j=1,\ldots,m$, we can obtain a lower bound on the maximum calibration rate with a time complexity $\mathcal{O}((M+4|\mathcal{E}|+5N)^{2.5})$. Such a lower bound guarantees the input parameter region $\mathcal{D}$ is preserved.
\end{prop} 

With the obtained constraints calibration rate (or its lower bound), we show how to obtain the DNN model with sufficient size to ensure the universal feasibility over the entire parameter input region despite the approximation errors in next subsection.

\subsection{Feasibility Guarantee of DNN}\label{ssec:feasibilityDNN}
In this section, we first model DNN with ReLU activations. Then, we introduce a method to determine the sufficient DNN size for guaranteeing solution feasibility.
\subsubsection{DNN Model}
After determining the proper constraints calibrations rate, we need train a DNN to learn the input-solution mapping for the problem with calibrated constraints. As discussed in Sec.~\ref{sec:OPCC_formulation}, the mapping between the input and the optimal solution of OPCC is continuous if OPCC admits a unique solution for each input $\boldsymbol{\theta}\in\mathcal{D}$~\cite{pan2020deepopf,bemporad2006algorithm}. Existing works~\cite{bemporad2006algorithm,hanin2017approximating,kidger2020universal,hornik1991approximation,karg2020efficient} show that the feed-forward neural networks demonstrate universal approximation capability and can approximate real-valued continuous functions arbitrary well for the sufficient large neural network size, indicating that there always exists a DNN size such that universal feasibility can be achieved.

We employ a DNN model with $N_{{\text{hid}}}$ hidden layers (depth) and $N_{{\text{neu}}}$ in each hidden layer (width),\footnote{The DNNs with different number of neurons can be cast to this structure by setting $N_{{\text{neu}}}$ as the maximum number of neurons among each layer and keep some parameters of the DNN as constant.} using multi-layer feed-forward neural network structure with ReLU activation function\footnote{The ReLU activation function is widely adopted with the advantage of accelerating the convergence and alleviate the vanishing gradient problem~\cite{krizhevsky2012imagenet}} to approximate the input-solution mapping for OPCC, which is defined as:
\begin{equation}
\begin{split}
    &\boldsymbol{h_0}=\boldsymbol{\theta}, \\
    &\boldsymbol{h_i}=\sigma \left( \boldsymbol{W_ih_{i-1}}+\boldsymbol{b_{i}} \right), \text{for} \ i=1,\ldots, N_{{\text{hid}}}, \\
    &\boldsymbol{\tilde{h}}=\sigma\left( \boldsymbol{W_oh_{N_{{\text{hid}}}}}+\boldsymbol{b_o}-\underline{\boldsymbol{x}}\right)+\underline{\boldsymbol{x}},\\
    & \hat{\boldsymbol{x}}=-\sigma\left( \bar{\boldsymbol{x}}-\boldsymbol{\tilde{h}}\right)+\bar{\boldsymbol{x}}.
\end{split}
\label{dnn.model}
\end{equation}
where $\boldsymbol{\theta}$ is the input parameter of the OPCC and forms the input of the DNN. $\boldsymbol{h_i}$ is the output the $i$-th layer. $\boldsymbol{W_i}$ and $\boldsymbol{b_i}$ are the $i$-th layer's weight matrix and bias, respectively. $\sigma(x)=\max(x,0)$ is the ReLU activation function, taking element-wise max operation over the input vector. Here $\tilde{\boldsymbol{h}}$ is the intermediate vector enforcing the lower bound feasibility of predictions. The final output $\hat{\boldsymbol{x}}=\{\hat{x}_i\}_{i=1,...,N}$ further satisfies upper bounds. $\bar{\boldsymbol{x}}$, $\underline{\boldsymbol{x}}$ are the upper bounds and lower bounds of the decision variables respectively. We remark that the last two operations in \eqref{dnn.model} enforces the feasibility of predicted solution $\hat{{x}}_i$ w.r.t.~\eqref{equ:OPCC.formulation-4} to be within its lower bound $\underline{x}_i$ and upper bound $\bar{x}_i$. Here we present the last two \textit{clamp}-equivalent actions as (\ref{dnn.model}) for further DNN analysis.

\subsubsection{The Input-Output Relations of DNN with ReLU Activation}
\rev{To better include the DNN equations in our designed optimization to analysis DNN's worst case feasibility guarantee performance,} we adopt the technique in~\cite{tjeng2018evaluating} to reformulate the ReLU activations expression in~(\ref{dnn.model}).\footnote{We remark that there exist other $\max()$ reformulation methods, e.g., MPEC reformulation (which can also be cast as the integer formulation equivalently). In this work, we focus on the mixed-integer linear expression as shown in \eqref{equ.DNNinteger-1}--\eqref{equ:DNNinteger-upper} for an analysis. Such an expression shows benefits when designing the framework as discussed in Sec.~\ref{sssec:opcc.size} and Sec.~\ref{sssec:opcc.algorithm}} For $i=1,\ldots,N_{\text{hid}}$, let $\boldsymbol{\hat{h}}_i$ denote $\boldsymbol{W_ih_{i-1}+b_{i}}$. The output of neuron with ReLU activation is represented as: for $k=1,\ldots,N_{\text{neu}}$ and $i=1,\ldots,N_{\text{hid}}$,
\begin{align} 
    \hat{h}_{i}^{k} \leq h_{i}^{k} & \le \hat{h}_{i}^{k}-h_{i}^{\min ,k}(1-z_{i}^{k}), \label{equ.DNNinteger-1}\\
    h_{i}^{k}& \le h_{i}^{\max ,k}z_{i}^{k},\label{equ.DNNinteger-2}\\
    h_{i}^{k}& \ge 0, \, z_{i}^{k}\in \{0,1\}. \label{equ.DNNinteger-3}
\end{align} 
Here we use the superscript $k$ to denote the $k$-th element of a vector. $z_{i}^{k}$ are (auxiliary) binary variables indicates the state of the corresponding neuron, i.e., 1 (resp. 0) indicates activated (resp. non-activated).  That is, when the input to the $i$-th neuron in layer $k$, $\hat{h}^i_k\leq 0$, the corresponding binary variable $z^i_k$ is $0$ such that the last two inequalities~\eqref{equ.DNNinteger-2}--\eqref{equ.DNNinteger-3} contain it to zero while the first two are not binding if $\hat{h}^i_k< 0$. Similarly, when $\hat{h}^i_k\geq 0$, the corresponding binary variable $z^i_k$ is $1$ such that the first two inequalities in~\eqref{equ.DNNinteger-1} contain it to $\hat{h}^i_k$ while the last two are not binding if $\hat{h}^i_k> 0$.

 We remark that given the value of DNN weights and bias, \rev{$z_{i}^{k}$ can be determined ($z_{i}^{k}$ can be either 0/1 if $\hat{h}^k_i=0$) for each input $\boldsymbol{\theta}$.} $h^{\max,k}_i/h^{\min,k}_i$ are the upper/lower bound on the neuron outputs. \rev{See Appendix~\ref{appen.ssec.upper.lower.neuron} for a discussion.}  Similarly, the last two operations in \eqref{dnn.model} can also be reformulated. Let $\boldsymbol{\hat{h}}_{\text{out}}$ denote $\boldsymbol{W_oh_{N_{\text{hid}}}+b_{o}}-\underline{\boldsymbol{x}}$, for $k^l=1,\ldots,N$ and $k^u=1,\ldots,N$:
\begin{equation} 
\begin{split} {\hat{h}}_{\text{out}}^{k^l}+\underline{x}_{k^l} \leq \tilde{h}^{k^l} &\le {\hat{h}}_{\text{out}}^{k^l}+\underline{x}_k-h_{\text{out}}^{\min ,k}(1-z_{i}^{k^l}), \\
    \tilde{h}^{k^l}& \le h_{\text{out}}^{\max ,{k^l}}z_{\text{out}}^{k^l}+\underline{x}_{k^l},\\
    \tilde{h}^{k^l}& \ge \underline{x}_{k^l}, \, z_{\text{out}}^{k^l}\in \{0,1\},
\end{split}\label{equ:DNNinteger-lower}
\end{equation} 
\begin{equation} 
\begin{split}
    \tilde{h}^{k^u}+\hat{x}^{\min ,k}(1-z_{i}^{k}) \leq \hat{x}^{k^u} &\le\tilde{h}^{k^u}, \\
    \hat{x}^{k^u}& \geq -\hat{x}^{\max ,k}z_{\text{out}}^{k}+\bar{x}_{k^u},\\
    \hat{x}^{k^u}& \leq \bar{x}_{k^u}, \, z_{\text{out}}^{k^u}\in \{0,1\},
\end{split}\label{equ:DNNinteger-upper}
\end{equation} 
where $h^{\max,k}_{\text{out}}$, $h^{\min,k}_{\text{out}}$, $\hat{x}^{\max,k}$, and $\hat{x}^{\min,k}$ are the corresponding upper/lower bounds.
With~(\ref{equ.DNNinteger-1})-(\ref{equ.DNNinteger-2}), the input-output relationship in DNN can be represented with a set of mixed-integer linear inequalities. We discuss how to employ~(\ref{equ.DNNinteger-1})-(\ref{equ.DNNinteger-2}) to determine the sufficient DNN size in guaranteeing universal feasibility {in Sec.~\ref{ssec.suff.dnn.size}.} For ease of representation, we use $(\mathbf{W},\bold{b})$ to denote DNN weights and bias in the following.

Typically, the DNN is trained to minimize the average of the specified loss function
among the training set by optimizing the the value of $(\mathbf{W},\bold{b})$.
{In the previous work, the training (test) set is generally obtained by sampling the input data according to some distribution to train (evaluate) the DNN performance~\cite{pan2020deepopf,zhao2020deepopf}. However, the DNN model obtained from such approaches may not achieve good feasibility performance over the entire input domain $\mathcal{D}$ especially considering the worst-case {scenarios~\cite{venzke2020learning,nellikkath2021physics,nellikkath2021physics2}.} In the following, we study the worst-case performance of DNN and determine the sufficient DNN size so that for any possible input from the input region, the resulting DNN solution is guaranteed to be feasible w.r.t. inequality constraints.}

\subsubsection{Sufficient DNN Size in Guaranteeing Universal Feasibility}\label{ssec.suff.dnn.size}
As an essential methodological contribution, we propose an iterative approach to determine the sufficient DNN size for guaranteeing universal solution feasibility in the input region. The idea is to iteratively verify whether the worst-case prediction error of the given DNN model is within the room of error (maximum calibration rate), and doubles the DNN's width (with fixed depth) if not, until the worst-case prediction error does not exceed the tolerated range. We outline the design of the proposed approach below, under the setting where all hidden layers share the same width. Let the depth and (initial) width of the DNN model be $N_{\text{hid}}$  and $N_{\text{neu}}$, respectively. \rev{Here we define \textit{universal solution feasibility} as that for any input $\boldsymbol{\theta}\in\mathcal{D}$, the output of DNN always satisfies (\ref{equ:OPCC.formulation-3})-(\ref{equ:OPCC.formulation-4}).}

For each iteration, the proposed approach first evaluates the least maximum relative violations among all constraints for all $\boldsymbol{\theta}\in\mathcal{D}$ for the current DNN model via solving the following bi-level program: 
\begin{align}
    \min_{\mathbf{W}, \bold{b}} \max_{\boldsymbol{\theta}\in\mathcal{D}} \;\; & {\nu}^f  \label{equ:DNNsize-1}\\
    \mathrm{s.t.} \;\; & \eqref{equ.DNNinteger-1}-\eqref{equ.DNNinteger-3},  1\leq i\leq N_{{\text{hid}}}, 1\leq k\leq N_{\text{neu}}, \nonumber\\
    &\eqref{equ:DNNinteger-lower},\eqref{equ:DNNinteger-upper},  1\leq k^l\leq N, 1\leq k^u\leq N, \nonumber\\
    &{\nu}^f=\max_{j\in\mathcal{E}}\{(g_j(\boldsymbol{\theta},\hat{\boldsymbol{x}})- \hat{e}_j)/|e_j|\}.
    \label{equ:DNNsize-2}
\end{align}
where \eqref{equ.DNNinteger-1}-\eqref{equ:DNNinteger-upper} express the outcome of the DNN as a function of input $\boldsymbol{\theta}$. \eqref{equ:DNNsize-2} denotes the relative violation on each constraint considering the limits calibration, where $\hat{e}_j=(1_{e_j\geq0}(1-\Delta)+1_{e_j<0}(1+\Delta))\cdot e_j$ denotes the constraint limit after calibration and $\Delta$ represents the determined calibration rate via \eqref{equ:calibration.rate-1}--\eqref{equ:calibration.rate-2}. Thus, solving (\ref{equ:DNNsize-1})-(\ref{equ:DNNsize-2}) gives the least maximum DNN constraint violation over the input region $\mathcal{D}$. Here recall that for the class of OPLC, the obtained $\Delta$ is no greater than the maximum one\footnote{For OPLC, the obtained $\Delta$ is the global maximum or the lower bound of the maximum calibration rate. See Sec.~\ref{sssec:milp.calibrate} for the discussion.} with which the target input parameter region is still guaranteed to be preserved. We remark that the maximum violation \eqref{equ:DNNsize-2} can be reformulated as a set of mixed-integer inequalities. See Appendix~\ref{appen:proof.of.calibration} for details.

Consider the inner maximization problem, the objective ${\nu}^f$ hence finds the maximum violation among all the constraints consider the worst-case input $\boldsymbol{\theta}$ given the value of DNN parameters $(\mathbf{W}, \bold{b})$. Here \eqref{equ:DNNsize-1}--\eqref{equ:DNNsize-2} express the outcome of the DNN as a function of input $\boldsymbol{\theta}$. Thus, solving \eqref{equ:DNNsize-1}--\eqref{equ:DNNsize-2} gives the least maximum DNN constraint violation over the input region $\mathcal{D}$, representing the learning ability of given DNN size in ensuring feasibility of the predicted solutions considering the worst-case input in $\boldsymbol{\theta}$, given its best performance. We remark that \eqref{equ:DNNsize-1}--\eqref{equ:DNNsize-2} is a non-convex mixed-integer linear bi-level program due to the non-convex equality constraints related to the ReLU activations and the maximum operator in \eqref{equ:DNNsize-2}.
Since the inner maximization problem is a mixed-integer nonlinear program, the techniques for convex bi-level programs discussed in Sec.~\ref{ssec:calibrationrange} are not applicable, i.e., replacing the lower-level optimization problem by its KKT conditions. To solve such bi-level optimization problem, we apply the \textit{Danskin's Theorem} idea to optimize the upper-level variables $(\mathbf{W}, \bold{b})$ by gradient descent. This would simply require to 1) find the maximum of the inner problem, and 2) compute the normal gradient evaluated at this point~\cite{danskin,danskin2012theory}. We refer interested readers to~\cite{kolter2018adversarial,madry2018towards} and Appendix~\ref{appen:sec.danskin} for the detailed procedures and discussions. 

 Let $\rho$ be the obtained optimal objective value of the bi-level problem and $(\mathbf{W}^{f}, \bold{b}^{f})$ be the corresponding DNN parameters. With such DNN parameters, we can directly construct a DNN model. Recall that the determined calibration rate is $\Delta$. The proposed approach then verifies whether the constructed DNN model is sufficient for guaranteeing feasibility by the following proposition.
\begin{prop}\label{prop:violation.vs.calibration}
    Consider the DNN with $N_{\text{hid}}$ hidden layers each having $N_{\text{neu}}$ neurons and parameters $(\mathbf{W}^{f}, \bold{b}^{f})$. If $\rho \leq \Delta$, then for any $\boldsymbol{\theta}\in\mathcal{D}$, the solution of this DNN is feasible w.r.t \eqref{equ:OPCC.formulation-3}--\eqref{equ:OPCC.formulation-4}.\footnote{Note that the non-critical constraints are always respected from Definition~\ref{def:critical}.}
\end{prop}

The proof is shown in Appendix~\ref{appen:proof.of.calibration}. Proposition~\ref{prop:violation.vs.calibration} states that if $\rho \leq \Delta$, the worst-case prediction error of current DNN model is within the room of maximum calibration rate, i.e., the largest violation at the calibrated inequality constraints is no greater than the calibration rate. Therefore, the current DNN size is capable of achieving zero violation at all original inequality constraints for all inputs $\boldsymbol{\theta}\in\mathcal{D}$ and hence sufficient for guaranteeing universal feasibility; otherwise, it doubles the width of DNN and moves to the next iteration. We remark that solving \eqref{equ:DNNsize-1}--\eqref{equ:DNNsize-2} can be essentially seen as the training process of the DNN with the calibrated constraints (the iterative approach with gradient decent) such that the maximum violation is minimized from the outer minimization problem over the DNN parameters $(\mathbf{W}^{f}, \bold{b}^{f})$.

The above program helps to verify whether a certain DNN size is capable of achieving universal feasibility within the input parameter region. If $\rho> \Delta$, meaning that the test DNN fails to preserve universal feasibility, and we need to enlarge the DNN size, e.g., increase the number of neurons on each layer, such that universal feasibility of DNN solution can be guaranteed. Recall that the target mapping (from input $\boldsymbol{\theta}$ to (sub)optimal solution of OPCC with calibrated constraints) is continuous. As such, there exists a DNN such that the universal feasibility of the generated solution is guaranteed given the DNN size (width $N_{\text{neu}}$) is sufficiently large according to the universal approximation capability~\cite{bemporad2006algorithm,hanin2017approximating,kidger2020universal,hornik1991approximation,karg2020efficient} of DNNs. We highlight the claim of \textit{Universal Approximation} of DNNs in the following proposition.
\begin{prop}~\cite{bemporad2006algorithm,hanin2017approximating,kidger2020universal,hornik1991approximation,karg2020efficient}\label{prop:universal}
Assume the target function to learn is continuous, there always exists a DNN whose output function can approach the target function arbitrarily well, i.e., 
$${\displaystyle \max _{\boldsymbol{\theta}\in \mathcal{D}}\,\|h(\boldsymbol{\theta})-\hat{h}(\boldsymbol{\theta})\|<\varepsilon,}$$
hold for any $\epsilon$ arbitrarily small (distance from $h$ to $\hat{h}$ can be infinitely small). Here $h(\boldsymbol{\theta})$ and $\hat{h}(\boldsymbol{\theta})$ represent the target mapping to be approximated and the DNN function respectively. 

Furthermore, given the fixed depth $N_{\text{hid}}$ of the DNN, the learning ability of the DNN is increasing monotonically w.r.t. the width of the DNN. That is, consider two DNN width $N^1_{\text{neu}}$ and $N^2_{\text{neu}}$ such that $N^1_{\text{neu}}>N^2_{\text{neu}}$, we have
$${\displaystyle \min_{h\in\mathcal{C}^{N^1_{\text{neu}}}}\max _{\boldsymbol{\theta}\in \mathcal{D}}\,\|h(\boldsymbol{\theta})-\hat{h}(\boldsymbol{\theta})\|\leq\min_{h\in\mathcal{C}^{N^2_{\text{neu}}}}\max _{\boldsymbol{\theta}\in \mathcal{D}}\,\|h(\boldsymbol{\theta})-\hat{h}(\boldsymbol{\theta})\|,}$$
where $\mathcal{C}^{N^1_{\text{neu}}}$ and $\mathcal{C}^{N^2_{\text{neu}}}$ denote the class of all functions generated by a $N_{\text{hid}}$ depth neural network with width $N^1_{\text{neu}}$ and $N^2_{\text{neu}}$ respectively. 
\end{prop}
Proposition~\ref{prop:universal} provides us a strong observation and theoretical basis for further designing the iterative approach to determine the sufficient DNN size in guaranteeing universal feasibility.

\begin{algorithm}[!t]
\caption{Proposed Approach for Sufficient DNN Size}
\label{alg:double.sufficient.dnn.size}
\begin{algorithmic}[1]
   \STATE {\bfseries Input:} $\Delta$; Initial width $N_{\text{neu}}^{\text{init}}$
   \STATE {\bfseries Output:} Determined DNN width: $N_{\text{neu}}^*$
   \STATE Set $t=0$
   \STATE Set $N_{\text{neu}}^{t} = N_{\text{neu}}^{\text{init}}$
   \STATE Obtain $\rho$ via solving \eqref{equ:DNNsize-1}--\eqref{equ:DNNsize-2}
    \WHILE{$\rho \geq \Delta$}
   \STATE Set $N_{\text{neu}}^{t+1} = 2\times N_{\text{neu}}^{t}$
   \STATE Set $t=t+1$
   \STATE Solve \eqref{equ:DNNsize-1}--\eqref{equ:DNNsize-2} and update $\rho$ 
    \ENDWHILE
    \STATE Set $N_{\text{neu}}^*=N_{\text{neu}}^{t}$
    \STATE {\bfseries Return:} $N_{\text{neu}}^*$
\end{algorithmic}
\end{algorithm}

{\paragraph{Iterative Approach for Sufficient DNN Size} In the following, we propose an iterative approach to determine the sufficient DNN size such that universal feasibility is guaranteed. We start with the initial DNN model with depth $N_{\text{hid}}$ and width $N_{\text{neu}}$ at iteration $t=0$ (line 3-4).
\begin{mdframed}[backgroundcolor=gray!10, skipabove=6mm, skipbelow=3mm]
\begin{itemize}
    \item Step 1. At iteration $t$, verify the universal feasibility guarantee of DNN with depth $N^t_{\text{hid}}$ and width $N^t_{\text{neu}}$ by solving \eqref{equ:DNNsize-1}--\eqref{equ:DNNsize-2}. If the obtained value $\rho\leq\Delta$, stop the iteration (line 5 and line 9).
\item  Step 2. If $\rho>\Delta$, double the DNN width $N_{\text{neu}}^{t+1} = 2\cdot N_{\text{neu}}^{t}$ and proceed to the next iteration $t+1$. Go to Step 1 (line 6-8).
\end{itemize}
\end{mdframed}

The details of the proposed approach are shown in Algorithm~\ref{alg:double.sufficient.dnn.size}. It repeatedly compare the obtained maximum constraints violation ($\rho$) with the calibration rate ($\Delta$), doubles the DNN width, and return the width as $N_{\text{neu}}^*$ until $\rho \leq \Delta$. The above approach is expected to determine the sufficient DNN size (width) that is capable of achieving universal feasibility w.r.t. the input domain $\mathcal{D}$, i.e., $\rho\leq\Delta$, if the DNN width is large enough~\cite{hornik1991approximation}.\footnote{{One can also increase the DNN depth to achieve universal approximation for more degree of freedom in DNN parameters. In this work, we focus on increasing the DNN width for sufficient DNN learning ability.}}  Thus, we could construct a feasibility-guaranteed DNN
model by the proposed approach, namely DNN-FG as shown in Fig~\ref{fig:framework}. Note that if the initial tested DNN size guarantees universal feasibility, we do not need the above doubling approach to further expand the DNN size but keep it as the sufficient one.

We remark that the obtained sufficient DNN size by doubling the DNN width may be substantial, introducing additional training time to train the DNN model and higher computational time when applied to solve OPCC. One can also determine the corresponding minimal sufficient DNN size by a simple and efficient binary search between 
\begin{mdframed}[backgroundcolor=gray!10, skipabove=6mm, skipbelow=3mm]
\begin{itemize}
    \item the obtained sufficient DNN size $N_{\text{neu}}^*$ and the pre-obtained DNN size $N_{\text{neu}}^*/2$ (before doubling the DNN width) which fails to achieve universal feasibility, if the initial tested DNN can not guarantee universal feasibility;
    \item the initial tested DNN size and some small DNN, e.g., zero width DNN, if the initial tested DNN size is sufficient in guaranteeing universal feasibility.
\end{itemize}
\end{mdframed}
Such a minimal sufficient DNN size denotes the minimal width required for a given DNN structure with depth $N_{\text{hid}}$ to achieve universal feasibility within the entire input domain. We use $\hat{N}_{\text{neu}}$ to denote the determined minimal sufficient DNN size and propose the following proposition.
\begin{prop}\label{prop:opcc:size}
Consider the DNN width $\hat{N}_{\text{neu}}$ and assume \eqref{equ:DNNsize-1}--\eqref{equ:DNNsize-2} is solved global optimally such that $\rho\leq\Delta$, any DNN with depth $N_{\text{hid}}$ and a smaller width than $\hat{N}_{\text{neu}}$ can not guarantee universal feasibility for all input $\boldsymbol{\theta}\in\mathcal{D}$. Meanwhile, any DNN with depth $N_{\text{hid}}$ and at least $\hat{N}_{\text{neu}}$ width can always achieve universal feasibility. Furthermore, one can construct a feasibility-guaranteed DNN with the corresponding obtained DNN parameters $(\mathbf{W}^{f}, \bold{b}^{f})$ such that for any $\theta\in\mathcal{D}$, the solution of this DNN is feasible w.r.t. \eqref{equ:OPCC.formulation-3}--\eqref{equ:OPCC.formulation-4}.
\end{prop}

It is worth noticing that the above result is based on the condition that we can obtain the global optimal solution of \eqref{equ:DNNsize-1}--\eqref{equ:DNNsize-2}.  However, one should note that the applied procedures based on \textit{Danskin's Theorem} are not guaranteed to provide the global optimal one. In addition, the inner maximization of \eqref{equ:DNNsize-1}--\eqref{equ:DNNsize-2} is indeed a non-convex mixed-integer nonlinear program due to non-convex equality constraints related to the ReLU activations for general OPCC. The existing solvers, e.g., APOPT, YALMIP, or Gurobi, may not be able to generate the global optimal solution. We refer to Appendix~\ref{appen:sec.danskin} for a discussion on the relationship between the obtained value and the global optimal one for general OPCC. Despite the non-global optimality of the solvers/approach, we remark that for the class of OPLC, we can still obtain a useful upper bound for further analysis.}

\subsubsection{Special Case: OPLC}\label{sssec:opcc.size}
We remark that for the class of OPLC, i.e., $g_j$ are all linear, $\forall j=1,\ldots,m$, the inner problem of \eqref{equ:DNNsize-1}--\eqref{equ:DNNsize-2} is indeed an MILP. Though it is challenging to solve the bi-level problem (\ref{equ:DNNsize-1})-(\ref{equ:DNNsize-2}) exactly, we can actually obtain $\rho$ as an upper bound on its optimal objective value if program \eqref{equ:DNNsize-1}--\eqref{equ:DNNsize-2} is not solved to global optimum, meaning that the maximum violation is not beyond such a rate.


Though such an upper bound might not be tight, as discussed in the following proposition, it is still useful for analyzing universal solution feasibility, indicating that it is guaranteed to achieve universal feasibility with such a DNN size if it is no greater than $\Delta$. In addition, despite the difficulty of the mixed-integer non-convex programs that we need to solve repeatedly, we can always obtain a feasible (sub-optimal) solution for further use for both general OPCC and OPLC.\footnote{Such a feasible (sub-optimal) solution can be easily obtained by a heuristic trial of some particular $\boldsymbol{\theta}$, e.g., the worst-case input at the previous round as the initial point and the associate integer values in the constraints, which are fixed given the specification of DNN parameters.} Our simulation results in Sec.~\ref{sec:simulations} demonstrate such observation. We highlight the result in the following proposition.
\begin{prop}\label{prop:sufficient.size}
    Consider the OPLC, i.e., $g_j$ are all linear, $\forall j=1,\ldots,m$, and assume $\Delta>0$, Algorithm~\ref{alg:double.sufficient.dnn.size} is guaranteed to terminate in finite number of iterations. At each iteration $t$, consider the DNN with $N_{\text{hid}}$ hidden layers each having $N^t_{\text{neu}}$ neurons, we can obtain $\rho$ as an upper bound to the optimal objective of \eqref{equ:DNNsize-1}--\eqref{equ:DNNsize-2} with a time complexity $\mathcal{O}((M+|\mathcal{E}|+2N_{\text{hid}} N^t_{\text{neu}}+4N)^{2.5})$. If $\rho \leq \Delta$, then the DNN with depth $N_{\text{hid}}$ and width $N^t_{\text{neu}}$ is sufficient in guaranteeing universal feasibility. Furthermore, one can construct a feasibility-guaranteed DNN with  the corresponding obtained DNN parameters $(\mathbf{W}^{f}, \bold{b}^{f})$ such that for any $\theta\in\mathcal{D}$, the solution of this DNN is feasible w.r.t. \eqref{equ:OPCC.formulation-3}--\eqref{equ:OPCC.formulation-4}.
\end{prop}
Proposition~\ref{prop:sufficient.size} indicates $\rho$ can be obtained in polynomial time. If $\rho \leq \Delta$, it means the current DNN size is sufficient to preserve universal solution feasibility in the input region; otherwise, it means the current DNN size may not be sufficient for the purpose and it needs to double the DNN width. In our case study in Sec.~\ref{sec:simulations}, we observe that the evaluated initial DNN size can always guarantee universal feasibility with a non-positive worst-case constraints violation, and we hence further conduct simulations with such determined sufficient DNN size $N_{\text{neu}}^*$
and leave the analysis of finding the minimal sufficient DNN size (width) $\hat{N}_{\text{neu}}$ and solving the problem \eqref{equ:DNNsize-1}--\eqref{equ:DNNsize-2} global optimally for general OPCC for future investigation.

\subsection{Adversarial Sample-Aware Algorithm}\label{ssec:ASAA}

While we can directly construct a feasibility-guaranteed DNN (without training) as shown in Proposition~\ref{prop:opcc:size} and Proposition~\ref{prop:sufficient.size}, it may not achieve strong optimality performance. We investigate the performances and approximation accuracy of such DNN in the case study in Sec.~\ref{ssec:effect.asaa}. To address this issue, we propose an \textit{Adversarial Sample-Aware} algorithm to further improve the optimality performance while guaranteeing universal feasibility within the input domain in this subsection. Overall, we can obtain two DNNs from the framework. Though the first one constructed from the step of determining the sufficient DNN size in Sec.~\ref{ssec:feasibilityDNN} can guarantee universal feasibility (DNN-FG), the other DNN obtained from the proposed \textit{Adversarial Sample-Aware} training algorithm in this subsection further improves optimality without sacrificing DNN's universal feasibility guarantee (DNN Optimality Enhanced).

\begin{algorithm}[!th]
\caption{Adversarial Sample-Aware algorithm}\label{alg:alg.asaa}
\algorithmfootnote{*We only include feasible $\boldsymbol{\theta}^{i,k}\in\mathcal{D}$ into $S^i$. Here $\epsilon_0$ is set to be $0$ such that $(\boldsymbol{\theta}^{i},\text{OPCC}^c(\boldsymbol{\theta}^{i}))\in S^i$ and each element of $\epsilon_k\in\mathcal{R}^{M}$ is from a uniform distribution $U(-a,a)$ for $k=1,\ldots,K$. 
$\odot$ denotes the element-wise multiplication operation among two vectors (Hadamard product).
}
\LinesNumbered 
\KwIn{Training epochs $T$, Number of iterations $I$, Initial training set $\mathcal{T}^0$, Number of auxiliary training samples $K$, Constant $a$ for constructing auxiliary training set, Calibration rate $\Delta$}
\KwOut{Feasibility-guaranteed DNN model with parameters  $(\mathbf{W}^{*}, \bold{b}^{*})$}
Pre-train the DNN model on $\mathcal{T}^{0}$ using loss function ~\eqref{equ:loss.function} for $T$ epochs\\
Save the parameters of the pre-trained initial DNN model as $(\mathbf{W}^{0}, \bold{b}^{0})$\\
\For{$i=0$ {\bfseries to} $I$}{
       Find the maximum violation of $(\mathbf{W}^{i}, \bold{b}^{i})$ by solving:
      \begin{align*}
        \boldsymbol{\theta}^i= \arg \max_{\boldsymbol{\theta}\in\mathcal{D}} \;\; {\nu}^f   \quad \mathrm{s.t.} \;\; &\eqref{equ:DNNsize-2},
        \\&\eqref{equ.DNNinteger-1}-\eqref{equ.DNNinteger-3},  1\leq i\leq N_{{\text{hid}}}, 1\leq k\leq N_{\text{neu}},\\
        &\eqref{equ:DNNinteger-lower},\eqref{equ:DNNinteger-upper},  1\leq k^l\leq N, 1\leq k^u\leq N
        \end{align*}\\
        {Set $\gamma = \nu^f(\boldsymbol{\theta}^i)$ as the optimal value of the above program at solution $\boldsymbol{\theta}^i$\\}
        \eIf{$\gamma\leq\Delta$}{
   Set $\mathbf{W}^{*}=\mathbf{W}^{i}, \bold{b}^{*} = \bold{b}^{i}$; Break
   }{
   {Construct  $(\boldsymbol{\theta}^{i}_k,\boldsymbol{x}^{i}_k)$ pair set $\mathcal{S}^i$ by uniformly sampling centered around $\boldsymbol{\theta}^i$ with calibrated OPCC ($\text{OPCC}^c$) solutions,} for $k=0,1,\ldots,K$:
   \begin{equation*}
        \boldsymbol{\theta}^{i}_k=\boldsymbol{\theta}^i \odot (\boldsymbol{1}+\epsilon_k)\in\mathcal{D}, \epsilon^{(j)}_k\sim U(-a,a), \boldsymbol{x}^{i}_k=\text{OPCC}^c(\boldsymbol{\theta}^{i}_k)   
    \end{equation*}
   }
        Set $\mathcal{T}^{i+1} = \mathcal{T}^{i} \cup \mathcal{S}^i$ and $(\mathbf{W}^{i}_0, \bold{b}^{i}_0)=(\mathbf{W}^{i}, \bold{b}^{i})$\\
		\For{$t=0$ {\bfseries to} $T$}{
             Initial DNN with parameters $(\mathbf{W}^{i}_t, \bold{b}^{i}_t)$ and train on $\mathcal{T}^{i+1}$ using loss function \eqref{equ:loss.function} and update parameters to $(\mathbf{W}^{i}_{t+1}, \bold{b}^{i}_{t+1})$;\\
             Feed each $\boldsymbol{\theta}^{i}_k\in S^i$ in the DNN model with parameters $(\mathbf{W}^{i}_{t+1}, \bold{b}^{i}_{t+1})$ to obtain predicted solution $\hat{\boldsymbol{x}}^{i}_k$;\\
             \uIf{Each $\hat{\boldsymbol{x}}^{i}_k$ is feasible w.r.t the original constraints \eqref{equ:OPCC.formulation-3}--\eqref{equ:OPCC.formulation-4}}
             {Set $(\mathbf{W}^{i+1}, \bold{b}^{i+1})=(\mathbf{W}^{i}_{t+1}, \bold{b}^{i}_{t+1})$;
             Break}
             \ElseIf{$t=T-1$}{Set $(\mathbf{W}^{i+1}, \bold{b}^{i+1})=(\mathbf{W}^{i}_{t+1}, \bold{b}^{i}_{t+1})$}
		}
     }
\end{algorithm}

The proposed algorithm adopts adversarial learning idea~\cite{chakraborty2018adversarial}, e.g., adaptively incorporates adversarial inputs with violation for improving the DNN robustness. {Furthermore, it leverage the technique of active learning~\cite{ren2020survey} to improve the training efficiency by sampling around such identified adversarial inputs and apply the preventive training scheme to enhance the feasibility performance.}  In particular, the algorithm identifies the worst-case inputs identification and attempts to improve the DNN approximation ability around these adversarial inputs with violations, i.e., better learning the specific mapping information enclosing some particular input points. The corresponding pseudocode is given in Algorithm~\ref{alg:alg.asaa}. We outline the algorithm in the following. Denote the initial training set as $\mathcal{T}^{0}$, containing randomly-generated input and the corresponding ground-truth obtained by solving the calibrated OPCC (with calibration rate $\Delta$). The proposed \textit{Adversarial Sample-Aware} algorithm adopts the supervised learning approach and first pre-trains a DNN model with the sufficient size determined by the approach discussed in Sec.~\ref{ssec.suff.dnn.size}, using the initial training set $\mathcal{T}^{0}$ and the following loss function $\mathcal{L}$ for each instance:
\begin{equation}
\mathcal{L} =\frac{w_1}{N}\left\| \boldsymbol{\hat{x}}-\boldsymbol{x}^* \right\|^2 _2+\frac{w_2}{|\mathcal{E} |}\sum_{j\in \mathcal{E}}{\max}(g_j(\boldsymbol{\hat{x}},\boldsymbol{\theta })-\hat{e}_j,0).
\label{equ:loss.function}
\end{equation}
{We leverage the penalty-based training idea in~(\ref{equ:loss.function}).} The first term is the mean square error between DNN prediction $\hat{\boldsymbol{x}}$ and {the ground-truth $\boldsymbol{x}^*$ provided by the solver for each input.} The second term is the inequality constraints violation w.r.t calibrated limits $\hat{e}_j$. $w_1$ and $w_2$ are positive weighting factors to balance prediction error and penalty. We remark that after the constraints calibration, the penalty loss is with respect to the adjusted limits $\hat{e}_j$ discussed in Sec.~\ref{ssec:calibrationrange}.  The training processing can be regarded as minimizing the average value of loss function with the given training data by tuning the parameters of the DNN model, {including each layer's connection weight matrix and bias vector.} {Hence, training DNN by minimizing~(\ref{equ:loss.function}) can pursue a strong optimality performance as DNN prediction error is also minimized.} {This step corresponds to line 1-2 in Algorithm~\ref{alg:alg.asaa}}. {However, traditional penalty-based training by only minimizing~(\ref{equ:loss.function}) can not guarantee universal feasibility~\cite{venzke2020learning,deepopf2}. To address this issue,} the \textit{Adversarial Sample-Aware} algorithm then repeatedly updates the DNN model, containing the following two techniques:

\textbf{Adversarial sample identification.} The framework sequentially identifies the worst-case input in {the entire input} $\mathcal{D}$, at which constraints violations happens given the specification of DNN parameters. This step helps test whether universal feasibility is achieved and find out the potential adversarial inputs that cause infeasibility. {This step corresponds to line 4-5 in Algorithm~\ref{alg:alg.asaa}}.

\textbf{Training based on adversarial inputs.} We correct the DNN approximation behavior by involving the specific mapping information around the identified adversarial samples. In particular, we sequentially include the worst-case inputs identified in the previous step into the existing training set, anticipating the post-trained DNNs on the new training set can eliminate violations around such inputs by improving its approximation ability around them. This step corresponds to line 6-17 in Algorithm~\ref{alg:alg.asaa}.

 Specifically, given current DNN parameters, the algorithm finds the worst-case input $\boldsymbol{\theta}^i\in\mathcal{D}$ by solving the inner maximization problem of~\eqref{equ:DNNsize-1}--\eqref{equ:DNNsize-2}. Let $\gamma$ be the obtained optimal objective value. Recall that the calibration rate is $\Delta$. If $\gamma \leq \Delta$, the algorithm terminates; otherwise, it incorporates a subset of samples {randomly sampled around $\boldsymbol{\theta}^i$ and solves the calibrated OPCC with $\Delta$,} and starts a new round of training. We remark that the proposed \textit{Adversarial Sample-Aware} algorithm is expected to achieve universal feasibility within the entire input domain while preserving desirable optimality performance. The underlining reason lies in that during the adversarial sample identification-training process, both optimality (represented by the prediction error in the first term in \eqref{equ:loss.function}) and feasibility (represented by the penalty w.r.t. the constraints violations in the second term in \eqref{equ:loss.function}) are considered by training the DNN on such algorithmic designed training set. Therefore, the obtained DNN can improve the feasibility and optimality performance simultaneously by having better approximation accuracy on these samples. We present the detailed steps in the following.

For better representation, we use OPCC($\boldsymbol{\theta}$) and $\text{OPCC}^c(\boldsymbol{\theta})$ to denote the optimal solutions of the OPCC problem and the OPCC problem with constraints calibrations given the input $\boldsymbol{\theta}$. We remark that with the obtained lower bound on the maximum calibration rate $\Delta$, such constraints calibrations only leads to the (sub)-optimal solutions that are interior points within the original feasible region (the inequality constraints are expected to be not binding) while the input parameter region $\mathcal{D}$ in consideration is not reduced.\footnote{It is expected that a larger calibration rate can help to improve the DNN solution feasibility during training. With a smaller calibration rate (lower bound), one may need to increase the DNN model size and the amount of training data/time for better approximation ability to achieve satisfactory feasibility performance.} Overall, our algorithm, start from round $i=0$, contains the following steps.
\begin{mdframed}[backgroundcolor=gray!10, skipabove=6mm, skipbelow=3mm]
\begin{itemize}
    \item Step 1. Prepare the initial training set (denoted as $\mathcal{T}^{0}$) via uniform sampling in the input domain $\mathcal{D}$ and train the DNN using training set $\mathcal{T}^{0}$ (line 1-2).
    \item Step 2. At round $i$, identify the worst-case input $\boldsymbol{\theta}^i$ {within the entire input domain $\mathcal{D}$}. If the obtained optimal objective value $\gamma\leq\Delta$, stop the iteration (line 4-7).
    \item  Step 3. If $\gamma>\Delta$, construct an auxiliary subset $S^i$ containing training pairs $(\boldsymbol{\theta}^{i}_k,\boldsymbol{x}^{i}_k)$ by uniformly sampling centered around $\boldsymbol{\theta}^i$ and the associated calibrated OPCC solutions (line 6-9).
    \item Step 4. Further train the DNN on the new training set $\mathcal{T}^{i+1}$ that combines $\mathcal{S}^i$ and the pre-obtained set $\mathcal{T}^{i}$ (line 11) using back-propagation to minimize the loss function \eqref{equ:loss.function} considering constraints calibrations with the chosen training algorithm, e.g., stochastic gradient descent (SGD) with momentum~\cite{qian1999momentum} (line 10-12).
    \item Step 5. Check whether feasibility in $\mathcal{S}^i$ (constructed around the identified adversarial sample $\boldsymbol{\theta}^i$ at Step 2 with violations) is restored by the post-trained DNN (line 13-17). If so, proceed to the next round $i+1$ and go to Step 2. 
\end{itemize}
\end{mdframed}


We expect that after a few training epochs, the post-trained DNN can restore feasibility at the identified adversarial sample $\boldsymbol{\theta}^i$ and the points around it in $\mathcal{S}^i$. This is inspired by the observation that after adding the previously identified training pairs $\mathcal{S}^i$ into the existing training set, the DNN training loss is dominated by the approximation errors and the penalties at the samples in $\mathcal{S}^i$. {Though the training loss may not be optimized to $0$, e.g., still has violations w.r.t. the calibrated constraints limits, the DNN solution is expected to satisfy the original inequality constraints after such preventive training procedure.} Therefore, the post-trained DNN is capable of preserving feasibility and good accuracy at these input regions. We remark that the algorithm terminates when the maximum relative violation is no greater than the calibration rate, i.e., $\gamma\leq\Delta$ (line 6), such that universal feasibility is guaranteed. Thus, we can construct a DNN model with desirable optimality without sacrificing feasibility by the proposed algorithm, namely DNN Optimality Enhanced as shown in Fig~\ref{fig:framework}. Simulation results in Sec.~\ref{ssec:effect.asaa} show the effectiveness of the propose algorithm.

We highlight the difference between the DNN model obtained in Sec.~\ref{ssec.suff.dnn.size} and that obtained by the \textit{Adversarial Sample-Aware} algorithm as follows. The former is directly constructed via solving \eqref{equ:DNNsize-1}--\eqref{equ:DNNsize-2}, which guarantees universal feasibility whilst without considering optimality. In contrast, the latter is expected to enhance optimality performance while preserving universal feasibility as both optimality and feasibility are considered during the training. We further provide theoretical guarantee of it in ensuring universal feasibility of DNN in the following proposition.
\begin{prop}\label{prop:infinity}
Consider a DNN model with $N_{\text{hid}}$ hidden layers each having $N^*_{\text{neu}}$ neurons. For each iteration $t$, assume such a DNN trained with the \textit{Adversarial Sample-Aware} algorithm can maintain feasibility at the constructed neighborhood $\hat{\mathcal{D}}^j=\{\boldsymbol{\theta }|\boldsymbol{\theta }^j\cdot (1-a)\le \boldsymbol{\theta }\le \boldsymbol{\theta }^j\cdot (1+a),\boldsymbol{\theta }\in \mathcal{D} \}$ around $\boldsymbol{\theta}^j$ with some small constant $a>0$ for $\forall j\leq i$. There exists a constant $C$ such that the algorithm is guaranteed to ensure universal feasibility as the number of iterations is larger than $C$.
\end{prop}

The proof idea is shown in Appendix~\ref{appendix.prop3}. Proposition~\ref{prop:infinity} indicates that, with the iterations is large enough, the \textit{Adversarial Sample-Aware} algorithm can ensure universal feasibility by progressively improving the DNN performance around each region around worst-case input that causes infeasibility. Therefore, the DNN gradually better learns the global mapping information at each iteration benefiting from the ideas of adversarial learning and active learning~\cite{chakraborty2018adversarial,ren2020survey}. {It provides a theoretical understanding of the justifiability of the \textit{ASA} algorithm.} Though the results claim the feasibility guarantee as the number of iterations is large enough, in practice, we can terminate the ASA training algorithm
whenever the maximum solution violation is smaller than the inequality calibration rate, which implies universal feasibility guarantee. We note that the feasibility enforcement in the empirical/heuristic algorithm achieves strong theoretical grounding while its performance can be affected by the training method chosen. Nevertheless, as observed in the case study in Sec.~\ref{sec:simulations} and Appendix~\ref{appen:sec:details.deepopf+}, the proposed \textit{Adversarial Sample-Aware} algorithm terminates in at most 52 iterations with 7\% calibration rate, i.e., $\gamma\leq\Delta$, which indicates the efficiency and usefulness of the proposed training algorithm in practical application. 

Note that at the \textit{Adversarial sample identification} step, the involved program (line 4) is a mixed-integer non-convex problem. The existing solvers, e.g., Gurobi, CPLEX, or APOPT, may not be able to generate the global optimal solution due to the high complexity of the non-convex combinatory problem. In the following, we present that we can still obtain a useful upper bound for the class of OPLC.

\subsubsection{Special Case: OPLC}\label{sssec:opcc.algorithm}
We remark that for the class of OPLC, i.e., $g_j$ are all linear, $\forall j=1,\ldots,m$, the concerned inner maximization problem in~\eqref{equ:DNNsize-1}--\eqref{equ:DNNsize-2} of the \textit{Adversarial sample identification} step in Algorithm~\ref{alg:alg.asaa} (line 4) is the form of MILP. Though the MILP may not be solved to global optimum, we can still use the obtained \textit{upper bound} to verify the performance of the obtained DNN. If the obtain optimal objective (or its upper bound) is no greater than the calibration rate, then universal feasibility of the trained DNN is guaranteed. In addition, despite the difficulty of the mix-integer non-convex programs that we need to solve repeatedly, we can always obtain a feasible (sub-optimal) solution for further use for both general OPCC and OPLC.\footnote{See the footnote in Sec.~\ref{sssec:opcc.size} for a discussion.} Our simulation results is Sec.~\ref{sec:simulations} demonstrate such observation. We highlight the result in the following proposition.


\begin{prop} \label{prop:asaa}
     Consider the OPLC, i.e., $g_j$ are all linear, $\forall j=1,\ldots,m$, and a DNN model with $N_{\text{hid}}$ hidden layers each having $N^*_{\text{neu}}$ neurons. We can obtain $\gamma$ as an upper bound to the optimal objective of the \textit{Adversarial sample identification} problem in Algorithm~\ref{alg:alg.asaa} (line 4) with a time complexity $\mathcal{O}((M+|\mathcal{E}|+2N_{\text{hid}} N^*_{\text{neu}}+4N)^{2.5})$ at each iteration $i$. If $\gamma\leq\Delta$, then the obtained DNN with parameters $(\mathbf{W}^{*}, \bold{b}^{*})$ guarantees universal feasibility such that for any $\theta\in\mathcal{D}$, the solution of this DNN is feasible w.r.t. \eqref{equ:OPCC.formulation-3}--\eqref{equ:OPCC.formulation-4}.
\end{prop}
Proposition~\ref{prop:asaa} indicates $\gamma$ can be obtained in polynomial time. If $\gamma \leq \Delta$, it means the obtained DNN with parameters $(\mathbf{W}^{*}, \bold{b}^{*})$ can guarantee universal feasibility; otherwise, it means we need to further include the adversarial samples and retrain the DNN for better performance. {Our simulation results in Sec.~\ref{sec:simulations} show the effectiveness of the algorithm.}

Overall, the framework can ensure DNN solution feasibility based on the preventive learning approach and is expected to maintain good DNN optimality performance without sacrificing feasibility guarantee via the proposed \textit{Adversarial Sample-Aware} training algorithm. Our simulation results in Sec.~\ref{sec:simulations} show the effectiveness of the framework.

\section{Performance Analysis of the Preventive Learning Framework}
\subsection{Summary of Results under Different Settings and Universal Feasibility Guarantee for OPLC}
In this subsection, we briefly summarize the results that can be obtained in polynomial time under the setting of general OPCC and OPLC ($g_j$ are all linear, $\forall j=1,\ldots,m$) if the corresponding program is not solved to global optimum. We discuss the results at \textbf{Determine calibration rate}, \textbf{Determine DNN size for ensuring universal feasibility}, and \textbf{Adversarial Sample-Aware training algorithm} steps in the proposed framework after i), ii), and iii) in the following respectively.
\begin{itemize}
    \item For general OPCC, we can get i) an upper bound on the maximum calibration rate (with the feasible (sub-optimal) solution); ii) a lower bound (with the feasible (sub-optimal) solution) on the worst-case violation given DNN parameters while we may not get the valid and useful result for the bi-level program \eqref{equ:DNNsize-1}--\eqref{equ:DNNsize-2}; iii) a lower bound on the worst-case violation (with the feasible (sub-optimal) solution) at the \textit{Adversarial sample identification} problem in Algorithm~\ref{alg:alg.asaa} (line 4). In summary, these bounds/results can be loose and may not be utilized with desired performance guarantee.
    \item For OPLC, we can get i) a \textit{valid lower bound} on the maximum calibration rate (may or may not not with the feasible solution) that can preserve the input parameter region; ii) a \textit{valid upper bound} on the worst-case violation given DNN parameters (may or may not not with the feasible solution) and on the global optimum of the bi-level MILP program \eqref{equ:DNNsize-1}--\eqref{equ:DNNsize-2} that can help determine the sufficient DNN size for universal feasibility if it is no greater than the calibration rate; 3) a \textit{valid upper bound} on the worst-case violation (may or may not with the feasible solution) at the \textit{Adversarial sample identification} problem in Algorithm~\ref{alg:alg.asaa} (line 4) that guarantees universal feasibility if it is no greater than the calibration rate. In summary, the bounds obtained from the setting of OPLC can still be useful and valid for further analysis. Therefore, we can construct the DNNs with provable universal solution feasibility guarantee. We provide the following proposition showing that the preventive learning framework generates {two DNN models} with universal feasibility guarantees.
\begin{prop} \label{prop:analysis}
Consider the OPLC, i.e., $g_j$ are all linear, $\forall j=1,\ldots,m$. Let $\Delta$, $\rho$, and $\gamma$ be the determined maximum calibration rate, the obtained objective value via solving \eqref{equ:DNNsize-1}--\eqref{equ:DNNsize-2} to determine the sufficient DNN size, and the obtained maximum relative violation of the trained DNN from \textit{Adversarial Sample-Aware} algorithm following steps in preventive learning framework, respectively. Assume (i) $\Delta > 0$, (ii) $\rho \leq \Delta$, and (iii) $\gamma\leq\Delta$. The DNN-FG obtained from determining sufficient DNN size can provably guarantee universal feasibility and the DNN from ASA algorithm further improves optimality without sacrificing feasibility guarantee $\forall\boldsymbol{\theta}\in\mathcal{D}$. 
\end{prop}
Proposition~\ref{prop:analysis} indicates the DNN models obtained by preventive learning framework is expected to guarantee the universal solution feasibility, which is verified by the simulation results in Sec.~\ref{sec:simulations}. 
\end{itemize}

Furthermore, we state that for each involved program, we can always obtain a feasible (sub-optimal) solution for further use. These results are discussed in the corresponding parts in the paper. {Overall, the preventive learning framework provides two DNNs. The first one constructed from the step of determining the sufficient DNN size in Sec.~\ref{ssec:feasibilityDNN} can guarantee universal feasibility (DNN-FG), while the other DNN obtained from the proposed \textit{Adversarial-Sample Aware} training algorithm in Sec.~\ref{ssec:ASAA} further improves optimality and maintains DNN's universal feasibility simultaneously (DNN Optimality Enhanced). Such feasibility guarantee can be verified from the obtained valid bounds specific for OPLC.}

\subsection{Run-time Complexity of the Framework}
\label{ssec:complexity}
To better understand the advantage of the proposed framework for solving OPCC, we further analyze its computational complexity as follows.

The computational complexity of the framework consists of the complexity of using DNN to predict the solutions, which is $\mathcal{O} \left(N_{{\text{hid}}}N_{{\text{neu}}}^2\right)$~\cite{deepopf2}. {Recall that $N_{{\text{hid}}}$ denote the number of hidden layers in DNN (depth), and $N_{{\text{neu}}}$ denotes the number of neurons at each layer (width).} In practice, we set $N_{{\text{hid}}}$ to be 3 and observe that the DNN with width $N_{{\text{neu}}}$ of $\mathcal{O} \left(N\right)$ can achieve satisfactory optimality performance with universal feasibility guarantee. Therefore, the complexity of using DNN to predict the $N$ variables is $\mathcal{O} \left(N^2\right)$.
    

We then provide the complexity of the traditional method in solving the optimization problems with convex constraints. To the best of our knowledge, OPCC in its most general form is NP-hard cannot be solved in polynomial tie unless P$=$NP. To better deliver the results here, we consider the specific case of OPCC, namely the multiparameter quadratic program (mp-QP), with linear constraints and quadratic objective function, formulated as \eqref{appen:equ.mpqp-obj}--\eqref{appen:box} for an analysis.
 The mp-QP is wildly adopted with many applications, e.g., DC-OPF problems in power systems and model-predictive control (MPC) problems in general control systems. See Appendix~\ref{appen:mpqp-f} for the formulation of mp-QP. 
We remark that the complexity of solving mp-QP provides a lower bower for the general OPCC problem.

Note that the number of decision variables to be optimized is $N$ in the formulated mp-QP.  After taking $\mathcal{O}\left(|\mathcal{E}|M \right)$ operations to calculate the value of $\boldsymbol{b^T_j}\boldsymbol{\theta}$ in $g_j(\boldsymbol{x},\boldsymbol{\theta})\triangleq \boldsymbol{a^T_j}\boldsymbol{x}+\boldsymbol{b^T_j}\boldsymbol{\theta}\leq e_j$ for each $j\in\mathcal{E}$, the best-known interior-point based iterative algorithm~\cite{ye1989extension} requires a computational complexity of $\mathcal{O} \left( N^4\right)$ for solving such programs, measured by the number of elementary operations assuming that it takes a fixed time to execute each operation.  Therefore, the traditional iterative method for solving mp-QP has a computational complexity of $\mathcal{O}\left(N^4+|\mathcal{E}|M\right)$.


We remark that the computational complexity of the proposed framework is lower than that of traditional algorithms. Our simulation results in Sec.~\ref{sec:simulations} on DC-OPF problems verify such observation. As seen, the proposed framework provides close-to-optimal solutions ($<0.19\%$ optimality loss) in a fraction of the time compared with the state-of-the-art solver (up to two order of magnitude speedup).


\rev{
\subsection{Trade-off between Feasibility Guarantee and Optimality}\label{ssec.tradeoff}
We remark that to guarantee universal feasibility, the preventive learning framework shrinks the feasible region used in preparing training data. Therefore, the learned solution may have larger optimality loss due to the (sub)-optimal training data. It indicates a trade-off between optimality and feasibility, i.e., larger calibration rate leads to better feasibility but worse optimality. To further enhance DNN optimality performance, one can choose a smaller calibration rate than $\Delta$ while enlarging DNN size for better approximation ability and hence achieve satisfactory optimality performance while guarantee universal feasibility simultaneously. 
}


\section{Application to Solve DC-OPF Problems and Numerical Experiments}\label{sec:simulations}

DC-OPF is a fundamental problem for modern grid operation. It aims to determine the least-cost generator dispatch to meet the load in a power network subject to physical and operational constraints.\footnote{Despite having the most accurate description of the power system, the OPF problem with a full AC power flow formulation (AC-OPF) is a non-convex problem, whose complexity obscures its practicability. Meanwhile, based on linearized power flows, the DC-OPF problem is a convex problem 
and is widely adopted in a variety of applications, including electricity market clearing and power transmission network management. See e.g., \cite{frank2012optimal1,frank2012optimal2} for a survey.} With the penetration of renewables and flexible load, the system operators need to handle significant uncertainty in load input during daily operation. They need to solve DC-OPF problem under
many scenarios more frequently and quickly in a short interval, e.g., 1000 scenarios in 5 minutes, to obtain a stochastically optimized solution for stable and economical operations. However, iterative solvers may fail to solve a large number of DC-OPF problems for large-scale power networks fast enough for the purpose. Although recent DNN-based approaches obtain close-to-optimal solution much faster than conventional methods, they do not guarantee solution feasibility. We here design \text{DeepOPF+} by employing the preventive learning framework to tackle this issue.

\subsection{Problem Formulation}
The DC-OPF problem determines optimal generator operations that achieve the least cost while satisfying the physical and operational constraints for each load input $\boldsymbol{P_{D}}\in\mathcal{R}^{|\mathcal{B}|}$:
\begin{align}
    \min_{\boldsymbol{P_{G}},\ {\Phi }} &  \mathrm{\ }\sum_{i\in\mathcal{G}}{c_i\left( P_{Gi} \right)} \\
    \mathrm{s.t.} \ \ &\boldsymbol{P_{G}^{\min}}\le \boldsymbol{P_{G}}\le \boldsymbol{P_{G}^{\max}},\label{equ:pg}\\
    & \mathbf{M}\cdot \Phi =\boldsymbol{P_{G}}-\boldsymbol{P_{D}}, \label{equ:branch.balance}\\
    & -\boldsymbol{P^{\max}_{\text{line}}} \leq\mathbf{B}_{\text{line}}\cdot \Phi \leq \boldsymbol{P^{\max}_{\text{line}}}, \label{equ:branch.limit}
\end{align}
where $\mathcal{B}$ and $\mathcal{G}$ denote the set of buses and generators respectively. $\boldsymbol{P_{G}^{\min}}\in\mathcal{R}^{|\mathcal{B}|}$ (resp. $\boldsymbol{P_{G}^{\max}}$) and $\boldsymbol{P^{\max}_{\text{line}}}\in\mathcal{R}^{|\mathcal{K}|}$ denote the minimum (resp. maximum) generation output limits of the generators\footnote{{$P_{G_i}=P^{\min}_{G_i}=P^{\max}_{G_i}=0, \forall i \notin \mathcal{G}$, and $P_{D_i}=0, \forall i \notin \mathcal{A}$, where $\mathcal{A}$ denotes the set of load buses.}} and the line transmission capacity limits of the branches in the power network, where $\mathcal{K}$ is the set of transmission lines. $\mathbf{M}\in\mathcal{R}^{|\mathcal{B}|\times|\mathcal{B}|}$, $\mathbf{B}_{\text{line}}\in\mathcal{R}^{|\mathcal{K}|\times|\mathcal{B}|}$, and $\Phi\in\mathcal{R}^{|\mathcal{B}|}$ denote the bus admittance matrix, line admittance matrix, and bus phase angles respectively.\footnote{{Here we only consider the branches where they could reach the limits.}} The objective is the total generation cost and $c_i\left(\cdot \right)$ is the cost function of each generator, which is usually \rev{strictly quadratic~\cite{260897,tpcwTrey4} from generator's heat rate curve.} Constraints \eqref{equ:pg}--\eqref{equ:branch.limit} enforce nodal power balance equations and the limits on the active power generation $P_G$ and line transmission capacity. Note that the slack bus voltage phase angle $\phi^{\text{slack}}$ is fixed to be zero. The DC-OPF problem is hence a quadratic programming and admits a unique optimal solution w.r.t. each load input $\boldsymbol{P_{D}}$. Analogy to OPCC (\ref{equ:OPCC.formulation-1})-(\ref{equ:OPCC.formulation-4}), $\sum_{i\in\mathcal{G}}{c_i\left( P_{Gi} \right)}$
is the objective function $f$ in (\ref{equ:OPCC.formulation-1}). $\boldsymbol{P_{D}}$ is the problem input $\mathbf{\theta}$ and $(\boldsymbol{P_{G}},\Phi)$ are the decision variables $\boldsymbol{x}$.

\subsubsection{Proposed \textsf{DeepOPF+}}

We apply the proposed preventive-learning framework to design a DNN scheme, named \textsf{DeepOPF+}, for solving DC-OPF problems. We refer interested readers to Appendix~\ref{appendix.deepopf+} for details. We first remove the non-critical inequality constraints for DC-OPF problem as discussed in Appendix~\ref{appen:ssec.remove}. We then determine the inequality constraints calibration rate as discussed in Sec.~\ref{ssec:calibrationrange}. Following the steps in Sec.~\ref{ssec:feasibilityDNN} and Sec.~\ref{ssec:ASAA}, we obtain the sufficient DNN size that can guarantee solution universal feasibility and apply the \textit{Adversarial Sample-Aware} algorithm to train the DNN with such size for stronger optimality performance.  Suppose $\Delta$, $\rho$, and $\gamma$ denote the determined maximum calibration rate, the obtained objective value via solving \eqref{equ:DNNsize-1}--\eqref{equ:DNNsize-2} using the determined sufficient DNN size, and the maximum relative violation of the trained DNN from \textit{Adversarial Sample-Aware} algorithm in the design of \textsf{DeepOPF+}, respectively.
We highlight the feasibility guarantee and computational efficiency of \textsf{DeepOPF+} in following corollary.

\begin{corollary}\label{cor1}
For the DC-OPF problem and DNN model defined in \eqref{dnn.model}. Assume (i) $\Delta > 0$, (ii) $\rho \leq \Delta$, and (iii) $\gamma\leq\Delta$, then the \textsf{DeepOPF+} generates a DNN  guarantees universal feasibility for any $\boldsymbol{P_{D}}\in\mathcal{D}$. 
Furthermore, suppose the DNN width is the same order of number of bus, $B$, the proposed DNN based framework \textsf{DeepOPF+} has a smaller computational complexity of $\mathcal{O}\left(B^2\right)$ compared with that of traditional method $\mathcal{O}\left(B^4\right)$, where $B$ is the number of buses.
\end{corollary}
Corollary~\ref{cor1} says that \textsf{DeepOPF+} can solve DC-OPF problems with universal feasibility guarantee with lower computational complexity,\footnote{\rev{We remark that the training of DNN is conducted offline; thus, its complexity is minor as amortized over many DC-OPF instances, e.g., 1000 scenarios per 5 mins.  Meanwhile, the extra cost to solve the new-introduced optimizations in our design is also minor observing that existing solvers like Gurobi could solve the problems efficiently, e.g., $<$20 minutes to solve the MILPs to determine calibration rate and DNN size. Thus, we consider the run-time complexity of the DNN-based scheme, which is widely used in existing studies.}} as compared to conventional iterative solvers {\rev{as DNNs with width $\mathcal{O}(B)$ can achieve desirable feasibility/optimality. Such an assumption is validated in existing literature \cite{deepopf1} and our simulation.}}. To our best knowledge, \textsf{DeepOPF+} is the first DNN scheme for solving DC-OPF problems that guarantees solution feasibility without post-processing. In the following subsections, we further apply the steps in the proposed \textsf{DeepOPF+} framework to the DC-OPF problems. \rev{We remark that the design of \textsf{DeepOPF+} can be easily generalized to other linearized OPF models~\cite{cain2012history,yang2018general,bolognani2015fast} with DNN solution feasibility guarantee.}

\begin{table}[!t]
	\centering
	\caption{Parameters for test cases.} 
	\renewcommand{\arraystretch}{1}
	\begin{threeparttable}
		\begin{tabular}{c|c|c|c|c|c|c}
			\toprule
			\hline
			Case & \tabincell{c}{$|\mathcal{B}|$} & \tabincell{c}{$|\mathcal{G}|$} & \tabincell{c}{$|\mathcal{A}|$} &\tabincell{c}{$|\mathcal{K}|$} &\tabincell{c}{$N_{{\text{hid}}}$} &\tabincell{c}{Neurons \\per hidden layer}
			\\
			\hline
			\tabincell{c}{Case30} & 30 & 6 & 20 &41 & 3&32/16/8\\
			\hline
			\tabincell{c}{Case118} & 118 & 19 & 99 &186&3&128/64/32\\
			\hline
			\tabincell{c}{Case300} & 300 & 69 & 199 &411&3&256/128/64\\
			\hline
			\bottomrule
		\end{tabular}
		\begin{tablenotes}
			\footnotesize
			\item[*] $\mathcal{A}$ and $\mathcal{K}$ denote the set of load bus and the set of branches respectively.
			\item[*] The number of load buses is calculated based on the default load on each bus. A bus is considered a load bus if its default active power consumption is non-zero.
		\end{tablenotes}
	\end{threeparttable}
	\label{tab.testcase.DNN.1}
\end{table}

\subsection{Experiment Setup}\label{ssec:simulation.setup}
The experiments are conducted in CentOS 7.6 on the quad-core (i7-3770@3.40G Hz) CPU workstation with 16GB RAM. {We evaluate \textsf{DeepOPF+} on three representative test power networks: IEEE 30-, 118-, and 300-bus test cases~\cite{tpcwTrey4}.}\footnote{As IEEE 118-bus and 300-bus test cases provided by MATPOWER~\cite{zimmerman2011matpower} do not specify the line capacities, we use IEEE 118-bus test case and the line capacity setting for IEEE 300-bus test case with the same branch provided by Power Grid Lib~\cite{babaeinejadsarookolaee2019power} 
(version 19.05).} For each test case, the amount of training data and test data are 50,000 and 10,000, training data and test data contain 50,000 and 10,000 instances, of which the load input obtained via uniformly sampling strategy and the solution are generated by a conventional solver.
The load data is sampled within $[100\%, 130\%]$ of the default load on each load bus uniformly at random, which covers both light-load ($[100\%, 115\%]$) and heavy-load ($[115\%, 130\%]$) regimes such that some transmission lines and slack generation reach the allowed operational limits. {Note that the existing DNN-based approaches may not be able to provide feasible solution especially under the heavy-load regimes.} {According to the sizes of the power network, we design DNNs with different neurons on each hidden layer. The parameters are given in Table~\ref{tab.testcase.DNN.1}.  When implementing the \textit{Adversarial Sample-Aware} algorithm in DNN training, we apply the widely-used stochastic gradient descent (SGD) with momentum on the Pytorch platform. The number of training epochs is 200. Based on empirical experience, we set the weighting factors $w_1=1$ and $w_2=1$ in the loss function.
 We evaluate the obtained DNNs from different training approaches considering the following metrics:
\begin{itemize}
    \item Feasibility: What is percentage of the feasible solutions provided by DNN on the test set?
    \item Universal Feasibility: Whether the obtained DNN can guarantee universal feasibility within the entire load input domain?
    \item Prediction error: What is the average relative optimality difference between the objective values obtained by DNN and the ground truth provided by Pypower?
    \item Optimality loss: What is the average relative optimality difference between the objective values obtained by DNN and Pypower?
    \item Speedup: How fast can the DNN achieve in obtaining the final feasible solution? That is, what is the average speedup, i.e., the average running-time ratios of Pypower to DNN-based approach for the test instances, respectively? 
    \item Worst-case violation rate: What is the largest constraints violation rate of DNN solutions in the entire load domain?
\end{itemize}
}

\subsection{Summary of the \textsf{DeepOPF+} Design}

\begin{figure} [!t]
  \centering
  \subfigure[Case30.]{
    \label{fig.case30danskin} 
    \includegraphics[width = 0.3\textwidth]{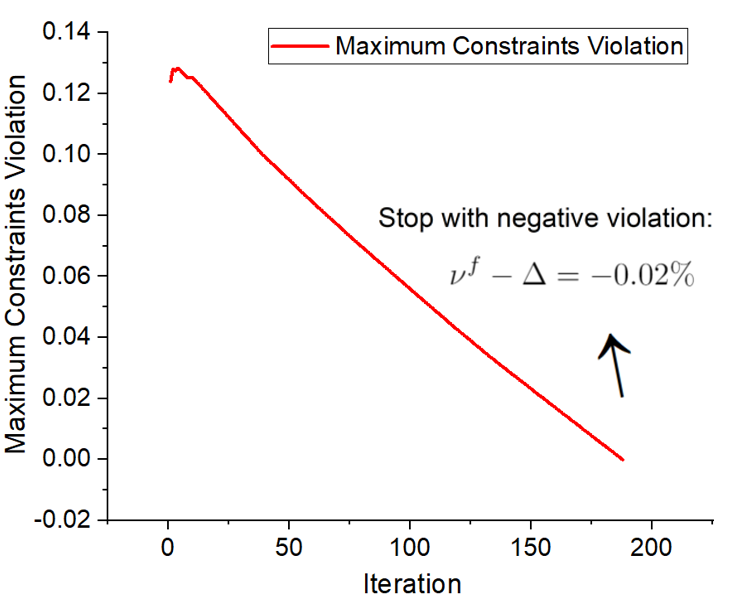}
  }
  \subfigure[Case118.]{
    \label{fig.case118danskin} 
    \includegraphics[width = 0.3\textwidth]{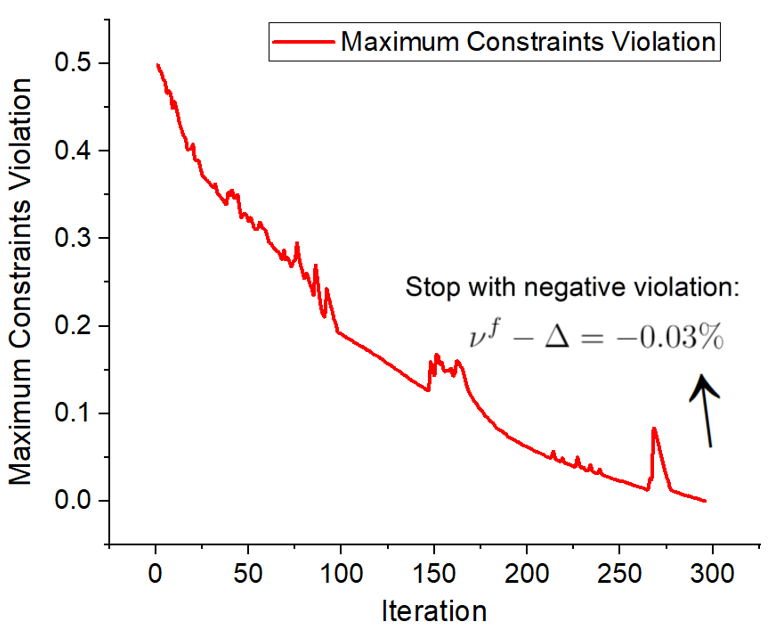}
  }
  \subfigure[Case300.]{
    \label{fig.case300danskin} 
    \includegraphics[width = 0.3\textwidth]{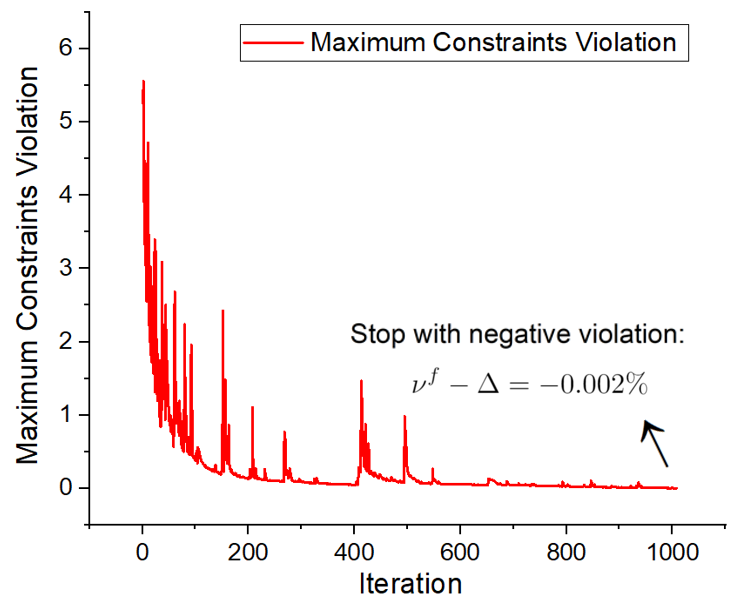}
  }
  \caption{Maximum relative constraints violation compared with calibration rate ($\nu^f-\Delta$) at each iteration for IEEE Case30, Case118, and Case300 test case.}\label{fig.danskin}
\end{figure}

We summarize the detailed design of \textsf{DeepOPF+} to solve the DC-OPF problems following the three steps discussed in Sec.~\ref{ssec:calibrationrange} to Sec.~\ref{ssec:ASAA} in the following. First, the maximum calibration rates for three IEEE test cases are shown in Table~\ref{tab:violaton.calibration}, representing the room for DNN prediction error that the critical inequality constraints can be calibrated by at most $7.0\%$, $16.7\%$, and $21.6\%$ for IEEE 30-, 118-, and 300-bus test cases respectively by solving the programs \eqref{equ:calibration.rate-1}--\eqref{equ:calibration.rate-2} for each test case. {Note that as DC-OPF problem is a convex problem with linear constraints, the program \eqref{equ:calibration.rate-1}--\eqref{equ:calibration.rate-2} involved in this step can be reformulated as MILP as discussed in Sec.~\ref{sssec:milp.calibrate}. In this experiment, we note that the off-the-shell solver returns exact solutions for the problem in \eqref{equ:calibration.rate-1}--\eqref{equ:calibration.rate-2} with zero optimality gap.} Therefore, the obtained calibration rate is tight and global optimal. Second, we directly construct DNNs to ensure universal feasibility for the three IEEE test cases, which all have 3 hidden layers but with 32/16/8 neurons, 128/64/32 neurons, and 256/128/64 neurons, respectively. We also show the change of the difference between maximum relative constraints violation and calibration rate during solving process in Fig.~\ref{fig.danskin}. From Fig.~\ref{fig.danskin}, we observe that for all three test cases, the proposed approach succeeds in reaching a {relative constraints} violation no larger than the corresponding calibration rate $\Delta$, i.e., $\rho\leq\Delta$, indicating that the DNNs in Table~\ref{tab.testcase.DNN.1} have enough size to guarantee feasibility within the given load input domain. Third, we evaluate the performance of the DNN model obtained following the steps in Sec.~\ref{ssec.suff.dnn.size} without using the \textit{Adversarial Sample-Aware} algorithm. While ensuring universal feasibility, it suffers from an undesirable optimality loss, up to $2.42\%$ and more than $130\%$ prediction error. In contrast, the DNN model trained with the \textit{Adversarial Sample-Aware} algorithm achieves lower optimality loss (up to $0.19\%$) while preserving universal feasibility. The observation justifies the effectiveness of \textit{Adversarial Sample-Aware} algorithm. The detailed results are discussed in the following subsections.

\begin{table}[!t]
\caption{Maximum calibration rate on the critical constraints for three test cases.}
	\renewcommand{\arraystretch}{1.1}
	\centering
	\begin{tabular}{c|c|c|c}
		\toprule
		\hline
        \multicolumn{1}{c|}{Variants} & \tabincell{c}{IEEE Case30} & \tabincell{c}{IEEE Case118} & IEEE Case300\\
        \hline
        {\tabincell{c}{Maximum calibration rate}} &\tabincell{c}{7.0\%} &16.7\% &21.6\% \\
        \cline{2-4}
		\hline
		\bottomrule		
		\end{tabular}
	\label{tab:violaton.calibration}
\end{table}

\subsection{Performance Evaluation}

\subsubsection{Effectiveness of Inequality Constraint Calibration and Drawback of Traditional Training}\label{ssec:evaluation}
To better deliver the advantage of \textsf{DeepOPF+}, we first evaluate the effectiveness of constraints calibration on improving the feasibility performance on the test set and discuss the drawback of traditional training without considering adversarial inputs. Specifically, we adopt the \textsf{DeepOPF} approach in~\cite{deepopf1} and compare the performance of its variants that are with different calibration rate on the critical constraints (named \textsf{DeepOPF-Cal}). According to the maximum constraints calibration results shown in Table~\ref{tab:violaton.calibration}, the test calibration rates are set as $1.0\%$, $3.0\%$, $5.0\%$, and $7.0\%$ in \eqref{equ:inequality.calibration}, respectively, {which are all no greater than the maximum calibration rate shown in Table~\ref{tab:violaton.calibration}}. The DC-OPF problem solution provided by Pypower~\cite{tpcwTrey1} is regarded as ground truth. For each power network, we train a DNN on the uniformly sampled training set derived in Sec.~\ref{ssec:simulation.setup} with loss function \eqref{equ:loss.function} to approximate its load-generation mapping. The DNN inputs the load profile and outputs the generation prediction. 

 \begin{table}[!t]
	\caption{Performance evaluation of DNN-based approach with different constraints calibration under typical training manner.}
	\renewcommand{\arraystretch}{1}
	\centering
	\begin{threeparttable}
 		\scalebox{0.9}{
		\begin{tabular}{c|c|c|c|c|c|c|c|c|c}
			\toprule
			\hline
			\multirow{2}{*}{Case}&
			\multirow{2}{*}{\tabincell{c}{Limit \\calibration (\%)}} &
			\multirow{2}{*}{\tabincell{c}{Feasibility \\rate (\%)}} &
			\multirow{2}{*}{\tabincell{c}{Feasibility rate \\without calibration (\%)}} &
			\multicolumn{3}{c|}{\tabincell{c}{Average cost (\$)}} &
			\multicolumn{2}{c|}{\tabincell{c}{Average running \\time (ms)}} &
			\multirow{2}{*}{\tabincell{c}{Average \\speedup}} \\
			\cline{5-9}
			&&&&\tabincell{c}{\textsf{$\text{\textsf{DeepOPF-Cal}}^*$}}&Ref.&Loss(\%)&\textsf{DeepOPF-Cal}&Ref.&\\ 
			\hline
 			\multirow{4}{*}{Case30}&
 			\tabincell{c}{1.0}& 98.52 &\multirow{4}{*}{97.65}& 675.4 &\multicolumn{1}{|c|}{\multirow{4}{*}{675.2}} & 0.03 &0.51 & \multicolumn{1}{|c|}{\multirow{4}{*}{43}} & $\times$85\\
 			
  			 &\tabincell{c}{3.0}& 99.73 & & 675.4 &  & 0.03 &0.50 && $\times$86 \\
  			 &\tabincell{c}{5.0}& 99.99 & & 675.4 &  & 0.03 &0.50 && $\times$86 \\
  			 &\tabincell{c}{7.0}& 100 & & 675.5 &  & 0.04 &0.50 && $\times$86 \\
 			\hline
 			\multirow{4}{*}{Case118}&
 			\tabincell{c}{1.0}& 80.87 &\multirow{4}{*}{54.34}& 111377.8 &	\multicolumn{1}{|c|}{\multirow{4}{*}{111165.3}} & 0.19 &1.27 & \multicolumn{1}{|c|}{\multirow{4}{*}{123}} & $\times$164\\
  			 &\tabincell{c}{3.0}& 98.60 & & 111472.8 &  & 0.28 &0.71 && $\times$185 \\
  			 &\tabincell{c}{5.0}& 99.94 & & 111606.4 &  & 0.40 &0.58 && $\times$213 \\
  			 &\tabincell{c}{7.0}& 100 & & 111724.9 &  & 0.50 &0.58 && $\times$214 \\
 			\hline
 			\multirow{4}{*}{Case300}&
 		\tabincell{c}{1.0}& 94.06 &\multirow{4}{*}{87.73}& 851247.5 &	\multicolumn{1}{|c|}{\multirow{4}{*}{850882.6}} & 0.04 &1.34& \multicolumn{1}{|c|}{\multirow{4}{*}{84}} & $\times$127\\
 		
  			 &\tabincell{c}{3.0}& 98.61 & & 851401.3 &  & 0.06 &0.79 && $\times$134 \\
  			 &\tabincell{c}{5.0}& 99.73 & & 851695.8 &  & 0.10 &0.65 && $\times$136 \\
  			 &\tabincell{c}{7.0}& 100 & & 852099.6 &  & 0.14 &0.60 && $\times$140 \\
 			\hline
			\bottomrule
		\end{tabular}}
		\begin{tablenotes}
			\footnotesize
 			\item[*] {\textsf{DeepOPF-Cal} stands for the adopted \textsf{DeepOPF} approach with critical constraints calibration.}
		\end{tablenotes}
		
	\end{threeparttable}
	\label{tab:DeepOPF+}
\end{table}

For the DNNs from such typical training approach, we observe infeasibility without any calibrations, and feasibility improvement is up to $45.66\%$ on the test set with the preventive calibrations (from $54.34\%$ to $100\%$).\footnote{If the DNN generates infeasible solutions, we apply an efficient $\ell_1$-projection post-processing procedure to ensure the feasibility of the final solution~\cite{deepopf2}, which is essentially an LP. The average running time includes the post-processing time if DNN obtains infeasible solutions.} 
Also, the differences between the average cost of the DNN solutions and that of the reference ones are minor (at most $0.50\%$). The DNNs speed up the computational time by two orders of magnitude on the test set.\footnote{Note that Case118 takes a longer computational time to obtain the optimal solution with the conventional solver compared to Case300. This observation comes from the observation that Case118 requires more iteration steps to converge (on average 25 times) than Case300 (on average 11 times), while the average running time per iteration of Case118 (4.7 ms) is less than that of Case300 (7.5 ms).}
Note that the existing DNN-based schemes may not achieve high speedups due to the expensive post-processing procedure to recover the feasibility of infeasible DNN solutions. 
Moreover, we observe that larger calibration rates contribute to higher feasibility rates but lead to larger optimality losses.\footnote{As we employ DNN to approximate the load-solution mapping of calibrated OPF problems, the optimality loss may increase when calibration rate is larger, i.e., the sample ground truth deviates more from the optimal solution.} 
Our results demonstrate the effectiveness of critical constraints calibrations and show that all three test cases achieve $100\%$ feasibility with a $7.0\%$ calibration rate on the test set. However, we remark that they cannot guarantee the universal feasibility within the entire input region as their performances under adversarial load inputs could be discouraging, as shown in the next part.

\begin{figure} [!t]
  \centering
  \subfigure[Case118.]{
    \label{fig.case118asaa} 
    \includegraphics[width = 0.45\textwidth]{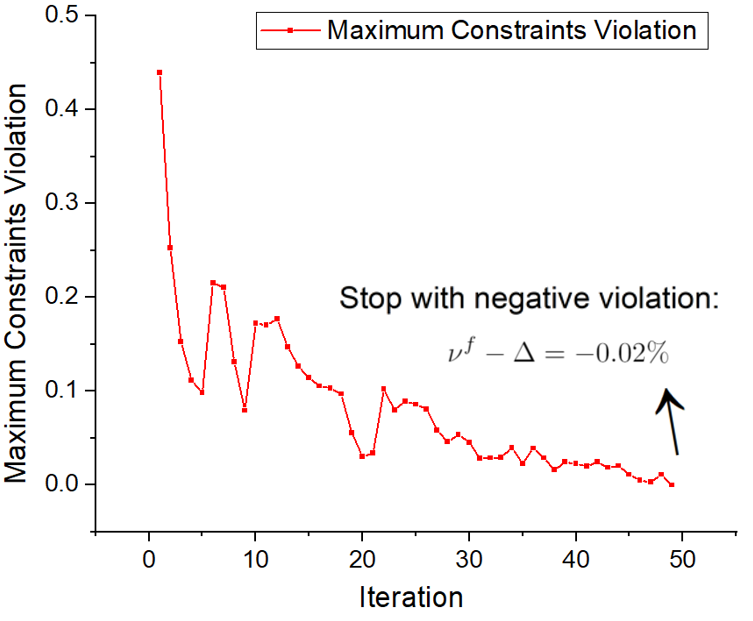}
  }
  \subfigure[Case300.]{
    \label{fig.case300asaa} 
    \includegraphics[width = 0.45\textwidth]{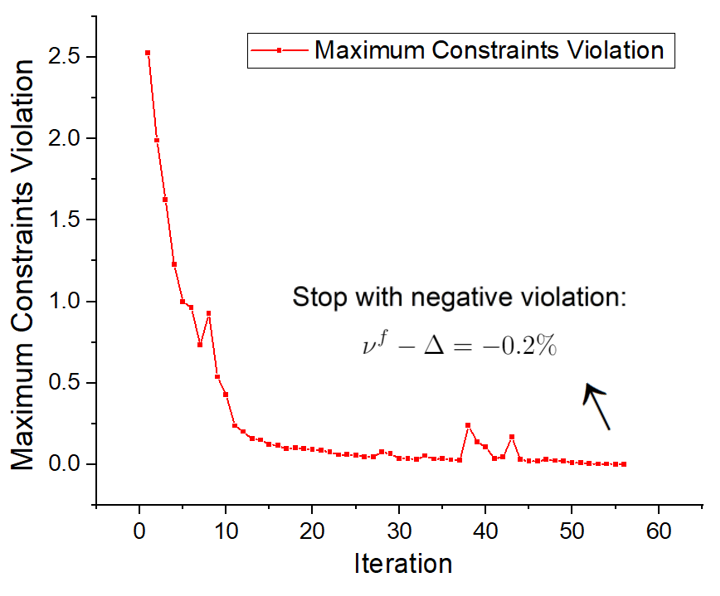}
  }
  \caption{Iteration of \textit{Adversarial Sample-Aware} algorithm for IEEE Case118 and IEEE Case300.}\label{fig.asaa}
\end{figure}

\subsubsection{Performance Comparisons Between \textsf{DeepOPF+} and Existing DNN Scheme}\label{ssec:effect.asaa}
We further evaluate the performance of the proposed \textsf{DeepOPF+}. Specifically, we compare the following three DNN-based approaches:
\begin{itemize}
    \item \textsf{DeepOPF}: the adopted \textsf{DeepOPF} approach~\cite{deepopf1} without constraints calibration.
    
    \item \textsf{DeepOPF-Cal}: the adopted \textsf{DeepOPF} approach~\cite{deepopf1} with $7.0\%$ calibration rate.
    \item \textsf{DNN-FG}: the proposed approach \eqref{equ:DNNsize-1}--\eqref{equ:DNNsize-2} to determine the sufficient DNN size in guaranteeing universal feasibility considering DNN violation minimization.
    \item \textsf{DeepOPF+}: the proposed \textsf{DeepOPF+} approach using the \textit{Adversarial Sample-Aware} training algorithm. Specifically, the initial model is chosen as the DNN model with $7.0\%$ calibration rate. Maximum training epochs $T=200$. Initial training data set $\mathcal{T}^0$ is the same as the one obtained in Sec.~\ref{ssec:simulation.setup}, the auxiliary set $\mathcal{S}^i$ size $K=100$, and the sampling rate $a=0.01$ around the adversarial input $\boldsymbol{P^i_{D}}$.
\end{itemize}
\begin{table*}[!t]
	\centering
	\caption{Performance comparisons of \textsf{DeepOPF+} and existing DNN-based approaches on test cases.}
	\begin{threeparttable}
		\scalebox{0.9}{
			\begin{tabular}{c|c|c|c|c|c|c|c|c|c|c}
			\toprule
			\hline
			\multirow{3}{*}{Variants}&
			\multirow{3}{*}{\tabincell{c}{Universal \\Feasibility}}&
			\multicolumn{3}{c|}{\tabincell{c}{Case30}} &
			\multicolumn{3}{c|}{\tabincell{c}{Case118}} &
			\multicolumn{3}{c}{\tabincell{c}{Case300}}  \cr
			\cline{3-11}
			& &\tabincell{c}{Prediction\\ error (\%)} &\tabincell{c}{Optimality\\loss (\%)} 	& Speedup&\tabincell{c}{Prediction\\error (\%)} &\tabincell{c}{Optimality \\ loss (\%)}& Speedup&\tabincell{c}{Prediction\\error (\%)} &\tabincell{c}{Optimality\\loss (\%)}& Speedup\\
			\hline
			\tabincell{c}{\textsf{DeepOPF-Cal}}&\tabincell{c}{\XSolidBrush}&3.4&0.04 &86&9.9&0.50 &214&2.9&0.14 &140\\
			\hline
			\tabincell{c}{DNN-FG}&\tabincell{c}{\checkmark}&4.5&0.08 &86&39.0&2.42 &220&14.6&0.98 &138\\
			\hline
			\tabincell{c}{DeepOPF+}&\tabincell{c}{\checkmark}&3.5&0.03 &87&9.3&0.43 &220&0.98&0.07 &139\\
			\hline
			\bottomrule
			\end{tabular}}
	\end{threeparttable}
	\label{tab:DeepOPF+.variants}
\end{table*}

\paragraph{Universal Feasibility Guarantee and Desirable Optimality of \textsf{DeepOPF+} on the Entire Load Domain}
We have following observations from Table~\ref{tab:DeepOPF+.variants} for the performance of \textsf{DeepOPF+} on the entire load region [$115\%, 130\%$]. For \textsf{DeepOPF-Cal}, though it performs well on the test set, it fails to provide universal feasibility in satisfying the operational constraints under the worst-case input, i.e., up to $0.71\%$, $43.96\%$, and $252.88\%$ violation for Case30, Case118, and Case300, respectively. Such adversarial inputs are always element-wise at the boundary of $\mathcal{D}$, which can be rarely spotted by the applied uniform sampling strategy. As compared with it, both \textsf{DNN-FG} and \textsf{DeepOPF+} can obtain universal feasibility.

For \textsf{DNN-FG}, though desirable universal feasibility is maintained, its optimality and prediction accuracy performances are substandard, due to only focusing on diminishing violations when optimizing the DNN parameters via \eqref{equ:DNNsize-1}--\eqref{equ:DNNsize-2}. We remark that the performance of \textsf{DNN-FG} is closely related to the initialization of the DNNs. In this experiment, we initialize with the pre-trained DNNs in~\cite{zhao2020deepopf}. Here \textsf{DNN-FG} achieves universal violation within the entire load domain after 187, 295, and 1008 iterations for IEEE Case30, Case118, and Case300, respectively. We refer to Fig.~\ref{fig.danskin} for the relative violation ($\nu^f-\Delta$) for each test case at each iteration for illustration.

For \textsf{DeepOPF+}, we train the DNN at $7.0\%$ calibration
rate on set $\mathcal{T}^{i+1}$ at each round. We observe that the post-trained DNN with parameters $(\mathbf{W}^{i+1}, \bold{b}^{i+1})$ can always restore feasibility at the adversarial input $P^i_D$ and $\mathcal{S}^i$ after a few training epochs (maximum 130 epochs while 24 epochs on average for Case300) and benefit from
including the adversarial load samples in training for smaller worst-case violations.\footnote{For \textsf{DeepOPF-Cal}, its training loss converges after 200 training epochs and hardly decreases if we keep increasing the number of training epochs.}
We note that the DNN model \textsf{DeepOPF+} obtained from the \textit{Adversarial Sample-Aware} algorithm can preserve desirable accuracy and optimality performance ($0.03\%$ loss for Case30, $0.43\%$ loss for Case118 and $0.07\%$ for Case300) while maintaining universal feasibility within the entire load domain. This observation indicates the effectiveness of the \textit{Adversarial Sample-Aware} algorithm. We refer to Fig.~\ref{fig.asaa} for the relative violation ($\nu^f-\Delta$) on Case118 and Case300-bus at each iteration for illustration. For Case30, the \textit{Adversarial Sample-Aware} algorithm helps achieve universal feasibility just after one iteration (from $0.71\%$ to $-0.62\%$).

\paragraph{Performance Improvement of \textsf{DeepOPF+} under Light-Load and Heavy-Load Settings}

We further evaluate the performance of \textsf{DeepOPF+} over IEEE 30-/118-/300- bus test cases~\cite{tpcwTrey4} on the input load region of $[100\%, 130\%]$ of the default load covering both the light-load ($[100\%, 115\%]$) and heavy-load ($[115\%, 130\%]$) regimes, respectively. In particular, the load input region of $[100\%, 115\%]$ of the default load is considered the light-load regime, where only a small portion of test instance constraints are binding in the test data set (e.g., $6.46\%$ test instances for Case300). The load input region of $[115\%, 130\%]$ of the default load is considered the heavy-load regime, where the system's constraints are highly binding (e.g., $100\%$ test instances in the test set have binding inequality constraints for Case300). \rev{We follow the proposed steps and design two different DNNs from \textsf{DeepOPF+} for each test case in the light-load regime and heavy-load regime separately.} We compare \textsf{DeepOPF+} with five baselines {on the same training/test setting}: (i) Pypower: the conventional {iterative} OPF solver; (ii) \textsf{DNN-P}: A DNN scheme adapted from~\cite{deepopf1}. It learns the load-solution mapping using penalty approach without constraints calibration {and incorporates a projection post-processing if the DNN solution is infeasible; (iii) \textsf{DNN-D}: A penalty-based DNN scheme adapted from~\cite{donti2021dc3}. It includes a correction step for infeasible solutions in training/testing;} (iv) \textsf{DNN-W}: A hybrid method adapted from~\cite{dong2020smart}. It trains a DNN to predict the primal and dual variables as the warm-start points to the conventional solver; {(v) DNN-G: A gauge-function based DNN scheme adapted from~\cite{li2022learning}.
It enforces solution feasibility by first solving a linear program to find a feasible interior point, and then constructing the mapping between DNN prediction in an $l_{\infty}$ unit ball and the optimum.} \rev{For better evaluation, we implement two \textsf{DeepOPF+} schemes with different DNN sizes and calibration rate (3\%, 7\%) that are all within the maximum allowable one, namely \textsf{DeepOPF+-3}, and \textsf{DeepOPF+-7}.} The detailed designs are summarized in Appendix~\ref{appen:sec:details.deepopf+}. 

The results are shown in Table~\ref{tab:light.heavy} with the following observations. First, \textsf{DeepOPF+} improves over \textsf{DNN-P/DNN-D} in that it achieves consistent speedups in both light-load and heavy-load regimes. {\textsf{DNN-P/DNN-D} achieves a lower speedup in the heavy-load regime than in the light-load regime as a large percentage of its solutions are infeasible, and it needs to involve a post-processing procedure to recover the feasible solutions. Note that though \textsf{DNN-P/DNN-D} may perform well on the test set in light-load regime with a higher feasibility rate, its worst-case performance over the entire input domain can be significant,} e.g., more than $443\%$ constraints violation for Case300 in the heavy-load region. In contrast, \textsf{DeepOPF+} guarantees solution feasibility in both light-load and heavy-load regimes, \vspace{-0.002in}eliminating the need for post-processing and hence achieving consistent speedups. {Second, though the warm-start/interior point based scheme \textsf{DNN-W}/\textsf{DNN-G} ensures the feasibility of obtained solutions, they suffer low speedups/large optimality loss. As compared, \textsf{DeepOPF+} achieves noticeably better speedups as avoiding the iterations in conventional solvers}. Third, the optimality loss of \textsf{DeepOPF+} is minor and {comparable with these of the existing state-of-the-art DNN schemes}, indicating the effectiveness of the proposed \textit{Adversarial-Sample Aware} training algorithm. Fourth, we observe that the optimality loss of \textsf{DeepOPF+} increases with a larger calibration rate, which is consistent with the trade-off between optimality and calibration rate discussed in Sec.~\ref{ssec.tradeoff}. \rev{We remark that DC-OPF is an approximation to the original non-convex non-linear AC-OPF in power grid operation under several simplifications. DC-OPF is widely used for its convexity and scalability. Expanding the work to AC-OPF is a promising future work as discussed in Appendix~\ref{appendix.equality}.}

Moreover, we apply our framework to a non-convex problem in \cite{donti2021dc3} and show its performance advantage over existing schemes. Detailed design/results are shown in Appendix~\ref{appen.sec.nonconvex}.

\begin{table*}[!t]{
	\centering
	\caption{Performance comparison with SOTA DNN schemes in light-load and heavy-load regimes.}
    \renewcommand{\arraystretch}{1.1}
	\begin{threeparttable}
 	\resizebox{\textwidth}{!}{
			\begin{tabular}{c|c|c|c|c|c|c|c|c|c}
			\hline
			\multirow{2}{*}{Case}&
			\multirow{2}{*}{\tabincell{c}{Scheme}} &
			\multicolumn{2}{c|}{\tabincell{c}{Average speedups}}&
			\multicolumn{2}{c|}{\tabincell{c}{Feasibility rate (\%)}} &
			\multicolumn{2}{c|}{\tabincell{c}{Optimality loss (\%)}}&
			\multicolumn{2}{c}{\tabincell{c}{Worst-case violation (\%)}}\\
			\cline{3-10}
			&&\tabincell{c}{light-load }&\tabincell{c}{heavy-load }&\tabincell{c}{light-load }&\tabincell{c}{heavy-load }&\tabincell{c}{light-load }&\tabincell{c}{heavy-load }&\tabincell{c}{light-load }&\tabincell{c}{heavy-load}\\
			\hline
 			\multirow{6}{*}{Case30}&\tabincell{c}{\textsf{DNN-P}}& $\times$85&$\times$86&100&88.12&0.02&0.03&0&5.43\\
			\cline{2-10}
			&\tabincell{c}{\textsf{DNN-D}}& $\times$85&$\times$84&100&93.36&0.02&0.03&0&11.19\\
			\cline{2-10}
			&\tabincell{c}{\textsf{DNN-W}}& $\times$0.90&$\times$0.86&100&100&0&0&0&0 \\ 
			\cline{2-10}
			&\tabincell{c}{\textsf{DNN-G}}& $\times$24&$\times$26&100&100&0.13&0.04&0&0 \\ 
			\cline{2-10}
			&\tabincell{c}{\textsf{DeepOPF+}-3}& $\times$86&$\times$92&100&100&0.03&0.04&0&0 \\ 
			\cline{2-10}
			&\tabincell{c}{\textsf{DeepOPF+}-7}& $\times$86&$\times$93&100&100&0.03&0.09&0&0 \\ 
			\hline
 			\multirow{6}{*}{Case118}&\tabincell{c}{\textsf{DNN-P}}& $\times$137&$\times$125&68.84&54.92&0.17&0.21&19.5&44.8\\
 			\cline{2-10}
 			&\tabincell{c}{\textsf{DNN-D}}& $\times$138&$\times$124&73.42&55.37&0.20&0.24&16.69&43.1\\
			\cline{2-10}
			&\tabincell{c}{\textsf{DNN-W}}& $\times$2.08&$\times$2.26&100&100&0&0&0&0 \\ 
			\cline{2-10}
			&\tabincell{c}{\textsf{DNN-G}}& $\times$26&$\times$16&100&100&1.29&0.39&0&0 \\ 
			\cline{2-10}
			&\tabincell{c}{\textsf{DeepOPF+}-3}& $\times$201&$\times$226&100&100&0.18&0.19&0&0 \\
			\cline{2-10}
			&\tabincell{c}{\textsf{DeepOPF+}-7}& $\times$202&$\times$228&100&100&0.37&0.41&0&0 \\
			\hline
			\multirow{6}{*}{Case300}&\tabincell{c}{\textsf{DNN-P}}& $\times$115&$\times$98&91.29&78.42&0.06&0.08&261.1&443.0 \\
			\cline{2-10}
			&\tabincell{c}{\textsf{DNN-D}}& $\times$115&$\times$102&91.99&82.92&0.07&0.07&231.6&348.1\\
			\cline{2-10}
			&\tabincell{c}{DNN-W}& $\times$1.04&$\times$1.08&100&100&0&0&0&0 \\ 
			\cline{2-10}
			&\tabincell{c}{\textsf{DNN-G}}& $\times$2.44&$\times$2.65&100&100&0.32&0.06&0&0 \\ 
			\cline{2-10}
			&\tabincell{c}{\textsf{DeepOPF+}-3}& $\times$129&$\times$136&100&100&0.03&0.03&0&0 \\ 
			\cline{2-10}
			&\tabincell{c}{\textsf{DeepOPF+}-7}& $\times$130&$\times$138&100&100&0.10&0.06&0&0 \\ 

			\hline
		\end{tabular}}
		\begin{tablenotes}
			\scriptsize
			\item[*] Feasibility rate and Worst-case violation are the results \textit{before} post-processing. {Feasibility rates (resp Worst-case violation) after post-processing are 100\% (resp 0) for all DNN schemes. We hence report the results before post-processing to better show the advantage of our design. Speedup and Optimality loss are the results \textit{after} post-processing of the final obtained feasible solutions.}
			\item[*] The \textit{correction} step in \textsf{DNN-D} (with $10^{-3}$ rate) takes longer time compared with $l_1$-projection in \textsf{DNN-P}, resulting in lower speedups.
			\item[*] We empirically observe that \textsf{DNN-G} requires more training epochs for satisfactory performance. We report its best results at 500 epochs for Case118/300 in heavy-load and the results at 400 epochs for the other cases. The training epochs for the other DNN schemes are 200.
		\end{tablenotes}
	\end{threeparttable}
	\label{tab:light.heavy}}
\end{table*}

In summary, simulation results on Table~\ref{tab:DeepOPF+.variants} and Table~\ref{tab:light.heavy} demonstrate the improvement of the proposed \textsf{DeepOPF+} on ensuring universal feasibility over the existing DNN-based approach while achieving desirable optimality and speedup performance.\footnote{As DC-OPF problem is the linear approximation of the AC-OPF problem, we remark that though \textsf{DeepOPF+} guarantees universal solution feasibility for DC-OPF problem, the obtained DNN solution could not be feasible for the original AC-OPF problem~\cite{baker2021solutions}, e.g., due to the line losses ignored. We note that extending the preventive learning framework to AC-OPF problem and general non-convex problems is an interesting future direction as discussed in Section~\ref{sec:conclusion}.}

\section{Concluding remarks}\label{sec:conclusion}
In this paper, we propose a {preventive} learning framework for solving OPCC with solution feasibility guarantee. The idea is to systematically calibrate inequality constraints used in DNN training, thereby anticipating prediction errors and ensuring the resulting solutions remain feasible. The framework includes (i) deriving the maximum calibration rate to preserve the supported input region, (ii) determining the sufficient DNN size need for achieving universal feasibility, based on which a universal solution feasibility guaranteed DNN can be directly constructed without training, and (iii) a new \textit{Adversary Sample-Aware} training algorithm to improve the DNN's optimality performance while preserving the universal feasibility. Overall, the preventive learning framework provides two DNNs. The first one constructed from the step of determining the sufficient DNN size can guarantee universal feasibility (DNN-FG), while the other DNN obtained from the proposed \textit{Adversary Sample-Aware} training algorithm further improves optimality and maintains DNN's universal feasibility simultaneously (DNN Optimality Enhanced). We apply the preventive learning framework to develop \textsf{DeepOPF+} for solving the essential DC-OPF problem in grid operation.  It outperforms existing DNN-based schemes in ensuring feasibility and attaining consistent desirable speedup performance in both light-load and heavy-load regimes. Simulation results over IEEE test cases show that \textsf{DeepOPF+} generates $100\%$ feasible solutions with $<0.19\%$ optimality loss and up to two orders of magnitude computational speedup, as compared to a state-of-the-art iterative solver. We also apply our framework to a non-convex problem and show its performance advantage over existing schemes. We remark that the preventive learning framework can work for large-scale systems because of the desirable scalability of DNN. 

We note that, despite the potential of the framework and theoretical guarantee for OPLC, there are several limitations. First, we evaluate the constraints calibration and performance of DNNs via the proposed programs \eqref{equ:calibration.rate-1}--\eqref{equ:calibration.rate-2} and \eqref{equ:DNNsize-1}--\eqref{equ:DNNsize-2}, which are non-convex problems. The current solvers, e.g., Gurobi, CPLEX, or APOPT, may not provide the global optimal solutions and the valid bounds for general OPCC. Second, the proposed framework only focuses on the optimization problem with convex constraints; extending the preventive learning framework to ensure DNN solution feasibility for general non-linear constrained optimization problems like ACOPF and evaluate the performance over systems with several thousand buses and realistic loads as discussed in Appendix~\ref{appendix.equality} would be an interesting future direction.



\newpage
\bibliographystyle{IEEEtran}
\bibliography{IEEEabrv,8reference}

\begin{thebibliography}{100}
\providecommand{\url}[1]{#1}
\csname url@samestyle\endcsname
\providecommand{\newblock}{\relax}
\providecommand{\bibinfo}[2]{#2}
\providecommand{\BIBentrySTDinterwordspacing}{\spaceskip=0pt\relax}
\providecommand{\BIBentryALTinterwordstretchfactor}{4}
\providecommand{\BIBentryALTinterwordspacing}{\spaceskip=\fontdimen2\font plus
\BIBentryALTinterwordstretchfactor\fontdimen3\font minus
  \fontdimen4\font\relax}
\providecommand{\BIBforeignlanguage}[2]{{%
\expandafter\ifx\csname l@#1\endcsname\relax
\typeout{** WARNING: IEEEtran.bst: No hyphenation pattern has been}%
\typeout{** loaded for the language `#1'. Using the pattern for}%
\typeout{** the default language instead.}%
\else
\language=\csname l@#1\endcsname
\fi
#2}}
\providecommand{\BIBdecl}{\relax}
\BIBdecl

\bibitem{hornik1989multilayer}
K.~Hornik, M.~Stinchcombe, and H.~White, ``Multilayer feedforward networks are
  universal approximators,'' \emph{Neural networks}, vol.~2, no.~5, pp.
  359--366, 1989.

\bibitem{leshno1993multilayer}
M.~Leshno, V.~Y. Lin, A.~Pinkus, and S.~Schocken, ``Multilayer feedforward
  networks with a nonpolynomial activation function can approximate any
  function,'' \emph{Neural networks}, vol.~6, no.~6, pp. 861--867, 1993.

\bibitem{hanin2019universal}
B.~Hanin, ``Universal function approximation by deep neural nets with bounded
  width and relu activations,'' \emph{Mathematics}, vol.~7, no.~10, p. 992,
  2019.

\bibitem{deepopf1}
X.~Pan, T.~Zhao, and M.~Chen, ``Deepopf: Deep neural network for {DC} optimal
  power flow,'' in \emph{2019 IEEE International Conference on Communications,
  Control, and Computing Technologies for Smart Grids (SmartGridComm)}, 2019.

\bibitem{deepopf2}
X.~Pan, T.~Zhao, M.~Chen, and S.~Zhang, ``Deepopf: A deep neural network
  approach for security-constrained {DC} optimal power flow,'' \emph{IEEE
  Transactions on Power Systems}, vol.~36, no.~3, pp. 1725--1735, 2020.

\bibitem{pan2020deepopf}
X.~Pan, M.~Chen, T.~Zhao, and S.~H. Low, ``{DeepOPF: A Feasibility-Optimized
  Deep Neural Network Approach for {AC} Optimal Power Flow Problems},''
  \emph{arXiv preprint arXiv:2007.01002}, 2020.

\bibitem{donti2021dc3}
P.~L. Donti, D.~Rolnick, and J.~Z. Kolter, ``{DC3}: A learning method for
  optimization with hard constraints,'' \emph{arXiv preprint arXiv:2104.12225},
  2021.

\bibitem{chatzos2020high}
M.~Chatzos, F.~Fioretto, T.~W. Mak, and P.~Van~Hentenryck, ``High-fidelity
  machine learning approximations of large-scale optimal power flow,''
  \emph{arXiv preprint arXiv:2006.16356}, 2020.

\bibitem{lei2020data}
X.~Lei, Z.~Yang, J.~Yu, J.~Zhao, Q.~Gao, and H.~Yu, ``Data-driven optimal power
  flow: A physics-informed machine learning approach,'' \emph{IEEE Transactions
  on Power Systems}, vol.~36, no.~1, pp. 346--354, 2020.

\bibitem{huang2021deepopf}
W.~Huang, X.~Pan, M.~Chen, and S.~H. Low, ``{DeepOPF-V: Solving {AC-OPF}
  Problems Efficiently},'' \emph{accepted for IEEE Trans. Power Syst.}, 2021.

\bibitem{sun2018learning}
H.~Sun, X.~Chen, Q.~Shi, M.~Hong, X.~Fu, and N.~D. Sidiropoulos, ``Learning to
  optimize: Training deep neural networks for interference management,''
  \emph{IEEE Transactions on Signal Processing}, vol.~66, no.~20, pp.
  5438--5453, 2018.

\bibitem{xia2019deep}
W.~Xia, G.~Zheng, Y.~Zhu, J.~Zhang, J.~Wang, and A.~P. Petropulu, ``A deep
  learning framework for optimization of miso downlink beamforming,''
  \emph{IEEE Transactions on Communications}, vol.~68, no.~3, pp. 1866--1880,
  2019.

\bibitem{zhao2020deepopf}
T.~Zhao, X.~Pan, M.~Chen, A.~Venzke, and S.~H. Low, ``{DeepOPF+: A Deep Neural
  Network Approach for {DC} Optimal Power Flow for Ensuring Feasibility},''
  \emph{arXiv preprint arXiv:2009.03147}, 2020.

\bibitem{venzke2020learning}
A.~Venzke, G.~Qu, S.~Low, and S.~Chatzivasileiadis, ``Learning optimal power
  flow: Worst-case guarantees for neural networks,'' in \emph{2020 IEEE
  International Conference on Communications, Control, and Computing
  Technologies for Smart Grids (SmartGridComm)}, 2020.

\bibitem{tjeng2018evaluating}
V.~Tjeng, K.~Y. Xiao, and R.~Tedrake, ``Evaluating robustness of neural
  networks with mixed integer programming,'' in \emph{International Conference
  on Learning Representations}, 2018.

\bibitem{KotaryFHW21}
J.~Kotary, F.~Fioretto, P.~V. Hentenryck, and B.~Wilder, ``End-to-end
  constrained optimization learning: {A} survey,'' in \emph{Proceedings of
  {IJCAI} 2021, Virtual Event / Montreal, Canada, 19-27 August 2021}, Z.~Zhou,
  Ed., 2021, pp. 4475--4482.

\bibitem{zhou2022deepopf}
M.~Zhou, M.~Chen, and S.~H. Low, ``{DeepOPF-FT}: One deep neural network for
  multiple {AC-OPF} problems with flexible topology,'' \emph{IEEE Transactions
  on Power Systems}, vol.~38, no.~1, pp. 964--967, 2022.

\bibitem{guha2019machine}
N.~Guha, Z.~Wang, M.~Wytock, and A.~Majumdar, ``{Machine learning for {AC}
  optimal power flow},'' \emph{arXiv preprint arXiv:1910.08842}, 2019.

\bibitem{zamzam2020learning}
A.~S. Zamzam and K.~Baker, ``Learning optimal solutions for extremely fast {AC}
  optimal power flow,'' in \emph{2020 IEEE International Conference on
  Communications, Control, and Computing Technologies for Smart Grids
  (SmartGridComm)}, 2020.

\bibitem{fioretto2020predicting}
F.~Fioretto, T.~W. Mak, and P.~Van~Hentenryck, ``Predicting {AC} optimal power
  flows: Combining deep learning and lagrangian dual methods,'' in
  \emph{Proceedings of the AAAI Conference on Artificial Intelligence},
  vol.~34, no.~01, 2020, pp. 630--637.

\bibitem{dobbe2019towards}
R.~Dobbe, O.~Sondermeijer, D.~Fridovich-Keil, D.~Arnold, D.~Callaway, and
  C.~Tomlin, ``Toward distributed energy services: Decentralizing optimal power
  flow with machine learning,'' \emph{IEEE Transactions on Smart Grid},
  vol.~11, no.~2, pp. 1296--1306, 2019.

\bibitem{sanseverino2016multi}
E.~R. Sanseverino, M.~Di~Silvestre, L.~Mineo, S.~Favuzza, N.~Nguyen, and
  Q.~Tran, ``A multi-agent system reinforcement learning based optimal power
  flow for islanded microgrids,'' in \emph{2016 IEEE 16th International
  Conference on Environment and Electrical Engineering (EEEIC)}.\hskip 1em plus
  0.5em minus 0.4em\relax IEEE, 2016, pp. 1--6.

\bibitem{elmachtoub2022smart}
A.~N. Elmachtoub and P.~Grigas, ``Smart “predict, then optimize”,''
  \emph{Management Science}, vol.~68, no.~1, pp. 9--26, 2022.

\bibitem{deepopfngt}
W.~Huang and M.~Chen, ``{DeepOPF-NGT}: A fast unsupervised learning approach
  for solving {AC-OPF} problems without ground truth,'' in \emph{ICML 2021
  Workshop on Tackling Climate Change with Machine Learning}, 2021.

\bibitem{gutierrez2010neural}
V.~J. Gutierrez-Martinez, C.~A. Ca{\~n}izares, C.~R. Fuerte-Esquivel,
  A.~Pizano-Martinez, and X.~Gu, ``Neural-network security-boundary constrained
  optimal power flow,'' \emph{IEEE Transactions on Power Systems}, vol.~26,
  no.~1, pp. 63--72, 2010.

\bibitem{vaccaro2016knowledge}
A.~Vaccaro and C.~A. Ca{\~n}izares, ``A knowledge-based framework for power
  flow and optimal power flow analyses,'' \emph{IEEE Transactions on Smart
  Grid}, vol.~9, no.~1, pp. 230--239, 2016.

\bibitem{halilbavsic2018data}
L.~Halilba{\v{s}}i{\'c}, F.~Thams, A.~Venzke, S.~Chatzivasileiadis, and
  P.~Pinson, ``Data-driven security-constrained {AC-OPF} for operations and
  markets,'' in \emph{2018 Power Systems Computation Conference (PSCC)}.\hskip
  1em plus 0.5em minus 0.4em\relax IEEE, 2018, pp. 1--7.

\bibitem{biagioni2020learning}
D.~Biagioni, P.~Graf, X.~Zhang, A.~S. Zamzam, K.~Baker, and J.~King,
  ``Learning-accelerated {ADMM} for distributed {DC} optimal power flow,''
  \emph{IEEE Control Systems Letters}, 2020.

\bibitem{pineda2020data}
S.~Pineda, J.~M. Morales, and A.~Jim{\'e}nez-Cordero, ``Data-driven screening
  of network constraints for unit commitment,'' \emph{IEEE Transactions on
  Power Systems}, vol.~35, no.~5, pp. 3695--3705, 2020.

\bibitem{jamei2019meta}
M.~Jamei, L.~Mones, A.~Robson, L.~White, J.~Requeima, and C.~Ududec,
  ``Meta-optimization of optimal power flow,'' \emph{ICML Workshop, Climate
  Change: How Can AI Help}, 2019.

\bibitem{deka2019learning}
D.~Deka and S.~Misra, ``Learning for {DC-OPF}: Classifying active sets using
  neural nets,'' in \emph{2019 IEEE Milan PowerTech}.\hskip 1em plus 0.5em
  minus 0.4em\relax IEEE, 2019, pp. 1--6.

\bibitem{karagiannopoulos2019data}
S.~Karagiannopoulos, P.~Aristidou, and G.~Hug, ``Data-driven local control
  design for active distribution grids using off-line optimal power flow and
  machine learning techniques,'' \emph{IEEE Transactions on Smart Grid},
  vol.~10, no.~6, pp. 6461--6471, 2019.

\bibitem{baker2019joint}
K.~Baker and A.~Bernstein, ``Joint chance constraints in {AC} optimal power
  flow: Improving bounds through learning,'' \emph{IEEE Transactions on Smart
  Grid}, vol.~10, no.~6, pp. 6376--6385, 2019.

\bibitem{ng2018statistical}
Y.~Ng, S.~Misra, L.~A. Roald, and S.~Backhaus, ``{Statistical learning for {DC}
  optimal power flow},'' in \emph{2018 Power Systems Computation Conference
  (PSCC)}.\hskip 1em plus 0.5em minus 0.4em\relax IEEE, 2018, pp. 1--7.

\bibitem{misra2018learning}
S.~Misra, L.~Roald, and Y.~Ng, ``{Learning for Constrained Optimization:
  Identifying Optimal Active Constraint Sets},'' \emph{arXiv preprint
  arXiv:1802.09639}, 2018.

\bibitem{zhai2010fast}
Q.~Zhai, X.~Guan, J.~Cheng, and H.~Wu, ``{Fast identification of inactive
  security constraints in SCUC problems},'' \emph{IEEE Transactions on Power
  Systems}, vol.~25, no.~4, pp. 1946--1954, 2010.

\bibitem{roald2019implied}
L.~A. Roald and D.~K. Molzahn, ``{Implied constraint satisfaction in power
  system optimization: the impacts of load variations},'' in \emph{2019 57th
  Annual Allerton Conference on Communication, Control, and Computing
  (Allerton)}.\hskip 1em plus 0.5em minus 0.4em\relax IEEE, 2019, pp. 308--315.

\bibitem{9091534}
L.~{Duchesne}, E.~{Karangelos}, and L.~{Wehenkel}, ``{Recent Developments in
  Machine Learning for Energy Systems Reliability Management},''
  \emph{Proceedings of the IEEE}, vol. 108, no.~9, pp. 1656--1676, 2020.

\bibitem{chen2020learning}
Y.~Chen and B.~Zhang, ``Learning to solve network flow problems via neural
  decoding,'' \emph{arXiv preprint arXiv:2002.04091}, 2020.

\bibitem{baker2019learning}
K.~Baker, ``Learning warm-start points for {AC} optimal power flow,'' in
  \emph{2019 IEEE 29th International Workshop on Machine Learning for Signal
  Processing (MLSP)}.\hskip 1em plus 0.5em minus 0.4em\relax IEEE, 2019, pp.
  1--6.

\bibitem{dong2020smart}
W.~Dong, Z.~Xie, G.~Kestor, and D.~Li, ``Smart-pgsim: using neural network to
  accelerate {AC-OPF} power grid simulation,'' in \emph{SC20: International
  Conference for High Performance Computing, Networking, Storage and
  Analysis}.\hskip 1em plus 0.5em minus 0.4em\relax IEEE, 2020, pp. 1--15.

\bibitem{balcan2018learning}
M.-F. Balcan, T.~Dick, T.~Sandholm, and E.~Vitercik, ``Learning to branch,'' in
  \emph{International conference on machine learning}.\hskip 1em plus 0.5em
  minus 0.4em\relax PMLR, 2018, pp. 344--353.

\bibitem{he2014learning}
H.~He, H.~Daume~III, and J.~M. Eisner, ``Learning to search in branch and bound
  algorithms,'' \emph{Advances in neural information processing systems},
  vol.~27, pp. 3293--3301, 2014.

\bibitem{li2016learning}
K.~Li and J.~Malik, ``Learning to optimize,'' \emph{arXiv preprint
  arXiv:1606.01885}, 2016.

\bibitem{chen2017learning}
Y.~Chen, M.~W. Hoffman, S.~G. Colmenarejo, M.~Denil, T.~P. Lillicrap,
  M.~Botvinick, and N.~Freitas, ``Learning to learn without gradient descent by
  gradient descent,'' in \emph{International Conference on Machine
  Learning}.\hskip 1em plus 0.5em minus 0.4em\relax PMLR, 2017, pp. 748--756.

\bibitem{nellikkath2021physics}
R.~Nellikkath and S.~Chatzivasileiadis, ``Physics-informed neural networks for
  minimising worst-case violations in {DC} optimal power flow,'' \emph{arXiv
  preprint arXiv:2107.00465}, 2021.

\bibitem{nellikkath2021physics2}
------, ``Physics-informed neural networks for ac optimal power flow,''
  \emph{arXiv preprint arXiv:2110.02672}, 2021.

\bibitem{zhang2020convex}
L.~Zhang, Y.~Chen, and B.~Zhang, ``A convex neural network solver for {DCOPF}
  with generalization guarantees,'' \emph{arXiv preprint arXiv:2009.09109},
  2020.

\bibitem{wang2019satnet}
P.-W. Wang, P.~Donti, B.~Wilder, and Z.~Kolter, ``Satnet: Bridging deep
  learning and logical reasoning using a differentiable satisfiability
  solver,'' in \emph{International Conference on Machine Learning}.\hskip 1em
  plus 0.5em minus 0.4em\relax PMLR, 2019, pp. 6545--6554.

\bibitem{chen2018neural}
R.~T. Chen, Y.~Rubanova, J.~Bettencourt, and D.~Duvenaud, ``Neural ordinary
  differential equations,'' \emph{arXiv preprint arXiv:1806.07366}, 2018.

\bibitem{ling2018game}
C.~K. Ling, F.~Fang, and J.~Z. Kolter, ``What game are we playing? end-to-end
  learning in normal and extensive form games,'' \emph{arXiv preprint
  arXiv:1805.02777}, 2018.

\bibitem{de2018end}
F.~de~Avila Belbute-Peres, K.~Smith, K.~Allen, J.~Tenenbaum, and J.~Z. Kolter,
  ``End-to-end differentiable physics for learning and control,''
  \emph{Advances in neural information processing systems}, vol.~31, pp.
  7178--7189, 2018.

\bibitem{bai2019deep}
S.~Bai, J.~Z. Kolter, and V.~Koltun, ``Deep equilibrium models,'' \emph{arXiv
  preprint arXiv:1909.01377}, 2019.

\bibitem{donti2017task}
P.~L. Donti, B.~Amos, and J.~Z. Kolter, ``Task-based end-to-end model learning
  in stochastic optimization,'' \emph{arXiv preprint arXiv:1703.04529}, 2017.

\bibitem{djolonga2017differentiable}
J.~Djolonga and A.~Krause, ``Differentiable learning of submodular models,''
  \emph{Advances in Neural Information Processing Systems}, vol.~30, pp.
  1013--1023, 2017.

\bibitem{tschiatschek2018differentiable}
S.~Tschiatschek, A.~Sahin, and A.~Krause, ``Differentiable submodular
  maximization,'' \emph{arXiv preprint arXiv:1803.01785}, 2018.

\bibitem{wilder2019melding}
B.~Wilder, B.~Dilkina, and M.~Tambe, ``Melding the data-decisions pipeline:
  Decision-focused learning for combinatorial optimization,'' in
  \emph{Proceedings of the AAAI Conference on Artificial Intelligence},
  vol.~33, no.~01, 2019, pp. 1658--1665.

\bibitem{gould2019deep}
S.~Gould, R.~Hartley, and D.~Campbell, ``Deep declarative networks: A new
  hope,'' \emph{arXiv preprint arXiv:1909.04866}, 2019.

\bibitem{amos2017optnet}
B.~Amos and J.~Z. Kolter, ``Optnet: Differentiable optimization as a layer in
  neural networks,'' in \emph{International Conference on Machine
  Learning}.\hskip 1em plus 0.5em minus 0.4em\relax PMLR, 2017, pp. 136--145.

\bibitem{agrawal2019differentiable}
A.~Agrawal, B.~Amos, S.~Barratt, S.~Boyd, S.~Diamond, and J.~Z. Kolter,
  ``Differentiable convex optimization layers,'' \emph{Advances in Neural
  Information Processing Systems}, vol.~32, pp. 9562--9574, 2019.

\bibitem{li2022learning}
M.~Li, S.~Kolouri, and J.~Mohammadi, ``Learning to solve optimization problems
  with hard linear constraints,'' \emph{arXiv preprint arXiv:2208.10611}, 2022.

\bibitem{ferrari2009multiobjective}
S.~Ferrari, ``Multiobjective algebraic synthesis of neural control systems by
  implicit model following,'' \emph{IEEE transactions on neural networks},
  vol.~20, no.~3, pp. 406--419, 2009.

\bibitem{yin2022learning}
H.~Yin, V.~Kekatos, M.~Jin \emph{et~al.}, ``Learning neural networks under
  input-output specifications,'' \emph{arXiv preprint arXiv:2202.11246}, 2022.

\bibitem{qin2019verification}
C.~Qin, B.~O'Donoghue, R.~Bunel, R.~Stanforth, S.~Gowal, J.~Uesato,
  G.~Swirszcz, P.~Kohli \emph{et~al.}, ``Verification of non-linear
  specifications for neural networks,'' \emph{arXiv preprint arXiv:1902.09592},
  2019.

\bibitem{limanond1998neural}
S.~Limanond and J.~Si, ``Neural network-based control design: An lmi
  approach,'' \emph{IEEE Transactions on Neural Networks}, vol.~9, no.~6, pp.
  1422--1429, 1998.

\bibitem{sotoudeh2021provable}
M.~Sotoudeh and A.~V. Thakur, ``Provable repair of deep neural networks,'' in
  \emph{Proceedings of the 42nd ACM SIGPLAN International Conference on
  Programming Language Design and Implementation}, 2021, pp. 588--603.

\bibitem{yang2021neural}
X.~Yang, T.~Yamaguchi, H.-D. Tran, B.~Hoxha, T.~T. Johnson, and D.~Prokhorov,
  ``Neural network repair with reachability analysis,'' \emph{arXiv preprint
  arXiv:2108.04214}, 2021.

\bibitem{sohn2019search}
J.~Sohn, S.~Kang, and S.~Yoo, ``Search based repair of deep neural networks,''
  \emph{arXiv preprint arXiv:1912.12463}, 2019.

\bibitem{sheikholeslami2020provably}
F.~Sheikholeslami, A.~Lotfi, and J.~Z. Kolter, ``Provably robust classification
  of adversarial examples with detection,'' in \emph{International Conference
  on Learning Representations}, 2020.

\bibitem{dvijotham2018dual}
K.~Dvijotham, R.~Stanforth, S.~Gowal, T.~A. Mann, and P.~Kohli, ``A dual
  approach to scalable verification of deep networks.'' in \emph{UAI}, vol.~1,
  no.~2, 2018, p.~3.

\bibitem{wong2018provable}
E.~Wong and Z.~Kolter, ``Provable defenses against adversarial examples via the
  convex outer adversarial polytope,'' in \emph{International Conference on
  Machine Learning}.\hskip 1em plus 0.5em minus 0.4em\relax PMLR, 2018, pp.
  5286--5295.

\bibitem{li2020sok}
L.~Li, X.~Qi, T.~Xie, and B.~Li, ``Sok: Certified robustness for deep neural
  networks,'' \emph{arXiv preprint arXiv:2009.04131}, 2020.

\bibitem{boyd2004convex}
S.~Boyd, S.~P. Boyd, and L.~Vandenberghe, \emph{Convex optimization}.\hskip 1em
  plus 0.5em minus 0.4em\relax Cambridge university press, 2004.

\bibitem{faisca2007multiparametric}
N.~P. Fa{\'\i}sca, V.~Dua, and E.~N. Pistikopoulos, ``Multiparametric linear
  and quadratic programming,'' pp. 3--23, 2007.

\bibitem{bemporad2000explicit}
A.~Bemporad, M.~Morari, V.~Dua, and E.~N. Pistikopoulos, ``The explicit
  solution of model predictive control via multiparametric quadratic
  programming,'' in \emph{Proceedings of the 2000 American Control Conference.
  ACC (IEEE Cat. No. 00CH36334)}, vol.~2.\hskip 1em plus 0.5em minus
  0.4em\relax IEEE, 2000, pp. 872--876.

\bibitem{ye1989extension}
Y.~Ye and E.~Tse, ``An extension of karmarkar's projective algorithm for convex
  quadratic programming,'' \emph{Mathematical Programming}, vol.~44, no.~1, pp.
  157--179, May 1989.

\bibitem{bemporad2006algorithm}
A.~Bemporad and C.~Filippi, ``An algorithm for approximate multiparametric
  convex programming,'' \emph{Computational optimization and applications},
  vol.~35, no.~1, pp. 87--108, 2006.

\bibitem{goodfellow2016deepma}
I.~Goodfellow, Y.~Bengio, A.~Courville, and Y.~Bengio, \emph{{Deep
  Learning}}.\hskip 1em plus 0.5em minus 0.4em\relax MIT Press Cambridge, 2016,
  vol.~1.

\bibitem{ben1990computational}
O.~Ben-Ayed and C.~E. Blair, ``Computational difficulties of bilevel linear
  programming,'' \emph{Operations Research}, vol.~38, no.~3, pp. 556--560,
  1990.

\bibitem{jeroslow1985polynomial}
R.~G. Jeroslow, ``The polynomial hierarchy and a simple model for competitive
  analysis,'' \emph{Mathematical programming}, vol.~32, no.~2, pp. 146--164,
  1985.

\bibitem{lawler1966branch}
E.~L. Lawler and D.~E. Wood, ``Branch-and-bound methods: A survey,''
  \emph{Operations research}, vol.~14, no.~4, pp. 699--719, 1966.

\bibitem{vaidya1989speeding}
P.~M. Vaidya, ``Speeding-up linear programming using fast matrix
  multiplication,'' in \emph{IEEE FOCS}, 1989, pp. 332--337.

\bibitem{hanin2017approximating}
B.~Hanin and M.~Sellke, ``Approximating continuous functions by relu nets of
  minimal width,'' \emph{arXiv preprint arXiv:1710.11278}, 2017.

\bibitem{kidger2020universal}
P.~Kidger and T.~Lyons, ``Universal approximation with deep narrow networks,''
  in \emph{Conference on learning theory}.\hskip 1em plus 0.5em minus
  0.4em\relax PMLR, 2020, pp. 2306--2327.

\bibitem{hornik1991approximation}
K.~Hornik, ``Approximation capabilities of multilayer feedforward networks,''
  \emph{Neural networks}, vol.~4, no.~2, pp. 251--257, 1991.

\bibitem{karg2020efficient}
B.~Karg and S.~Lucia, ``Efficient representation and approximation of model
  predictive control laws via deep learning,'' \emph{IEEE Transactions on
  Cybernetics}, vol.~50, no.~9, pp. 3866--3878, 2020.

\bibitem{krizhevsky2012imagenet}
A.~Krizhevsky, I.~Sutskever, and G.~E. Hinton, ``Imagenet classification with
  deep convolutional neural networks,'' in \emph{Proceedings of the
  International Conference on Neural Information Processing Systems}, vol.~1,
  Lake Tahoe, Nevada, USA, 2012, pp. 1097--1105.

\bibitem{danskin}
Y.~Dong, Z.~Deng, T.~Pang, J.~Zhu, and H.~Su, ``Adversarial distributional
  training for robust deep learning,'' in \emph{Advances in Neural Information
  Processing Systems}, vol.~33.\hskip 1em plus 0.5em minus 0.4em\relax Curran
  Associates, Inc., 2020, pp. 8270--8283.

\bibitem{danskin2012theory}
J.~M. Danskin, \emph{The theory of max-min and its application to weapons
  allocation problems}.\hskip 1em plus 0.5em minus 0.4em\relax Springer Science
  \& Business Media, 2012, vol.~5.

\bibitem{kolter2018adversarial}
Z.~Kolter and A.~Madry, ``Adversarial robustness: Theory and practice,''
  \emph{Tutorial at NeurIPS}, p.~3, 2018.

\bibitem{madry2018towards}
A.~Madry, A.~Makelov, L.~Schmidt, D.~Tsipras, and A.~Vladu, ``Towards deep
  learning models resistant to adversarial attacks,'' in \emph{International
  Conference on Learning Representations}, 2018.

\bibitem{chakraborty2018adversarial}
A.~Chakraborty, M.~Alam, V.~Dey, A.~Chattopadhyay, and D.~Mukhopadhyay,
  ``Adversarial attacks and defences: A survey,'' \emph{arXiv preprint
  arXiv:1810.00069}, 2018.

\bibitem{ren2020survey}
P.~Ren, Y.~Xiao, X.~Chang, P.-Y. Huang, Z.~Li, X.~Chen, and X.~Wang, ``A survey
  of deep active learning,'' \emph{arXiv preprint arXiv:2009.00236}, 2020.

\bibitem{qian1999momentum}
N.~Qian, ``On the momentum term in gradient descent learning algorithms,''
  \emph{Neural networks}, vol.~12, no.~1, pp. 145--151, 1999.

\bibitem{frank2012optimal1}
S.~Frank, I.~Steponavice, and S.~Rebennack, ``Optimal power flow: a
  bibliographic survey i,'' \emph{Energy Systems}, vol.~3, no.~3, pp. 221--258,
  Sep 2012.

\bibitem{frank2012optimal2}
------, ``Optimal power flow: a bibliographic survey ii,'' \emph{Energy
  Systems}, vol.~3, no.~3, pp. 259--289, Sep 2012.

\bibitem{260897}
J.~H. Park, Y.~S. Kim, I.~K. Eom, and K.~Y. Lee, ``Economic load dispatch for
  piecewise quadratic cost function using hopfield neural network,'' \emph{IEEE
  Transactions on Power Systems}, vol.~8, no.~3, pp. 1030--1038, Aug 1993.

\bibitem{tpcwTrey4}
``{Power Systems Test Case Archive},'' 2018,
  \url{http://labs.ece.uw.edu/pstca/}.

\bibitem{cain2012history}
M.~B. Cain, R.~P. O’neill, and A.~Castillo, ``History of optimal power flow
  and formulations,'' \emph{Federal Energy Regulatory Commission}, vol.~1, pp.
  1--36, 2012.

\bibitem{yang2018general}
Z.~Yang, K.~Xie, J.~Yu, H.~Zhong, N.~Zhang, and Q.~Xia, ``A general formulation
  of linear power flow models: Basic theory and error analysis,'' \emph{IEEE
  Transactions on Power Systems}, vol.~34, no.~2, pp. 1315--1324, 2018.

\bibitem{bolognani2015fast}
S.~Bolognani and F.~D{\"o}rfler, ``Fast power system analysis via implicit
  linearization of the power flow manifold,'' in \emph{2015 53rd Annual
  Allerton Conference on Communication, Control, and Computing
  (Allerton)}.\hskip 1em plus 0.5em minus 0.4em\relax IEEE, 2015, pp. 402--409.

\bibitem{zimmerman2011matpower}
R.~D. Zimmerman, C.~E. Murillo-S{\'a}nchez, R.~J. Thomas \emph{et~al.},
  ``{MATPOWER: Steady-state operations, planning, and analysis tools for power
  systems research and education},'' \emph{IEEE Transactions on Power Systems},
  vol.~26, no.~1, pp. 12--19, 2011.

\bibitem{babaeinejadsarookolaee2019power}
S.~Babaeinejadsarookolaee, A.~Birchfield, R.~D. Christie, C.~Coffrin,
  C.~DeMarco, R.~Diao, M.~Ferris, S.~Fliscounakis, S.~Greene, R.~Huang
  \emph{et~al.}, ``The power grid library for benchmarking {AC} optimal power
  flow algorithms,'' \emph{arXiv preprint arXiv:1908.02788}, 2019.

\bibitem{tpcwTrey1}
``{pypower},'' 2018, \url{https://pypi.org/project/PYPOWER/}.

\bibitem{baker2021solutions}
K.~Baker, ``Solutions of {DC OPF} are never {AC} feasible,'' in
  \emph{Proceedings of the Twelfth ACM International Conference on Future
  Energy Systems}, 2021, pp. 264--268.

\bibitem{kotary2021learning}
J.~Kotary, F.~Fioretto, and P.~Van~Hentenryck, ``Learning hard optimization
  problems: A data generation perspective,'' \emph{Advances in Neural
  Information Processing Systems}, vol.~34, pp. 24\,981--24\,992, 2021.

\bibitem{pan2022deepopf}
X.~Pan, W.~Huang, M.~Chen, and S.~H. Low, ``Deepopf-{AL}: Augmented learning
  for solving {AC-OPF} problems with multiple load-solution mappings,'' in
  \emph{Proceedings of the Fourteenth ACM International Conference on Future
  Energy Systems (ACM e-Energy 2023)}, 2023.

\bibitem{fortuny1981representation}
J.~Fortuny-Amat and B.~McCarl, ``A representation and economic interpretation
  of a two-level programming problem,'' \emph{Journal of the operational
  Research Society}, vol.~32, no.~9, pp. 783--792, 1981.

\bibitem{chatzivasileiadis2018optimization}
S.~Chatzivasileiadis, ``Optimization in modern power systems,'' \emph{Lecture
  Notes. Tech. Univ. of Denmark. Available online: https://arxiv.
  org/pdf/1811.00943. pdf}, 2018.

\end{thebibliography}

\begin{appendices}

\section{Analytical formulation of $\mathcal{D}$, \rev{uniqueness of the OPCLC solution, and unbounded variable}}\label{appendix.D}
Set $\mathcal{D}$ is of problem dependent. For example, in DC-OPF problems, $\mathcal{D}$ represents the interested load input domain which is set by the system operator, e.g., feasible load within [100\%, 130\%] of the default load. For others applications, $\mathcal{D}$ represents region of \textit{possible} feasible problem inputs. Calculating the analytical representation of the feasible region of $\mathbf{\theta}$ is known as projection of a polyhedral set to lower dimension subspace. That is, $\mathcal{D}$ can be analytically obtained by projecting the following set
$$
\mathcal{P}=\{(\mathbf{\theta},\mathbf{x})| \mathbf{A}_{\boldsymbol{\theta}}\boldsymbol{\theta}\leq \boldsymbol{b_{\theta}}, \text{and} (\ref{equ:OPCC.formulation-3}),(\ref{equ:OPCC.formulation-4})\ \text{hold} \}
$$
onto the subspace of $\mathbf{\theta}$, which is still a convex polytope. The goal can be achieve using the Fourier–Motzkin elimination technique. Nevertheless, in our design, we do not need to access the full analytical formulation of $\mathcal{D}$. Instead, we introduce a set of auxiliary variable $\mathbf{\tilde{x}}$ associated with each $\mathbf{\theta}$. That is, the constraint $\mathbf{\theta}\in\mathcal{D}$ is indeed represented as $\{\mathbf{A}_{\boldsymbol{\theta}}\boldsymbol{\theta}\leq \boldsymbol{b_{\theta}}, g_j(\mathbf{\tilde{x}},\theta)\leq e_j, \forall j\in\mathcal{E}\}$.

\subsection{Uniqueness of the OPCC solution}
\rev{We would like to further discuss the assumption of the uniqueness of the OPCC solution. First, many OPCC are unique given their objective functions are strictly convex and constraints are linear. Such a condition holds for DC-OPF problems in power systems \cite{deepopf1} and model-predictive control problems in control systems \cite{bemporad2000explicit}. As proved in \cite{pan2020deepopf}, if the optimal solution is unique, the input-solution mapping is continuous while the DNN function is also continuous, which forms the underlying reason why DNN can be applied to learning such a mapping from the Universal Approximation Theorem of DNN for continuous functions. 

We would like to further discuss the situation if the optimal solution is not unique, which is an open problem and the challenge of the existing end-to-end DNN design.

Given a OPCC that admits multiple optimal solutions for the input, there indeed does not exist an injective mapping between input to solution, i.e., there exist multiple input-solution mappings. Consider the DNN training in this case, if the ground-truth training data are from different input-solution mappings, the DNN could present unsatisfactory performance as  solutions to closely related instances may exhibit large differences and the learning task can become inherently more difficult \cite{kotary2021learning,deepopfngt,pan2022deepopf}. Nevertheless, our approach is still applicable to such a scenario as the first obtained \textsf{DNN-FG} after determining the sufficient DNN size can still guarantee universal feasibility. As introduced in Sec. ~\ref{ssec:calibrationrange} and Sec.~\ref{ssec:feasibilityDNN}, deriving the calibration rate and determining the sufficient DNN size is only related to the OPCC constraints. These steps only require obtaining one of the continuous feasible mappings but not optimality. Towards the Adversarial-Sample Aware algorithm, it is straightforward to adopt the approaches in \cite{kotary2021learning,deepopfngt,pan2022deepopf} by improving the training data quality, applying the unsupervised learning idea, or learning the high-dimensional input+initial point to optimal solution mapping, which we leave for future work. Finally, the simulations on non-convex optimization (can have non-unique optimum) in Appendix~\ref{appen.sec.nonconvex} show that the ASA algorithm can still work well, showing the robustness of the design.

\subsection{Unbounded decision variables}
 There are two approaches to handle the unbounded variables: 1) setting $\underline{x}_i$ or $\bar{x}_i$ to be some arbitrarily small/large numbers. 2) only includes the bounded constraints into (4)-(5) and (6), e.g., for the variables 1) without lower bound, the DNN output is $\hat{\boldsymbol{x}}_i=-\sigma\left( \bar{\boldsymbol{x}}_i-(\boldsymbol{W_oh_{N_{{\text{hid}}}}}+\boldsymbol{b_o})_i\right)+\bar{\boldsymbol{x}}_i$; 2) without upper bound $\hat{\boldsymbol{x}}_i=\boldsymbol{\tilde{h}}_i$; 3) without both upper and lower bound, $\hat{\boldsymbol{x}}_i=(\boldsymbol{W_oh_{N_{{\text{hid}}}}}+\boldsymbol{b_o})_i$.}

\section{Handling Equality and Non-linear Constraints}\label{appendix.equality}
We remark that for general OPCC and other constrained optimizations, we can always removing the equality constraints explicitly/implicitly. Given $N+p$ variables and $p$ (linear) equality constraints, we can remove these equalities and representing $p$ variables by the remaining $N$ variables using the equality constraints, e.g., applying the coefficient matrix inversion as discussed in Appendix~\ref{appendix.deepopf+} without losing optimality. We thus focus on OPCC with inequality constraints only. The similar predict-and-reconstruct idea is proposed in~\cite{deepopf1,donti2021dc3}. In addition, we note that the proposed \textit{preventive leaning} framework is also applicable to non-linear inequality constraints, e.g., AC-OPF problems with several thousand buses, but with additional computational challenges in solving the related programs corresponding to the required steps. We leave the application to optimization with non-linear constraints and non-convex objective \rev{with large DNN size} for future study.

\rev{In this work, we consider the variation of the RHS of the linear inequality constraints. It is also interesting to study the varying $a_j, b_j, e_j$ in OPLC or other problem parameters in general OPCC and constrained optimizations. We believe our approach is still applicable to such a case while may have additional computational challenges as the problem turn to be non-linearly constrained. Nevertheless, it is also great interest to study problems whose parameters are not varying. For example, in DC-OPF, $a_j, b_j, e_j$ are determined by power network topology, which will not change significantly over a long time scale, e.g., months to years. Hence, it is reasonable and practical to study OPLC with varying inputs only.}

\section{Removing Non-Critical Inequality Constraints}\label{appen:removing.noncritical}
Considering the original OPCC without removing the non-critical constraints:
\begin{align}
    \min_{\boldsymbol{x}\in \mathcal{R}^N} \ &f(\boldsymbol{x},\boldsymbol{\theta})\label{appen:OPCC.formulation-1} \\
     \mathrm{s.t.} \quad &g_j(\boldsymbol{x,\theta})\leq e_j,\ j=1,\ldots,m, m+1,\ldots,m+q.\label{appen:OPCC.formulation-3}
     \\
     &\underline{x}_k\leq x_k\leq \bar{x}_k,\ k=1,...N.\label{appen:OPCC.formulation-4}
\end{align}

We solve the following problem for each inequality constraint $g_j, j=1,\ldots,m, m+1,\ldots,m+q$ to identify if it is critical, i.e,. whether it is active for at least one combination of the feasible input parameter $\boldsymbol{\theta}$ and  $\boldsymbol{x}$: 
\begin{align}
    \max_{\boldsymbol{\theta}, \boldsymbol{x}} \ &g_j(\boldsymbol{\theta}, \boldsymbol{x})-e_j \label{appen:equ:constraints.violation-1}\\
      \mathrm{s.t.} \quad & \eqref{appen:OPCC.formulation-4}, \ \boldsymbol{\theta}\in\mathcal{D} \notag
\end{align}
 \eqref{appen:OPCC.formulation-4} enforces the feasibility of the decision variables, which indicates the solution space of $\boldsymbol{x}$. 

It should be clear that if the optimal value of \eqref{appen:equ:constraints.violation-1} is non-positive for the $j$-th inequality constraint, i.e., $g_j\leq e_j$, then this inequality constraint is not critical in the sense that it can be removed without changing the optimal solution of OPCC for any input parameter in $\mathcal{D}$. {We remark that if some inequality constraints $g_j$ is linear w.r.t. $\boldsymbol{x}$ and $\boldsymbol{\theta}$, then program \eqref{appen:equ:constraints.violation-1} turns to be an LP, which can be efficiently solved global optimally by the existing solvers. Such condition holds for the DC-OPF problem studied in this work. For general convex $g_j$ constraints, the program \eqref{appen:equ:constraints.violation-1} is a non-convex optimization problem that can be NP-hard itself since we maximize with the convex objective. The existing solvers may not be able to solving the problem global optimally. Therefore, any (sub-optimal) solution provided by the solvers is a lower bound on \eqref{appen:equ:constraints.violation-1}. We remark that if for some $g_j$, the obtained (sub-optimal) solution is positive, then such constraints is ensured to be critical and should not be removed. We leave the study on how to solve \eqref{appen:equ:constraints.violation-1} global optimally for general convex constraints for future study.} By solving \eqref{appen:equ:constraints.violation-1} for all the inequality constraints, we obtain a set ${\mathcal{E}}$ of critical inequality constraints whose optimal objectives are positive.


In~\cite{venzke2020learning}, the authors adopt a similar idea to study the worst-case performance of DNNs in DC-OPF application, given the specification of DNN parameters. It is worth noticing that there exist several differences between these two problems.  First, we consider the individual inequality constraint instead of the overall maximum constraints violation within the entire input-solution combinations. Second, we restrict the predicted variables via \eqref{appen:OPCC.formulation-4}. The benefits lie in that 1) it helps target each critical inequality constraints given an input parameter region, which is necessary for the further constraints calibration procedure, 2) it considers 
all possible occurrence of decision variables $\boldsymbol{x}$, which is the case of any possible output of DNNs, and 3) it indicates the effectiveness of the two clamp-equivalent actions in \eqref{dnn.model} or the Sigmoid function at the output layer of DNNs, which helps guarantee predicted variables' feasibility.


\section{Formulation of Multiparameter Quadratic Program}\label{appen:mpqp-f}
We provide the formulation of multiparameter quadratic program (mp-QP):
\begin{align}
    z(\boldsymbol{\theta})=\min \quad & \frac{1}{2}\boldsymbol{x^T}\boldsymbol{Q}\boldsymbol{x}+\boldsymbol{d^T}\boldsymbol{x}\label{appen:equ.mpqp-obj}\\
    \mathrm{s.t.} \quad & \boldsymbol{a^T_j}\boldsymbol{x}+\boldsymbol{b^T_j}\boldsymbol{\theta}\leq e_j,\quad j=1,\ldots,m,\label{appen:equ.mpqp-ineq}\\
    &\underline{x}_k\leq x_k\leq \bar{x}_k,\ k=1,...,N, \label{appen:box}\\
    \mathrm{var.} \quad& \boldsymbol{x}\in \mathcal{R}^N,\notag 
\end{align}
where $\boldsymbol{x}\in\mathcal{R}^N$ are the decision variables, $\boldsymbol{\theta}\in\mathcal{R}^M$ are the input parameters, $\boldsymbol{a_j}\in\mathcal{R}^N,\boldsymbol{b_j}\in\mathcal{R}^M,e_j\in\mathcal{R}$ are the coefficients of the equality and inequality constraints. $\boldsymbol{d}\in\mathcal{R}^N$ is a constant vector, $\boldsymbol{Q}$ is an ($N\times N$) symmetric positive definite constant matrix.
The above parametric mp-QP problem asks for the least objective for each input $\boldsymbol{\theta}$. 

An applicable result from multiparametric programming that constrains the structure of the input-solution mapping $\Omega: \boldsymbol{\theta} \mapsto \boldsymbol{x}^*(\boldsymbol{\theta})$ is that $\boldsymbol{x}^*(\boldsymbol{\theta})$ is piece-wise continuous linear and the optimal objective $z(\boldsymbol{\theta})$ is piece-wise quadratic w.r.t. $\boldsymbol{\theta}$~\cite{faisca2007multiparametric}.

\section{Mixed-Integer Reformulation of Bi-Level Linear Programs}\label{appen:sec.bilevel.lp.reformulation}


Consider the following the linear constrained bi-level min-max problem: 
\begin{align}
    \min_{\boldsymbol{\theta}} \max_{\boldsymbol{x}}  \quad  &\boldsymbol{c^T} \boldsymbol{x} \\
    \mathrm{s.t.} \quad &\mathbf{A}\boldsymbol{x}\leq \boldsymbol{b}+\mathbf{F}\boldsymbol{\theta}, \\ 
    &\boldsymbol{\theta}\in{\mathcal{D}},
\end{align}
where $\mathbf{A}\in\mathcal{R}^{p\times N}$, $\boldsymbol{b}\in\mathcal{R}^p$, $\mathbf{F}\in\mathcal{R}^{p\times M}$.

The above linear constrained bi-level program can be reformulated by introducing the sufficient and necessary KKT conditions~\cite{boyd2004convex} of the inner maximization problem. We present the reformulated program in the following:
\begin{align}
    \min_{\boldsymbol{\theta},\boldsymbol{x}, \boldsymbol{y}} \quad  &\boldsymbol{c^T}\boldsymbol{x} \label{appen:equ.lp.kkt1}\\
    \mathrm{s.t.} \quad &\mathbf{A}\boldsymbol{x}\leq \boldsymbol{b}+\mathbf{F}\boldsymbol{\theta}, \ (\text{Primal feasibility})\\ 
    & \mathbf{A}^T\boldsymbol{y}=\boldsymbol{c}, \ (\text{Stationarity})\\
    & y_{i}\geq 0, \ i=1,\ldots,p, \ (\text{Dual feasibility})\\
    & y_{i}(\boldsymbol{a^T_i}\boldsymbol{x}-b_i-\boldsymbol{f^T_i}\boldsymbol{\theta})=0, \ i=1,\ldots,p, (\text{Complementary slackness}) \label{appen:equ.cs}\\
    &\boldsymbol{\theta}\in{\mathcal{D}}. \label{appen:equ.lp.kkt6}
\end{align}
Here $\boldsymbol{a_i}$ and $\boldsymbol{f_i}$ denote the $i$-th row of matrix $\mathbf{A}$ and $\mathbf{F}$ respectively. We remark that the nonlinear \textit{Complementary Slackness} condition in \eqref{appen:equ.cs} can be reformulated to be mixed-integer linear using the Fortuny-Amat McCarl linearization~\cite{fortuny1981representation}:
\begin{align}
    y_i\leq (1-r_i)C, \quad \boldsymbol{a^T_i}\boldsymbol{x}-b_i-\boldsymbol{f^T_i}\boldsymbol{\theta}\geq -r_i C.
\end{align}
Here the nonlinear complementary slackness conditions are reformulated with the binary variable $r_i$ and the large non-binding constant $C$ for each $i=1,\ldots,p$. Therefore, problem \eqref{appen:equ.lp.kkt1}--\eqref{appen:equ.lp.kkt6} can be reformulated to be the mixed-integer linear program (MILP).

\rev{We remark that if $\nu^{f*}=0$, implying that the system is too binding, e.g., for DC-OPF problem, some line/generator must always be at its capacity upper bound. Such a restrictive condition seldom happens in practice for the power system safety operation. Under such a scenario, one can consider a smaller input region $\mathcal{D}$ such that the input is not so extreme and there could always exist an interior for the input region.}

\section{Minimal Supporting Calibration Region}\label{appen:sec.mscr}
We first provide a toy example to demonstrate the non-uniqueness of the minimal supporting calibration region defined in Def.~\ref{def:mimial.region}. Consider the following modified network flow problem:
\begin{align}
    \min \quad & x^2_1+x^2_2+x^2_3 \\
    \mathrm{s.t.} \quad & 0\leq x_1\leq 90,\\
    & 0\leq x_2\leq 90,\\
    & x_3\leq 70,\label{appen:equ.ms1}\\
     & x_1+x_2\leq 90,\label{appen:equ.ms2}\\
     & x_2+x_3\leq 90,\label{appen:equ.ms3}\\
     & x_1+x_2+x_3=l.
\end{align}
Here $l$ is the input load within [$0,100$] and $x_1$, $x_2$, and $x_3$ can be seen as the network flow on the edges. Similar to the analysis in Sec.~\ref{ssec:calibrationrange}, the constraints \eqref{appen:equ.ms1}--\eqref{appen:equ.ms3} can be calibrated by at most $37.5\%$ uniformly. However, such a calibration region is not the minimal one, while forms the outer bound of it. Denote the calibration rate on \eqref{appen:equ.ms1}--\eqref{appen:equ.ms3} as ($x,y,z$), it is easy to see that any combination such that $7x+9y=6$ and $z=8/9-y$ is the minimal supporting one.

We further provide the follow procedures to determine (one of) the minimal supporting region.
\begin{itemize}
    \item  Step 1. Solve \eqref{equ:calibration.rate-1}--\eqref{equ:calibration.rate-2} to obtain the uniform maximum calibration rate $\Delta$. Let $k=1$.
    \item Step 2. For constraint $g_k$, solve
\begin{align}
    \min_{\boldsymbol{\theta}} \max_{{\boldsymbol{x}}}  \  &\frac{\hat{e}_k-g_k(\boldsymbol{\theta},\boldsymbol{x})}{|e_j|} \label{appen:equ.mc.steps1} \\
    \mathrm{s.t.} \quad & \eqref{equ:OPCC.formulation-4}, \ \ \boldsymbol{\theta}\in \mathcal{D},\notag\\ &g_j(\boldsymbol{\theta},\boldsymbol{x})\leq \hat{e}_j, \forall j\in \mathcal{E}, \label{appen:equ.mc.steps2}
\end{align}
where $\hat{e}_k=e_k\cdot(1_{e_k\geq0}(1-\Delta)+1_{e_k<0}(1+\Delta))$. Denote the optimal value of \eqref{appen:equ.mc.steps1}--\eqref{appen:equ.mc.steps2} as $\delta_k$, which represent the maximum additional individual calibration rate of constraint $g_k$ considering all other constraints' calibrations.
\item Update $\hat{e}_k$ to be $e_k\cdot(1_{e_k\geq0}(1-\Delta-\delta_k)+1_{e_k<0}(1+\Delta+\delta_k))$ and proceed to the next iteration $k+1$. Go to Step 2.
\end{itemize}
We remark that after each update of $\hat{e}_k$, the next $g_{k+1}$ is studied on a tighter region described by $\{\hat{e}_j, j=1,\ldots,k\}$. After solving the programs for each $g_k$,  one can easily see that the calibration region $\{\Delta+\delta_j\}_{j\in\mathcal{E}}$ is the minimal supporting calibration region.

\section{Mixed-Integer Reformulation of Maximum Violation and Proof of Proposition~\ref{prop:violation.vs.calibration}}
\label{appen:proof.of.calibration}
We first provide the mixed-integer linear reformulation of \eqref{equ:DNNsize-2} as follows. Consider
$${\nu}^f=\max_{j\in\mathcal{E}}\{(g_j(\boldsymbol{\theta},\hat{\boldsymbol{x}})- \hat{e}_j)/|e_j|\}.$$
The element-wise maximum in the objective can be reformulated to be the set of mixed-integer constraints:
\begin{align}
    & {\nu}^f\geq \frac{g_j(\boldsymbol{\theta},\hat{\boldsymbol{x}})- \hat{e}_j}{|e_j|}, \ \forall j=1,\ldots,|\mathcal{E}|,\\
    &{\nu}^f\leq \frac{g_j(\boldsymbol{\theta},\hat{\boldsymbol{x}})- \hat{e}_j}{|e_j|}+C\cdot(1-b_j), \ \forall j=1,\ldots,|\mathcal{E}|,\label{appen:equ.binary.max}\\ 
    & b_j\in\{0, 1\}, \ \forall j=1,\ldots,|\mathcal{E}|,\\
    & \sum^{|\mathcal{E}|}_{j=1}b_j=1. 
\end{align}

In \eqref{appen:equ.binary.max}, $b_j, j=0,\ldots,|\mathcal{E}|$ are binary variables that indicate the maximum among $g_j(\boldsymbol{\theta},\hat{\boldsymbol{x}})- \hat{e}_j)/|e_j|$ (e.g., $b_k=1$ if the violation on $g_k$ is the maximum one) and $C$ can be set as some big number.

We further provide the proof of Proposition~\ref{prop:violation.vs.calibration}.

\textsf{Proof:} Consider the DNN with $N_{\text{hid}}$ hidden layers each having $N_{\text{neu}}$ neurons and parameters $(\mathbf{W}^{f}, \bold{b}^{f})$ and $\rho \leq \Delta$. 
Since $\rho$ is the obtained optimal objective value of the bi-level problem \eqref{equ:DNNsize-1}--\eqref{equ:DNNsize-2}, we have
\begin{align}
    (g_j(\boldsymbol{\theta},\hat{\boldsymbol{x}})- \hat{e}_j)/|e_j|\leq \rho, \forall \boldsymbol{\theta}\in\mathcal{D}, \forall j\in\mathcal{E}.
\end{align}
Therefore, we have for any $\boldsymbol{\theta}\in\mathcal{D}$ and $j\in\mathcal{E}$
\begin{equation}
     \begin{cases}
          g_j(\boldsymbol{\theta},\hat{\boldsymbol{x}})-e_{j}(1-\Delta)\leq \rho\cdot e_j, & \mbox{if }e_{j}\geq0;\\
      g_j(\boldsymbol{\theta},\hat{\boldsymbol{x}})-e_{j}(1+\Delta)\leq -\rho\cdot e_j, & \mbox{otherwise},
    \end{cases}
\end{equation}
which is equivalent to 
\begin{equation}
     \begin{cases}
          g_j(\boldsymbol{\theta},\hat{\boldsymbol{x}})\leq e_j+(\rho-\Delta)\cdot e_j, & \mbox{if }e_{j}\geq0;\\
      g_j(\boldsymbol{\theta},\hat{\boldsymbol{x}})\leq e_j+(\Delta-\rho)\cdot e_j, & \mbox{otherwise}.
            \end{cases}
\end{equation}
Since $\rho\leq\Delta$, we have
\begin{align}
    g_j(\boldsymbol{\theta},\hat{\boldsymbol{x}})\leq e_j, \forall \boldsymbol{\theta}\in\mathcal{D}, \forall j\in\mathcal{E}.
\end{align}
This completes the proof of Proposition~\ref{prop:violation.vs.calibration}.

\section{Details of Applying \textit{Danskin's Theorem} to the Bi-Level Problem to Determine the Sufficient DNN Size}\label{appen:sec.danskin}

We provide the details of applying \textit{Danskin's Theorem} to solve the bi-level mined-integer nonlinear problem \eqref{equ:DNNsize-1}--\eqref{equ:DNNsize-2} and discuss the relationship between the obtained solution and the global optimal one for general OPCC.

To solve such bi-level optimization problem, we optimize the upper-level variables $(\mathbf{W},\bold{b})$ by gradient descent. This would simply involve repeatedly computing the gradient w.r.t. $(\mathbf{W},\bold{b})$ for the object function, and taking a step in this negative direction. That is, we want to repeat the update
\begin{align}\label{equ.gradientw}
   \mathbf{W} \coloneqq \mathbf{W}-\alpha\cdot\nabla_{\mathbf{W}} (\max_{\boldsymbol{\theta}} {\nu}^f(\mathbf{W}, \mathbf{b}, \boldsymbol{\theta})),
\end{align}
\begin{align}\label{equ.gradientb}
    \mathbf{b} \coloneqq \mathbf{b}-\alpha\cdot\nabla_{\mathbf{b}} (\max_{\boldsymbol{\theta}} {\nu}^f(\mathbf{W}, \mathbf{b}, \boldsymbol{\theta})).
\end{align}
Here $\max_{\boldsymbol{\theta}} {\nu}^f(\mathbf{W}, \mathbf{b}, \boldsymbol{\theta})$ denotes the maximum violation among the calibrated inequality constraints within the entire inputs domain $\mathcal{D}$, given the specific value of DNN parameters $(\mathbf{W},\bold{b})$.  Note that the inner function itself contains a maximization problem. We apply the \textit{Danskin's Theorem} to compute the gradient of the inner term. It states that the gradient of the inner function involving the maximization term is simply given by the gradient of the function evaluated at this maximum. In other words, to compute the (sub)gradient of a function containing a $\max(\cdot)$ term, we need to simply: 1) find the maximum, and 2) compute the normal gradient evaluated at this point~\cite{danskin,danskin2012theory}.
Hence, the relevant gradient is given by
\begin{align}\label{equ.danskinw}
    \nabla_{\mathbf{W}} (\max_{\boldsymbol{\theta}} {\nu}^f(\mathbf{W}, \mathbf{b}, \boldsymbol{\theta}))=\nabla_{\mathbf{W}} {\nu}^f(\mathbf{W}, \mathbf{b}, \boldsymbol{\theta}^*),
\end{align}
\begin{align}\label{equ.danskinb}
    \nabla_{\mathbf{b}} (\max_{\boldsymbol{\theta}} {\nu}^f(\mathbf{W}, \mathbf{b}, \boldsymbol{\theta}))=\nabla_{\mathbf{b}} {\nu}^f(\mathbf{W}, \mathbf{b}, \boldsymbol{\theta}^*),
\end{align}
where 
\begin{align}\label{equ.optimalx}
    \boldsymbol{\theta}^*=\arg\max_{\boldsymbol{\theta}} {\nu}^f(\mathbf{W}, \mathbf{b}, \boldsymbol{\theta}).
\end{align}
Here the optimal $\boldsymbol{\theta}^*$ depends on the choice of DNN parameters $(\mathbf{W},\bold{b})$. 
Therefore, at each iterative update of $(\mathbf{W},\bold{b})$, we need to solve the inner maximization problem once.  {Note that the optimal $\boldsymbol{\theta}^*$ may not be unique. However, the gradient of ${\nu}^f(\mathbf{W}, \mathbf{b}, \boldsymbol{\theta}^*)$ w.r.t. $(\mathbf{W}, \mathbf{b})$ can still be obtained given a specific $\boldsymbol{\theta}^*$, which is (one of the) gradient that optimizes the deep neural network. We remark that such approach is indeed widely adopted in existing literature~\cite{danskin,danskin2012theory}. In addition, though the involved program is a mixed-integer linear problem, we observe that the solver can indeed provide its optimum efficiently, e.g., $<$20 mins for Case300 in DC-OPF problem in simulation. Nevertheless, we remark that finding a (sub-optimal) feasible solution for the inner maximization problem can be easily obtained by a heuristic trial of some particular $\boldsymbol{\theta}$, e.g., the worst-case input at the previous round as the initial point and the associate integer values in the DNN constraints~(\ref{equ.DNNinteger-1})-(\ref{equ.DNNinteger-2}), which are fixed given the specification of DNN parameters. Such a solution can still be utilized for the further steps to calculate the sub-gradient of the DNN. One can see the analogy between it and DNN training with stochastic gradient decent method. 

In addition, note that to obtain the upper bound $\rho$, we do not need to access any feasible point of the inner maximization problem. The upper bound is provided by the relaxation in the branch-and-bound algorithm, e.g., relax (some) integer variables to continuous. This can be efficiently obtained by the solvers, e.g., APOPT or Gurobi. Such an upper bound is applied to verify whether universal feasibility guarantee is obtained and whether the DNN size is sufficient.}

For more discussion, at each iteration, we use $\nu^{f,t}$ to denote the corresponding objective value ${\nu}^f(\mathbf{W}, \mathbf{b}, \boldsymbol{\theta}^*)$. {We remark that if the value of $\nu^{f,t}-\Delta$ is non-positive after some number of iterations for some DNN size, then the evaluated DNN size is capable of achieving universal feasibility w.r.t. the entire input domain $\mathcal{D}$.  Otherwise if the value of $\nu^{f,t}-\Delta$ is always positive after $t$-th iteration with a large number of iterations $t$ for some DNN size, the evaluated DNN size may not be able to preserve universal feasibility.  Therefore, we need to increase the DNN size for better approximation ability.

It is worth noticing that the above result is based on the condition that we can obtain the global optimal solution of \eqref{equ:DNNsize-1}--\eqref{equ:DNNsize-2}. However, one should note that for general OPCC 1) the inner maximization of \eqref{equ:DNNsize-1}--\eqref{equ:DNNsize-2} is indeed a non-convex mixed-integer nonlinear program due to the ReLU activations and the non-linear inequalities associated with \eqref{equ:DNNsize-2}. The existing solvers, e.g., APOPT, YALMIP, or Gurobi, may not be able to provide the global optimal solution, meaning that for the given parameters of DNN $(\mathbf{W},\bold{b})$, we actually obtain a lower bound on the maximum violation among all possible inputs $\boldsymbol{\theta}\in\mathcal{D}$; 2) the iterative approach in \eqref{equ.gradientw}--\eqref{equ.optimalx} updating the DNN parameters of the outer problem characterizes the upper bound on such lower bound on the maximum violation from the inner problem given the DNN size. That is, for example if we can always solve the inner problem global optimally, the obtained value $\nu^{f,t}$ is the upper bound on $\nu^{f*}$, the optimal objective of \eqref{equ:DNNsize-1}--\eqref{equ:DNNsize-2}. If the inner problem only provides a lower bound on $\nu^{f*}|_{(\mathbf{W},\bold{b})}$, the optimal objective of inner problem given the specification of DNN parameters $(\mathbf{W},\bold{b})$, then the value of $\nu^{f,t}$ constructs the upper-lower bound on $\nu^{f*}$. Though such a bound might not be tight, it indicates that it could be possible to achieve universal feasibility with such a DNN size if $\nu^{f,t}-\Delta\leq0$. Otherwise if for some DNN size, the value of $\nu^{f,t}-\Delta$ is always positive, then such evaluated DNN size may fail to guarantee universal feasibility.}

\subsection{Determining the values of $h^{\max,k}_i/h^{\min,k}_i$}\label{appen.ssec.upper.lower.neuron}

 $h^{max,k}_i/h^{min,k}_i$ are constants and fixed during solving the (inner) MILP in optimization (\ref{equ:DNNsize-1})-(\ref{equ:DNNsize-2})~\cite{tjeng2018evaluating}. These numbers represent the maximum/minimum bounds on the values of the neuron outputs, which should be large/small enough numbers to let the DNN constraints not be binding in the reformulation (\ref{equ.DNNinteger-1})-(\ref{equ.DNNinteger-2}). In our design, we follow the technique in~\cite{venzke2020learning} to obtain such (tighter) upper/lower bounds for each updated $(\mathbf{W},\bold{b})$. In particular, we minimize and maximize the output of each neuron subject to the linear relaxation of the binary variables (to be continuous within 0 and 1) in the DNN constraints with parameters $(\mathbf{W},\bold{b})$ in (\ref{equ.DNNinteger-1})-(\ref{equ.DNNinteger-2}) and entire input region $\mathcal{D}$. Such upper/lower bounds can be efficiently obtained by solving the LPs after relaxation, which guarantees that the neuron output will not exceed the corresponding values. We note that for different DNN parameters $(\mathbf{W},\bold{b})$, $h^{max,k}_i/h^{min,k}_i$ could take different values that can always be efficiently obtained from the LPs after linear relaxation.

\section{Proof of Proposition~\ref{prop:infinity}}\label{appendix.prop3}
\textsf{Proof idea:} Here we consider the post-trained DNN with $N_{\text{hid}}$ hidden layers each having $N^*_{\text{neu}}$ neurons. Given current iteration $i$, for $ \forall j\leq i$, suppose it can always maintain feasibility at the correspondingly constructed neighborhoods around the identified worst-case input, i.e., $\hat{\mathcal{D}}^j$, by training on $\mathcal{T}^{i+1}$ that combines $\mathcal{T}^0$ and all the auxiliary subset $\mathcal{S}^j$ around the identified adversarial input $\boldsymbol{\theta}^j, \forall j\leq i$. Therefore, when the number of iterations is large enough, the union of the feasible regions $\tilde{\mathcal{D}}^{i>C}=\hat{\mathcal{D}}^1\cup\hat{\mathcal{D}}^2\cup\ldots\hat{\mathcal{D}}^i$ can cover the entire input domain $\mathcal{D}$. That is, the post-trained DNN can ensure feasibility for each small region $\hat{\mathcal{D}^i}$ within the input domain $\mathcal{D}$, and hence universal feasibility is guaranteed. Such observation is similar to the topic of minimum covering ball problem of the compact set in real analysis.

Such a condition generally requires the DNN to preserve feasibility within some small regions by especially including the input-solution information during training, which may not be hard to satisfy. This can be understood from the observation that the worst-case violation in the smaller inner domain can be reduced significantly by training on the broader outer input domain~\cite{venzke2020learning,nellikkath2021physics2} as the adversarial inputs are always element-wise at the boundary of the entire input domain $\mathcal{D}$, which echoes our simulation findings in Sec.~\ref{sec:simulations}. Therefore, the post-trained DNN is expected to perform good feasibility guarantee in all small regions $\hat{\mathcal{D}}^j, \forall j\leq i$ after the preventive training procedure on $\mathcal{T}^{i+1}$, the training set on the entire domain $\mathcal{D}$. We remark that after gradually including these subsets $\mathcal{S}^i$ into the existing training set, the loss function is determined by the joint loss among all samples in these regions. After the training process, the post-obtained DNN is hence expected to maintain feasibility at the points in the training set and the regions around them.

\section{Implementations of \textsf{DeepOPF+}}\label{appendix.deepopf+}
Recall that the DC-OPF formulation is given as
\begin{align}
    \min_{\boldsymbol{P_{G}},\ {\Phi }} &  \mathrm{\ }\sum_{i\in\mathcal{G}}{c_i\left( P_{Gi} \right)} \\
    \mathrm{s.t.} \ \ &\boldsymbol{P_{G}^{\min}}\le \boldsymbol{P_{G}}\le \boldsymbol{P_{G}^{\max}},\label{appen:equ:pg}\\
    & \mathbf{M}\cdot \Phi =\boldsymbol{P_{G}}-\boldsymbol{P_{D}}, \label{appen:equ:branch.balance}\\
    & -\boldsymbol{P^{\max}_{\text{line}}} \leq\mathbf{B}_{\text{line}}\cdot \Phi \leq \boldsymbol{P^{\max}_{\text{line}}}. \label{appen:equ:branch.limit}
\end{align}
We first reformulate the DC-OPF to remove the linear equality constraints and reduce the number of decision variables without losing optimality by adopting the predict-and-reconstruct framework~\cite{deepopf1}. Specifically, it leverages that the admittance matrix (after removing the entries corresponding to the slack bus) $\tilde{\mathbf{M}}$ is of full rank $B-1$, where $B=|\mathcal{B}|$ and is the size of the set of buses. Thus, given each $\boldsymbol{P_D}$, once the non-slack generations $\{P_{Gi}\}_{i\in\mathcal{G}\backslash n_0}$ ($n_0$ denotes the slack bus index) are determined, the slack generation and the bus phase angles of all buses can be uniquely reconstructed:
\begin{align}
    &P^{\text{slack}}_G=\sum_{i\in\mathcal{B}} P_{Di}- \sum_{i\in\mathcal{G}\backslash n_0} {P}_{Gi},\label{appen:equ.slackgeneration}\\
&\tilde{\Phi}= \mathbf{\tilde{M}}^{-1}\left( {\boldsymbol{\tilde{P}}_{{G}}}-\boldsymbol{\tilde{P}}_{{D}} \right),
\label{equ.DCOPF.reformulate} 
\end{align}
\noindent where $n_0$ and $P^{\text{slack}}_G$ denote the slack bus index and slack bus generation respectively. $\boldsymbol{\tilde{P}}_{{G}}$ and $\boldsymbol{\tilde{P}}_{{D}}$ are the $(B-1)$-dimensional generation and load vectors for all buses except the slack bus. Consequently, the line flow capacity constraints in (\ref{appen:equ:branch.limit}) can be reformulated as
\begin{equation}
-\boldsymbol{P^{\max}_{\text{line}}}\leq \tilde{\mathbf{B}}_{\text{line}}\mathbf{\tilde{M}}^{-1}\left( {\boldsymbol{\tilde{P}}_{{G}}}-\boldsymbol{\tilde{P}}_{{D}} \right)\leq \boldsymbol{P^{\max}_{\text{line}}},
\label{equ.PTDF.reformulate} 
\end{equation}
where $\tilde{\mathbf{B}}_{\text{line}}$ is the line admittance matrix after removing the column of slack bus.\footnote{{The matrix $\tilde{\mathbf{B}}_{\text{line}}\mathbf{\tilde{M}}^{-1}$ is well-known as ``Power Transfer Distribution Factors'' (PTDF) matrix~\cite{chatzivasileiadis2018optimization}.}} Therefore, the reformulated DC-OPF problem takes the form of
\begin{align}
    \min_{\boldsymbol{{\tilde{P}}_{{G}}}} &  \mathrm{\ }\sum_{i\in\mathcal{G}\backslash n_0}{c_i\left( P_{Gi} \right)}+c_{n_0}\left( \sum_{i\in\mathcal{B}} P_{Di}- \sum_{i\in\mathcal{G}\backslash n_0} {P}_{Gi} \right) \\
    \mathrm{s.t.} \ \ & (\ref{equ.PTDF.reformulate}), \notag\\ &P_{Gi}^{\min}\le P_{Gi}\le P_{Gi}^{\max}, \forall i\in \mathcal{G}\backslash n_0, \label{appen:equ.DCOPFmaxi-violation-pg}\\
    & P^{\min}_{\text{slack}}\leq\sum_{i\in\mathcal{B}} P_{Di}- \sum_{i\in\mathcal{G}\backslash n_0} {P}_{Gi}\leq P^{\max}_{\text{slack}}.\label{appen:equ:slack}
\end{align}
Therefore, we can solve DC-OPF by employing DNNs to depict the mapping between $\boldsymbol{P_D}$ and $\boldsymbol{{\tilde{P}}_{{G}}}$. We further note that any feasible active power generation $P_{Gi}$ that satisfies \eqref{equ:pg} can be written as
\begin{equation}
P_{Gi}=P_{Gi}^{\min}+\alpha _i\cdot\left( P_{Gi}^{\max}-P_{Gi}^{\min} \right),\ \alpha_i\in \left[ 0,1 \right] ,i\in\mathcal{G}.
\label{equ.DCOPF.scaling}
\end{equation}
Similar to~\cite{deepopf1}, instead of predicting $\{P_{Gi}\}_{i\in\mathcal{G}\backslash n_0}$, we use DNNs to generate the corresponding scaling factors and reconstruct the $\{P_{Gi}\}_{i\in\mathcal{G}\backslash n_0}$ and remaining variables in implementation. {Here one can apply the two clamp-equivalent actions in \eqref{dnn.model} or the equivalent Sigmoid function $\sigma'(x)=\frac{1}{1+e^{-x}}$ at the output layer of DNNs to predict the (0,1) scaling factors such that the feasibility of predicted $P_{Gi}, i\in\mathcal{G}\backslash n_0$ can always be guaranteed. The Sigmoid functions at the output layer present the same effect of the last two clipped ReLU operations in \eqref{dnn.model} to ensure feasibility of the predicted variables.}
\subsection{Removing Non-Critical Inequality Constraints}\label{appen:ssec.remove}
\subsubsection{Removing Non-Critical Branch Limits}

\begin{figure} [!t]
  \centering
  \subfigure[Case30.]{
    \label{appen:fig.case30.violation} 
    \includegraphics[width = 0.3\textwidth]{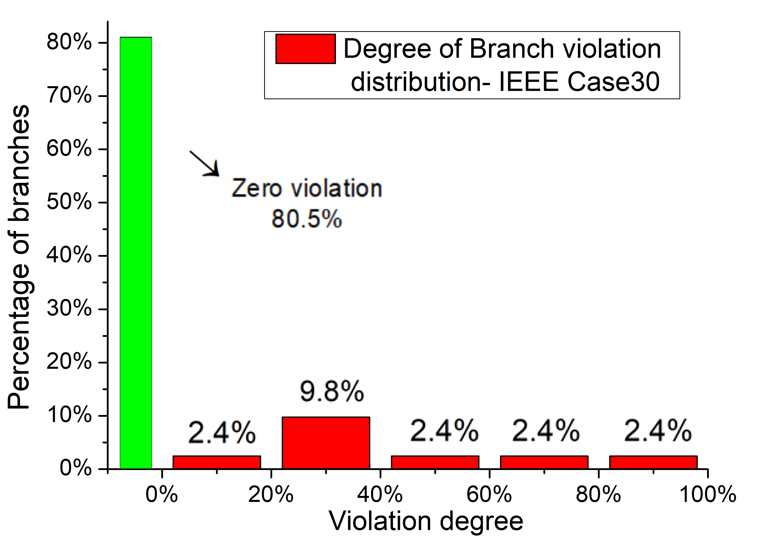}
  }
  \subfigure[Case118.]{
    \label{appen:fig.case118.violation} 
    \includegraphics[width = 0.3\textwidth]{./figs/case30-v2.png}
  }
  \subfigure[Case300.]{
    \label{appen:fig.case300.violation} 
    \includegraphics[width = 0.3\textwidth]{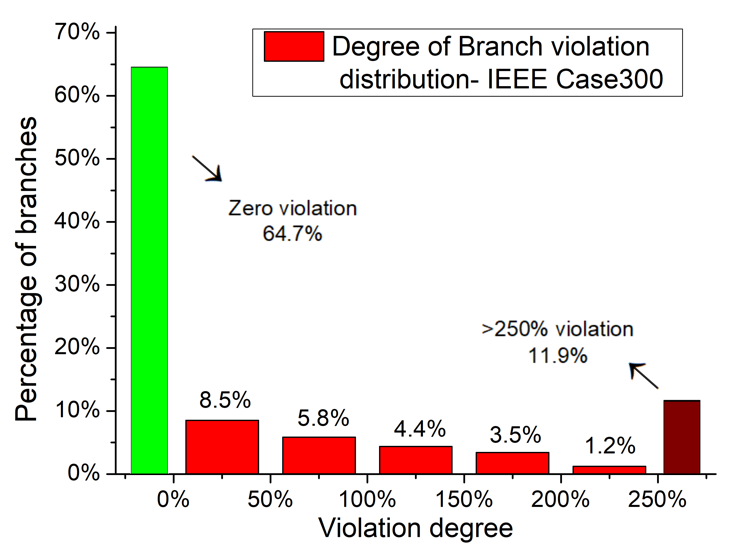}
  }
  \caption{Percentage and distribution of critical/non-critical transmission line of IEEE Case30, Case118, and Case300.}\label{appen:fig.violation}
\end{figure}

We propose the following program for each branch $i$ to remove the non-critical branch limits given the entire load and generation space:
\begin{align}
    &\max_{\boldsymbol{{\tilde{P}}_{{G}}}, \boldsymbol{P_D}} \quad \nu_i-1 \label{appen:equ.DCOPFmaxi-violation-obj}\\
    &\quad \mathrm{s.t.} \quad \ (\ref{appen:equ.DCOPFmaxi-violation-pg}),\notag\\
    &  \ \quad\quad \quad \ \boldsymbol{P_D}\in\mathcal{D},\label{appen:equ.DCOPFmaxi-violation-pd}\\
    &\ \quad\quad \quad \ \boldsymbol{\nu}={|\tilde{\mathbf{X}}\left( {\boldsymbol{\tilde{P}}_{{G}}}-\boldsymbol{\tilde{P}}_{{D}} \right)|}\label{appen:equ.DCOPFmaxi-violation-branch}.
\end{align}
Here we assume the load domain $ \mathcal{D}=\{\boldsymbol{P_D}|\mathbf{A}_d\boldsymbol{P_D}\leq \boldsymbol{b_d}, \exists \boldsymbol{{\tilde{P}}_{{G}}}: (\ref{equ.PTDF.reformulate}), (\ref{appen:equ.DCOPFmaxi-violation-pg}),(\ref{appen:equ:slack}\ \text{hold}\})$ is restricted to a convex polytope described by matrix $\boldsymbol{A_d}$ and vector $\boldsymbol{b_d}$ and the corresponding constraints.
(\ref{appen:equ.DCOPFmaxi-violation-pg}) enforces the feasibility of non-slack generations. (\ref{appen:equ.DCOPFmaxi-violation-branch}) represents the normalized power flow level at each branch, where $\tilde{\mathbf{X}}$ is obtained from (\ref{equ.PTDF.reformulate}) by dividing each row of matrix $\tilde{\mathbf{B}}_{\text{line}}\mathbf{\tilde{M}}^{-1}$ with the value of corresponding line capacity  and $\boldsymbol{\nu}\in \mathcal{R}^{|\mathcal{E}|}$.

We remark that problem (\ref{appen:equ.DCOPFmaxi-violation-obj})-(\ref{appen:equ.DCOPFmaxi-violation-branch}) can be reformulated as two linear programmings to perform the inference of the absolute sign of power flows in (\ref{appen:equ.DCOPFmaxi-violation-branch}):
\begin{align}
    \max_{\boldsymbol{{\tilde{P}}_{{G}}}, \boldsymbol{P_D}} / \min_{\boldsymbol{{\tilde{P}}_{{G}}}, \boldsymbol{P_D}} \quad &\tilde{\nu}_i \label{appen:equ.DCOPFmaxi-violation-minmaxobj}\\
    \quad \mathrm{s.t.} \quad\quad &(\ref{appen:equ.DCOPFmaxi-violation-pg}), (\ref{appen:equ.DCOPFmaxi-violation-pd}),\notag\\ 
    &\tilde{\boldsymbol{\nu}}={\tilde{\mathbf{X}}\left( {\boldsymbol{\tilde{P}}_{{G}}}-\boldsymbol{\tilde{P}}_{{D}} \right)}\label{appen:equ.DCOPFmaxi-violation-abs}.
\end{align}
If the optimal value of the above maximization (respectively minimization) problem is smaller or equal (respectively greater or equal) than 1 (respectively -1), then the optimal value of (\ref{appen:equ.DCOPFmaxi-violation-minmaxobj})-(\ref{appen:equ.DCOPFmaxi-violation-abs}) is non-positive for some branch $i$. Therefore, such non-critical inequality constraint does not affect the feasible solution space such that it is always respected given any input load $\boldsymbol{P_D}$ and can be removed from the DC-OPF problem.
By solving (\ref{appen:equ.DCOPFmaxi-violation-minmaxobj})-(\ref{appen:equ.DCOPFmaxi-violation-abs}), we can derive the set ${\mathcal{E}}$ of critical branch capacity constraints whose optimal objectives are positive.\footnote{For the critical branch constraints not in ${\mathcal{E}}$, it is possible to encounter such load input and generation solution profiles using the DNN scheme with the upper/lower bounds clipped ReLU functions in (\ref{dnn.model}) or the Sigmoid function at output layer under the worst-case scenarios with which the power flow on branch $i$ exceeds its transmission limit.}


The percentage and distribution of the critical/non-critical transmission lines in IEEE Case30, Case118, and Case300 are shown in Fig.~\ref{appen:fig.case30.violation}, Fig.~\ref{appen:fig.case118.violation}, and Fig.~\ref{appen:fig.case300.violation} respectively.  We observe that $80.5\%$, $76.9\%$ and $64.7\%$ of line limits in IEEE Case30, Case118, and Case300 are always inactive even under the worst-case scenarios.

\subsubsection{Removing Non-Critical Slack Bus Generation Limits}

We provide the formulation to identify the critical slack generation limits given the entire load and generation space and the possible violation degree w.r.t. the upper and lower bounds here.
\begin{align}
    &\max_{\boldsymbol{{\tilde{P}}_{{G}}}, \boldsymbol{P_D}} \quad {\nu}^u_{\text{slack}} \label{appen:equ.DCOPFmaxi-violation-slackup-obj}\\
    &\quad\mathrm{s.t.}
    \quad (\ref{appen:equ.slackgeneration}), (\ref{appen:equ.DCOPFmaxi-violation-pg}), (\ref{appen:equ.DCOPFmaxi-violation-pd}),\notag\\
    &\quad\ \quad\quad{\nu}^u_{\text{slack}}=\frac{P^{\text{slack}}_G-P^{\max}_{\text{slack}}}{P^{\max}_{\text{slack}}-P^{\min}_{\text{slack}}},\label{appen:equ.DCOPFmaxi-violation-slackup}
\end{align}
and 
\begin{align}
    &\max_{\boldsymbol{{\tilde{P}}_{{G}}}, \boldsymbol{P_D}} \quad {\nu}^l_{\text{slack}} \label{appen:equ.DCOPFmaxi-violation-slacklower-obj}\\
    &\quad\mathrm{s.t.}
    \quad (\ref{appen:equ.slackgeneration}), (\ref{appen:equ.DCOPFmaxi-violation-pg}), (\ref{appen:equ.DCOPFmaxi-violation-pd}),\notag\\
    &\quad\ \quad\quad\ {\nu}^l_{\text{slack}}=\frac{P^{\min}_{\text{slack}}-P^{\text{slack}}_G}{P^{\max}_{\text{slack}}-P^{\min}_{\text{slack}}},\label{appen:equ.DCOPFmaxi-violation-slacklower}
\end{align}
respectively. 
Here \eqref{appen:equ.DCOPFmaxi-violation-slackup} and \eqref{appen:equ.DCOPFmaxi-violation-slacklower} denote the (normalized) exceeding of slack bus generation exceeding its upper bound and lower bound, respectively. Therefore, if the optimal values of these proposed optimization problem is non-positive, such slack generation limit is non-critical and does not affect the load-solution feasible region.

We remark that problems \eqref{appen:equ.DCOPFmaxi-violation-slackup-obj}--\eqref{appen:equ.DCOPFmaxi-violation-slackup}, and \eqref{appen:equ.DCOPFmaxi-violation-slacklower-obj}--\eqref{appen:equ.DCOPFmaxi-violation-slacklower} are indeed linear programs and can be efficiently solved by the state-of-the-art solvers such as CPLEX or Gurobi. We find that all three test cases could have both critical upper bound and lower bound limits, i.e., both \eqref{appen:equ.DCOPFmaxi-violation-slackup-obj}--\eqref{appen:equ.DCOPFmaxi-violation-slackup} and \eqref{appen:equ.DCOPFmaxi-violation-slacklower-obj}--\eqref{appen:equ.DCOPFmaxi-violation-slacklower} have positive optimal values. The (normalized) exceeding degrees on slack bus generation limits and the percentage of critical limits among all inequalities for the three test cases are reported in Table~\ref{appen:tab.critical}.

\begin{table}[!t]
\caption{Relative slack bus generation limits exceeding and the percentage of critical inequality constraints.}
	\renewcommand{\arraystretch}{1.1}
	\centering
	\begin{tabular}{c|c|c|c}
		\toprule
		\hline
        \multicolumn{1}{c|}{Variants} & \tabincell{c}{IEEE Case30} & \tabincell{c}{IEEE Case118} & IEEE Case300\\
        \hline
        \multirow{1}{*}{\tabincell{c}{Upper bound exceeding}} &\tabincell{c}{2.1\%} &3.7\% &7.5\% \\
        \cline{2-4}
        \hline
        \multirow{1}{*}{\tabincell{c}{Lower bound exceeding}} &\tabincell{c}{0.8\%} &0.9\% &6.1\% \\
        \cline{2-4}
		\hline
		\multirow{1}{*}{\tabincell{c}{Percentage of critical constraints}} &\tabincell{c}{23.3\%} &23.9\% &35.6\% \\
        \cline{2-4}
		\hline
		\bottomrule		
		\end{tabular}
	\label{appen:tab.critical}
\end{table}

\subsection{Maximum Constraints Calibration Rate}\label{appen:ssec.calibration.dcopf}

Recall that in \textsf{DeepOPF+}, the DNN is trained on the samples from OPF problems with adjusted constraints. However, if we reduce the limits too much, some load $P_D\in\mathcal{D}$ will become infeasible under the calibrated constraints and hence lead to invalid training data with poor generalization, though they are feasible for the original limits. Therefore, it is critical to determine the appropriate calibration range without reducing the supported load feasible region. We propose the following bi-level optimization program to determine the maximum constraints calibration rate while preserving the input region:
\begin{align}
    &\min_{\boldsymbol{P_D}} \max_{\boldsymbol{P_G}}  \  \nu^{c} \label{appen:equ.DCOPFcalibration-obj}\\
    &\ \mathrm{s.t.}
    \quad\quad \  \eqref{equ:pg}, \ \eqref{appen:equ.slackgeneration}-\eqref{equ.PTDF.reformulate}, \ \eqref{appen:equ.DCOPFmaxi-violation-pd}, \notag\\
    &\quad\quad\quad\quad|PF_{ij}|=|\frac{1}{r_{ij}}\left( \phi_i-\phi_j \right)|,\,\, \forall\, (i,j)\in {\mathcal{E}}, \label{appen:equ.1}\\
    &\quad\quad\quad\quad{P}^u_{\text{slack}}=(P^{\max}_{\text{slack}}-P_{G}^{\text{slack}})/(P^{\max}_{\text{slack}}-P^{\min}_{\text{slack}}), \label{appen:equ.2}\\
    &\quad\quad\quad\quad{P}^l_{\text{slack}}=({P_{G}^{\text{slack}}-P^{\min}_{\text{slack}}})/(P^{\max}_{\text{slack}}-P^{\min}_{\text{slack}}), \label{appen:equ.3}\\
   &\quad\quad\quad\quad \nu^c\leq\frac{P_{Tij}^{\max}-|PF_{ij}|}{P_{Tij}^{\max}},\forall\, (i,j)\in {\mathcal{E}},\label{appen:equ.4}\\
  &\quad\quad\quad\quad \nu^c\leq{P}^u_{\text{slack}},\label{appen:equ.5}\\
 &\quad\quad\quad\quad \nu^c\leq {P}^l_{\text{slack}},\label{appen:equ.6}
\end{align}
where $PF_{ij}$ denotes the power flow on branch $(i,j)\in{\mathcal{E}}$. ${P}^u_{\text{slack}}$ and ${P}^l_{\text{slack}}$ represent the relative upper and lower bounds redundancy on slack bus.
Constraint \eqref{appen:equ.DCOPFmaxi-violation-pd} describes the convex polytope of $\boldsymbol{P_D}$. Constraints \eqref{equ:pg} and \eqref{appen:equ.slackgeneration}--\eqref{equ.DCOPF.reformulate} denote the feasibility of the corresponding $\boldsymbol{P_G}$. 
Consider the inner maximization problem, the objective finds the maximum of the element-wise least redundancy of the limits at ${\mathcal{E}}$, which is the largest possible constraints calibration rate at each given $P_D$. The outer minimization problem hence finds the largest possible calibration rate among all $P_D\in \mathcal{D}$, and correspondingly, the supported load feasible region is not reduced. We remark that the inner maximization problem is a linear program (LP).
as the set of inequalities containing the absolute operations on power flows $PF_{ij}$ in \eqref{appen:equ.4} can be reformulated to be linear. We employ the KKT-based approach in Sec.~\ref{ssec:calibrationrange} to solve the above bi-level problem and obtain the calibration rate for \textsf{DeepOPF+}. We remark that the above inner maximization problem is a primal feasible bounded LP (bounded feasible region of $\boldsymbol{P_G}$). Therefore, its dual problem is feasible and bounded, with strong duality hold. After solving \eqref{appen:equ.DCOPFcalibration-obj}--\eqref{appen:equ.6}, we derive the maximum calibration rate for each test case. Numerical results are summarized in Table~\ref{tab:violaton.calibration}.

\subsection{Constraints Calibration in DC-OPF Problem}

 In \textsf{DeepOPF+}, we adjust the system constraints preventively during the training stage. Therefore, even with approximation errors of DNN, the predicted solutions are anticipated to be feasible at the test stage. In particular, we first calibrate the system constraints, i.e., the critical transmission line capacity limits and slack bus generator's output limits, by an appropriate rate during the load sampling. As discussed in Appendix~\ref{appen:ssec.calibration.dcopf}, we reduce the line capacity limits by a certain rate $\eta_{ij}\geq0$, i.e.,
\begin{equation}\label{appen:equ.branchcalibrate}
|PF_{ij}|=|\frac{1}{r_{ij}}\left( \theta _i-\theta _j \right)| \le P_{Tij}^{\max}\cdot(1-\eta_{ij}),\,\, \forall\, (i,j)\in {\mathcal{E}},
\end{equation}
where $PF_{ij}$ and $r_{ij}$ are the power flow and line reactance at branch $(i,j)$ respectively. Set ${\mathcal{E}}$ contains the critical branch limits that need to be calibrated. We refer to Appendix~\ref{appen:ssec.remove} for the detailed construction of ${\mathcal{E}}$. The above formulation \eqref{appen:equ.branchcalibrate} is exactly the dedicated description of the second set of constraints describing the branch limits in \eqref{equ:branch.balance} for each branch $(i,j)\in{\mathcal{E}}$. The slack generation limits are also calibrated with $\xi\geq0$, i.e., \begin{equation}\label{appen:equ.slackcalibrate}
P^{\min}_{\text{slack}}+\xi\cdot k \leq P^{\text{slack}}_{G}\leq P^{\max}_{\text{slack}}-\xi\cdot k,
\end{equation}
where 
$P^{\min}_{\text{slack}}$ and $P^{\max}_{\text{slack}}$ are the slack bus generation limits and $k=P^{\max}_{\text{slack}}-P^{\min}_{\text{slack}}$. The choice of $\eta_{ij}$ and $\xi$ is analyzed in detail in Appendix~\ref{appen:ssec.calibration.dcopf}.   
 Then, we train the DNN on a dataset created with calibrated limits to learn the corresponding load-to-generation ($\boldsymbol{P_D}\mapsto \mathcal{S}$) mapping and evaluate its performance on a test dataset with the original limits. Thus, even with the inherent prediction error of DNN, the obtained solution can still remain feasible. {We remark that during the training stage, the operational limits calibration does not reduce the feasibility region of the load inputs $\boldsymbol{P_D}$ in consideration. The constraints calibration only leads to the (sub)optimal solutions that are interior points within the original feasible region (the operational constraints are expected to be not binding).}
Note that the slack generation and the voltage phase angles can be obtained from \eqref{equ.DCOPF.reformulate}--\eqref{equ.DCOPF.scaling} based on the predicted $\{\alpha_i\}_{i\in \mathcal{G}\backslash n_0}$ set-points. As benefits, the power balance equations in \eqref{equ:branch.balance} are guaranteed to be held, and the size of the DNN model and the amount of training data and time can be reduced.

\subsection{DNN Loss Function in DC-OPF Problem}
The target of DNN training is to determine the value of $\mathbf{W}$ and $\bold{b}$ which minimize the average of the specified loss function $\mathcal{L}_k$ among the training set, i.e.,
$$(\mathbf{W}^*,\bold{b}^*)=\arg\min_{\mathbf{W},\bold{b}}\frac{1}{|\mathcal{K}|}\sum^{|\mathcal{K}|}_{k=1}\mathcal{L}^k,$$
where $\mathcal{L}^k$ denotes the loss of training data $k$ and $|\mathcal{K}|$ is the number of training data.

In this work, we adopt the supervised learning approach in the \textit{Adversarial-Sample Aware} algorithm and design the loss function $\mathcal{L}$ consisting of two parts to guide the training process. Recall that similar to~\cite{deepopf1}, we first represent the feasible active power generation $P_{Gi}$ that satisfies \eqref{equ:pg} as:
$$
P_{Gi}=P_{Gi}^{\min}+\alpha _i\cdot\left( P_{Gi}^{\max}-P_{Gi}^{\min} \right),\ \alpha_i\in \left[ 0,1 \right] ,i\in\mathcal{G}.
$$
Therefore, instead of predicting $\{P_{Gi}\}_{i\in\mathcal{G}\backslash n_0}$, we use DNNs to generate the corresponding scaling factors and reconstruct the $\{P_{Gi}\}_{i\in\mathcal{G}\backslash n_0}$ and remaining variables in implementation. Hence, the first part is the sum of mean square error between the generated scaling factors $\hat{\alpha}_i$ and the reference ones ${\alpha}_i$ of the optimal solutions:
\begin{eqnarray}
\mathcal{L}_{P_G}=\frac{1}{|\mathcal{G}-1|}\sum_{i\in \mathcal{G}\backslash n_0}{\left( \hat{\alpha}_i-\alpha _i \right) ^2}.
\label{appen:equ.loss.first}
\end{eqnarray}
The second part consists of penalty terms related to the violations of the inequality constraints, i.e., line flow limits and slack bus generation:
\begin{equation}\label{appen:equ.loss.second}
\begin{split}
\mathcal{L}_{pen}=\mathcal{L}^{\text{line}}_{pen}+\mathcal{L}^{\text{slack}}_{pen},
\end{split}
\end{equation}
which are given as:
\begin{equation}
\begin{split}
\mathcal{L}^{\text{line}}_{pen}=&\frac{1}{|{\mathcal{E}|}}\sum_{k=1}^{|\mathcal{E}|}{\max \left( \left(\mathbf{X}\left( {\boldsymbol{\tilde{P}}_{{G}}}-\boldsymbol{\tilde{P}}_{{D}} \right) \right) _{k}^{2}-1,0 \right) },\\
\mathcal{L}^{\text{slack}}_{pen}=&\frac{1}{|{\mathcal{E}|}}\max\left(\frac{ P^{\text{slack}}_G-P^{\max}_{\text{slack}}}{P^{\max}_{\text{slack}}-P^{\min}_{\text{slack}}},0 \right)+\frac{1}{|{\mathcal{E}|}}\max\left(\frac{ P^{\min}_{\text{slack}}-P^{\text{slack}}_G}{P^{\max}_{\text{slack}}-P^{\min}_{\text{slack}}},0 \right),
\label{equ:loss.function.dcopf}
\end{split}
\end{equation}
respectively. Here matrix $\mathbf{X}$ is obtained from \eqref{equ.PTDF.reformulate} by dividing each row of matrix $\tilde{\mathbf{B}}_{\text{line}}\tilde{\mathbf{M}}^{-1}$ with the value of corresponding line capacity. The first and second terms of $\mathcal{L}^{\text{slack}}_{pen}$ denote (normalized) the violations of upper bound and lower bound on slack generation, respectively. 
We remark that after the constraints calibration, the penalty loss is with respect to the adjusted limits. Note here the non-slack generations are always feasible as we predict the $(0, 1)$ scaling factors in \eqref{equ.DCOPF.scaling}. 
The total loss is a weighted sum of the two:
\begin{eqnarray}
\mathcal{L}=w_1\cdot \mathcal{L}_{P_G}+w_2\cdot \mathcal{L}_{pen},
\label{appen:equ.loss.total}
\end{eqnarray}
\noindent where $w_1$ and $w_2$ are positive weighting factors representing the balance between prediction error and penalty. 
We apply the widely-used stochastic gradient descent (SGD) with momentum~\cite{qian1999momentum} method to update DNN's parameters $(\mathbf{W}, \bold{b})$ at each iteration.

\subsection{Run-time Complexity of \textsf{DeepOPF+}}

According to Sec.~\ref{ssec:complexity}, the computational complexity of \textsf{DeepOPF+} to predict the non-slack generations $\{P_{Gi}\}_{i\in\mathcal{G}\backslash n_0}$ is $\mathcal{O} \left(B^2\right)$. Reconstructing the phase angles $\Phi$ can be achieved by~(\ref{equ.DCOPF.reformulate}), which requires $\mathcal{O}\left(B^2 \right)$ operations. Overall, the computational complexity of \textsf{DeepOPF+} is $\mathcal{O}\left(B^2\right)$. For the traditional solver, the computational complexity of interior-point methods for solving DC-OPF is $\mathcal{O} \left( B^4\right)$, measured by the number of elementary operations. We remark that the computational complexity of \textsf{DeepOPF+} is lower than that of traditional algorithms.

\section{Details of \textsf{DeepOPF+} Design}\label{appen:sec:details.deepopf+}
We present the detailed result of each step in \textsf{DeepOPF+} design in this appendix. 

First, for determining the maximum calibration rate, the obtained result in shown in Table~\ref{tab:violaton.calibration}, representing the room for DNN prediction error. We note that the off-the-shell solver returns exact solutions for the problem in (\ref{equ:calibration.rate-1})-(\ref{equ:calibration.rate-2}).
\begin{table}[!t]
	\centering
	\fontsize{8}{9}\selectfont
	\caption{Parameters for test cases.} 
	\renewcommand{\arraystretch}{1}
	\begin{threeparttable}
		\begin{tabular}{c|c|c|c|c}
			\toprule
			\hline
			Case & \tabincell{c}{Number of \\buses} & \tabincell{c}{Number of \\generators} & \tabincell{c}{Number of \\load buses} &\tabincell{c}{Number of \\branches}  
			\\
			\hline
			\tabincell{c}{Case30} & 30 & 6 & 20 &41  \\
			\hline
			\tabincell{c}{Case118} & 118 & 19 & 99 &186\\
			\hline
			\tabincell{c}{Case300} & 300 & 69 & 199 &411\\
			\hline
			\bottomrule
		\end{tabular}
		\begin{tablenotes}
			\footnotesize
			\item[*] The number of load buses is calculated based on the default load on each bus. A bus is considered a load bus if its default active power consumption is non-zero.
		\end{tablenotes}
	\end{threeparttable}
	\label{tab.testcase}
\end{table}
\begin{table}[!t]
	\centering
	\caption{Parameters settings of \textsf{DeepOPF+} for IEEE Case30/118/300}
	\renewcommand{\arraystretch}{1.2}
	\begin{threeparttable}	
	\begin{tabular}{c|c|c|c}
	\hline
	\tabincell{c}{Test case}&				
	\tabincell{c}{Variants}&
    	\tabincell{c}{Calibration \\rate}&
    \tabincell{c}{Neurons
per \\hidden layer} \\
	\hline           
	\multirow{2}{*}{\tabincell{c}{Case30}} &\tabincell{c}{\textsf{DeepOPF+-3}} &3.0\%&60/30/15\\
    \cline{2-4}
    & \tabincell{c}{\textsf{DeepOPF+-7}} &7.0\%&32/16/8\\
    \hline
    \multirow{2}{*}{\tabincell{c}{Case118}} &\tabincell{c}{\textsf{DeepOPF+-3}} &3.0\%&200/100/50\\
    \cline{2-4}
    & \tabincell{c}{\textsf{DeepOPF+-7}} &7.0\%&128/64/32\\
    \hline
    \multirow{2}{*}{\tabincell{c}{Case300}} &\tabincell{c}{\textsf{DeepOPF+-3}} &3.0\%&360/180/90\\
    \cline{2-4}
    & \tabincell{c}{\textsf{DeepOPF+-7}} &7.0\%&256/128/64\\
    \hline
	\end{tabular}
	\end{threeparttable}	
	\label{tab:deepopf+setting}
\end{table}

\rev{
\begin{table}[!t]
	\centering
	\caption{Preprocessing time to setup \textsf{DeepOPF+} for IEEE Case30/118/300 in heavy-load regime}
	\renewcommand{\arraystretch}{1.2}
	\begin{threeparttable}	
	\begin{tabular}{c|c|c|c|c|c}
	\hline
	\tabincell{c}{Test case}&				
	\tabincell{c}{Variants}&
    	\tabincell{c}{Determine \\Calibration rate}&
    \tabincell{c}{Determine \\DNN size}&\tabincell{c}{ASA algorithm}&\tabincell{c}{Total time} \\
	\hline           
	\multirow{2}{*}{\tabincell{c}{Case30}} &\tabincell{c}{\textsf{DeepOPF+-3}} &0.2 seconds&0.15 hours&0.83 hour&0.98 hour \\
    \cline{2-6}
    & \tabincell{c}{\textsf{DeepOPF+-7}} &0.2 seconds&0.15 hours&0.73 hour&0.88 hour\\
    \hline
    \multirow{2}{*}{\tabincell{c}{Case118}} &\tabincell{c}{\textsf{DeepOPF+-3}} &20.9 seconds&5.47 hours&7.94 hour&13.42 hour\\
    \cline{2-6}
    & \tabincell{c}{\textsf{DeepOPF+-7}} &20.9 seconds&5.47 hours&5.31 hour&10.79 hour\\
    \hline
    \multirow{2}{*}{\tabincell{c}{Case300}} &\tabincell{c}{\textsf{DeepOPF+-3}} &1185.7 seconds&178.46 hours&25.72 hour&204.51 hour\\
    \cline{2-6}
    & \tabincell{c}{\textsf{DeepOPF+-7}} &1185.7 seconds&178.46 hours&10.52 hour&189.31 hour\\
    \hline
	\end{tabular}
	\end{threeparttable}	
	\label{tab:deepopf+time.complexity.heavy}
\end{table}

\rev{
\begin{table}[!t]
	\centering
	\caption{Preprocessing time to setup \textsf{DeepOPF+} for IEEE Case30/118/300 in light-load regime}
	\renewcommand{\arraystretch}{1.2}
	\begin{threeparttable}	
	\begin{tabular}{c|c|c|c|c|c}
	\hline
	\tabincell{c}{Test case}&				
	\tabincell{c}{Variants}&
    	\tabincell{c}{Determine \\Calibration rate}&
    \tabincell{c}{Determine \\DNN size}&\tabincell{c}{ASA algorithm}&\tabincell{c}{Total time} \\
	\hline           
	\multirow{2}{*}{\tabincell{c}{Case30}} &\tabincell{c}{\textsf{DeepOPF+-3}} &0.2 seconds&0.15 hours&0.81 hour&0.96 hour\\
    \cline{2-6}
    & \tabincell{c}{\textsf{DeepOPF+-7}} &0.2 seconds&0.15 hours&0.72 hour&0.87 hour\\
    \hline
    \multirow{2}{*}{\tabincell{c}{Case118}} &\tabincell{c}{\textsf{DeepOPF+-3}} &20.9 seconds&5.47 hours&6.99 hours&12.47 hours\\
    \cline{2-6}
    & \tabincell{c}{\textsf{DeepOPF+-7}} &20.9 seconds&5.47 hours&4.79 hours&10.27 hours\\
    \hline
    \multirow{2}{*}{\tabincell{c}{Case300}} &\tabincell{c}{\textsf{DeepOPF+-3}} &1185.7 seconds&178.46 hours&52.46 hours&231.25 hours\\
    \cline{2-6}
    & \tabincell{c}{\textsf{DeepOPF+-7}} &1185.7 seconds&178.46 hours&15.82 hours&194.61 hours\\
    \hline
	\end{tabular}
	\end{threeparttable}	
	\label{tab:deepopf+time.complexity.light}
\end{table}
}

\begin{table*}[!ht]
	\centering
	\caption{Average cost and runtime of SOTA DNN schemes in heavy-load regime.}
      \renewcommand{\arraystretch}{0.90}
	\begin{threeparttable}
			\begin{tabular}{c|c|c|c|c|c|}
			\toprule
			\hline
			\multirow{2}{*}{Case}&
			\multirow{2}{*}{\tabincell{c}{Scheme}} &
			\multicolumn{2}{c|}{\tabincell{c}{Average \\cost (\$/hr)}} &
			\multicolumn{2}{c|}{\tabincell{c}{Average running \\time (ms)}}  \\
			\cline{3-6}
			&&\tabincell{c}{DNN scheme}&Ref.&\textsf{DNN scheme}&Ref.\\ 
			\hline
 			\multirow{6}{*}{Case30}&
 			\tabincell{c}{DNN-P} & 732.5 &\multicolumn{1}{|c|}{\multirow{6}{*}{732.2}} & 0.58  & \multicolumn{1}{|c|}{\multirow{6}{*}{45.6}}\\
  			 &\tabincell{c}{DNN-D}  & 732.4 &   &0.63 & \\

  			 &\tabincell{c}{DNN-W}  & 732.2 &   &53.02 & \\
  			 &\tabincell{c}{DNN-G}  & 732.5 &   &1.78 & \\
   			 &\tabincell{c}{DeepOPF+-3}  & 732.4 &   &0.50 & \\
  			 &\tabincell{c}{DeepOPF+-7}  & 732.9 &   &0.49 & \\
 			\hline
 			\multirow{6}{*}{Case118}&
 			\tabincell{c}{DNN-P} & 121074.7 &\multicolumn{1}{|c|}{\multirow{6}{*}{120822.1}}  &2.13 & \multicolumn{1}{|c|}{\multirow{6}{*}{124.9}}\\
  			 &\tabincell{c}{DNN-D}  & 121112.1 &   &15.60 & \\

  			 &\tabincell{c}{DNN-W}  & 120822.1 &   &55.33 & \\
  			 &\tabincell{c}{DNN-G}  & 121299.6 &   &7.72 & \\
   			 &\tabincell{c}{DeepOPF+-3}  & 121051.3 &   &0.56 & \\
  			 &\tabincell{c}{DeepOPF+-7}  & 121313.9 &   &0.55 & \\
 			\hline
 			\multirow{6}{*}{Case300}&
 			\tabincell{c}{DNN-P} & 926660.6 &\multicolumn{1}{|c|}{\multirow{6}{*}{925955.0}}  &3.33 & \multicolumn{1}{|c|}{\multirow{6}{*}{83.5}}\\
  			 &\tabincell{c}{DNN-D}  & 926590.1 &   &57.92&  \\
  			 &\tabincell{c}{DNN-W}  & 925955.0 &   &77.48&  \\
  			 &\tabincell{c}{DNN-G}  & 926512.3 &   &31.55&  \\
   			 &\tabincell{c}{DeepOPF+-3}  & 926198.4 &   &0.61&  \\
  			 &\tabincell{c}{DeepOPF+-7}  & 926500.4 &   &0.60&  \\
 			\hline
			\bottomrule
		\end{tabular}
	\end{threeparttable}
	\label{table.detailed.cost.time.heavy}
\end{table*}

\begin{table*}[!ht]
	\centering
	\caption{Average cost and runtime of SOTA DNN schemes in light-load regime.}
     \renewcommand{\arraystretch}{0.90}
	\begin{threeparttable}
			\begin{tabular}{c|c|c|c|c|c|}
			\toprule
			\hline
			\multirow{2}{*}{Case}&
			\multirow{2}{*}{\tabincell{c}{Scheme}} &
			\multicolumn{2}{c|}{\tabincell{c}{Average \\cost (\$/hr)}} &
			\multicolumn{2}{c|}{\tabincell{c}{Average running \\time (ms)}}  \\
			\cline{3-6}
			&&\tabincell{c}{DNN scheme}&Ref.&\textsf{DNN scheme}&Ref.\\ 
			\hline
 			\multirow{6}{*}{Case30}&
 			\tabincell{c}{DNN-P} & 619.8 &\multicolumn{1}{|c|}{\multirow{6}{*}{619.7}} & 0.50  & \multicolumn{1}{|c|}{\multirow{6}{*}{42.4}}\\
  			 &\tabincell{c}{DNN-D}  & 619.8 &   &0.50 & \\

  			 &\tabincell{c}{DNN-W}  & 619.7 &   &46.93 & \\
  			 &\tabincell{c}{DNN-G}  & 620.4 &   &1.75 & \\
   			 &\tabincell{c}{DeepOPF+-3}  & 619.9 &   &0.50 & \\
  			 &\tabincell{c}{DeepOPF+-7}  & 619.8 &   &0.49 & \\
 			\hline
 			\multirow{6}{*}{Case118}&
 			\tabincell{c}{DNN-P} & 101843.2 &\multicolumn{1}{|c|}{\multirow{6}{*}{101673.0}}  &1.71 & \multicolumn{1}{|c|}{\multirow{6}{*}{115.4}}\\
  			 &\tabincell{c}{DNN-D}  & 101873.6 &   &5.02 & \\

  			 &\tabincell{c}{DNN-W}  & 101673.0 &   &55.55 & \\
  			 &\tabincell{c}{DNN-G}  & 102983.3 &   &4.37 & \\
   			 &\tabincell{c}{DeepOPF+-3}  & 101852.3 &   &0.58 & \\
  			 &\tabincell{c}{DeepOPF+-7}  & 102049.3 &   &0.57 & \\
 			\hline
 			\multirow{6}{*}{Case300}&
 			\tabincell{c}{DNN-P} & 778342.4 &\multicolumn{1}{|c|}{\multirow{6}{*}{777878.4}}  &1.71 & \multicolumn{1}{|c|}{\multirow{6}{*}{78.7}}\\
  			 &\tabincell{c}{DNN-D}  & 778404.3 &   &25.93&  \\
  			 &\tabincell{c}{DNN-W}  & 777878.4 &   &75.77&  \\
  			 &\tabincell{c}{DNN-G}  & 780368.9 &   &32.30&  \\
   			 &\tabincell{c}{DeepOPF+-3}  & 778070.6 &   &0.60&  \\
  			 &\tabincell{c}{DeepOPF+-7}  & 778675.2 &   &0.60&  \\
 			\hline
			\bottomrule
		\end{tabular}
 	\end{threeparttable}
	\label{table.detailed.cost.time.light}
\end{table*}
}


\begin{figure*} [!t]
  \centering
  \subfigure[Case118.]{
    \includegraphics[width = 0.3\textwidth]{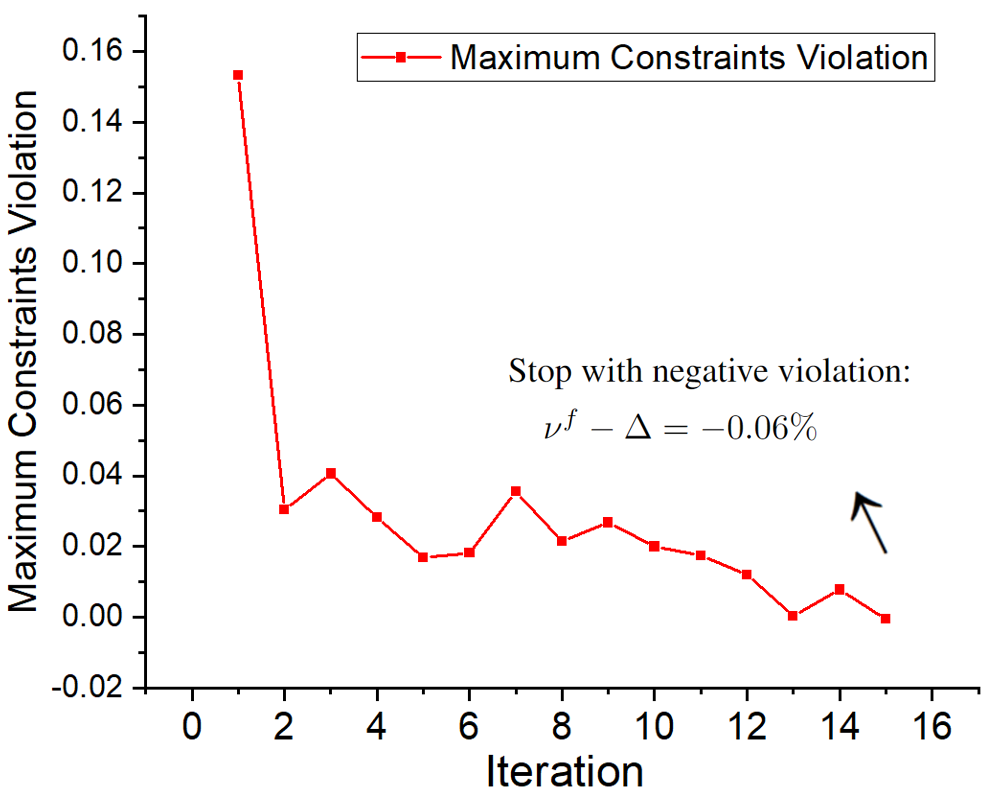}
  }
  \subfigure[Case300.]{
    \includegraphics[width = 0.3\textwidth]{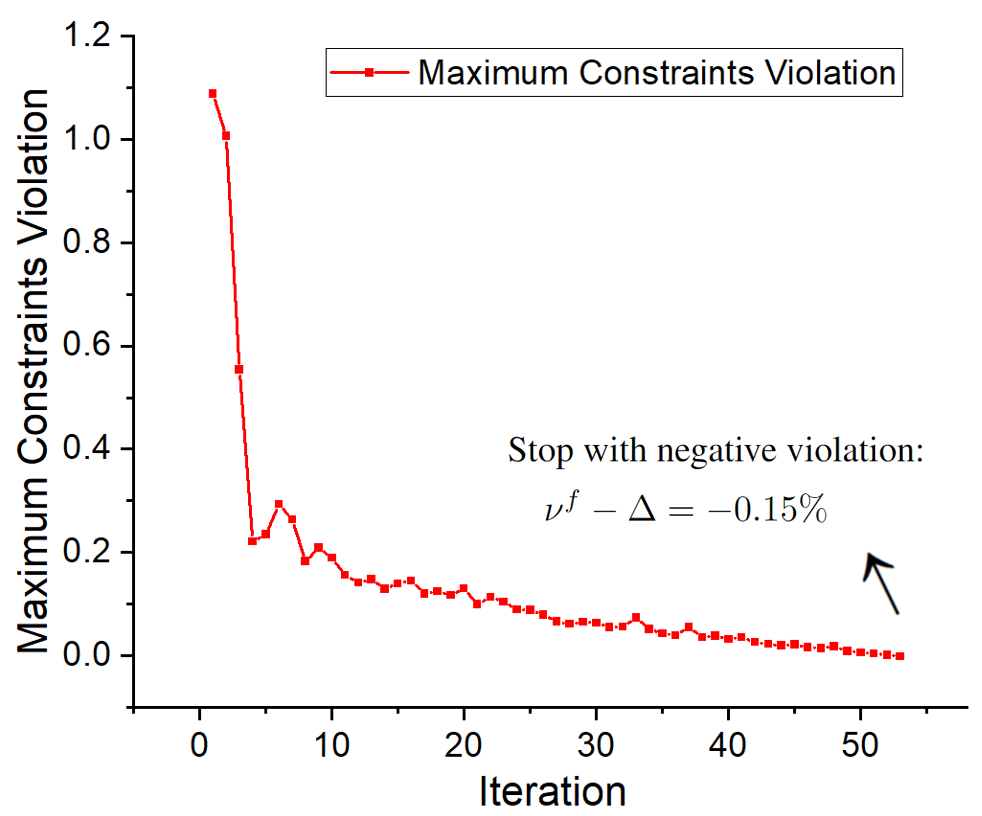}
  }\vspace{-0.1in}
  \caption{Worst-case violation of \textit{Adversarial-Sample Aware} algorithm at each iteration for IEEE Case118 and IEEE Case300 in light-load regime with 7\% calibration rate.}\label{fig.asaalight}\vspace{-0.1in}
\end{figure*}

\begin{figure*} [!t]
  \centering
  \subfigure[Case118.]{
    \includegraphics[width = 0.3\textwidth]{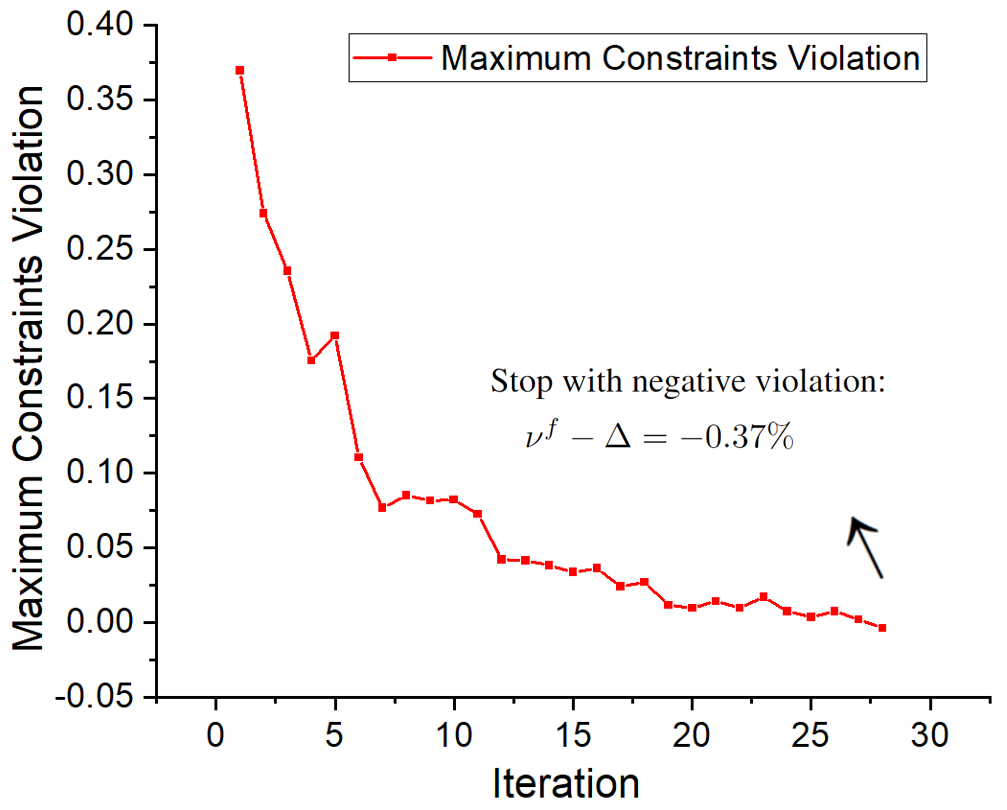}
  }
  \subfigure[Case300.]{
    \includegraphics[width = 0.3\textwidth]{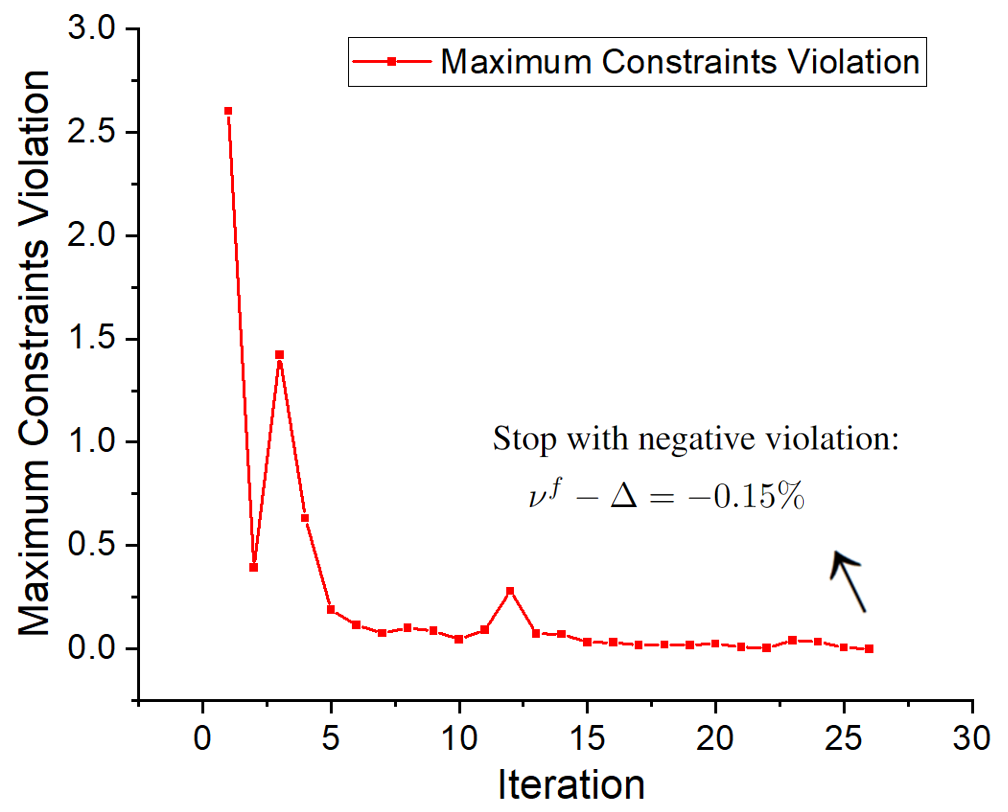}
  }\vspace{-0.1in}
  \caption{Worst-case violation of \textit{Adversarial-Sample Aware} algorithm at each iteration for IEEE Case118 and IEEE Case300 in heavy-load regime with 7\% calibration rate.}\label{fig.asaaheavy}\vspace{-0.1in}
\end{figure*}
Second, for determining the sufficient DNN size, we show the change of the difference between maximum relative constraints violation and calibration rate during iterative solving process via the \textit{Danskin's Theorem} in Fig.~\ref{fig.danskin}. From Fig.~\ref{fig.danskin}, we observe that for all three test cases, the proposed approach succeeds in reaching a {relative constraints} violation no larger than the corresponding calibration rate $\Delta$, i.e., $\rho\leq\Delta$, indicating that the verified DNNs, i.e., 32/16/8 neurons, 128/64/32 neurons
and 256/128/64 neurons, for IEEE 30-/118/300-bus test cases respective, have enough size to guarantee feasibility within the given load input domain of $[100\%, 130\%]$ of the default load. Note that we can directly construct DNNs to ensure universal feasibility for the three IEEE test cases. We further evaluate the {performance of the DNN model obtained following the steps in Sec.~\ref{ssec.suff.dnn.size} without using the \textit{Adversarial-Sample Aware} algorithm.} While ensuring universal feasibility, it suffers from an undesirable optimality loss, up to 2.31\% and more than 130\% prediction error.
\begin{figure*} [!t]
  \centering
  \subfigure[Case118.]{
    \includegraphics[width = 0.3\textwidth]{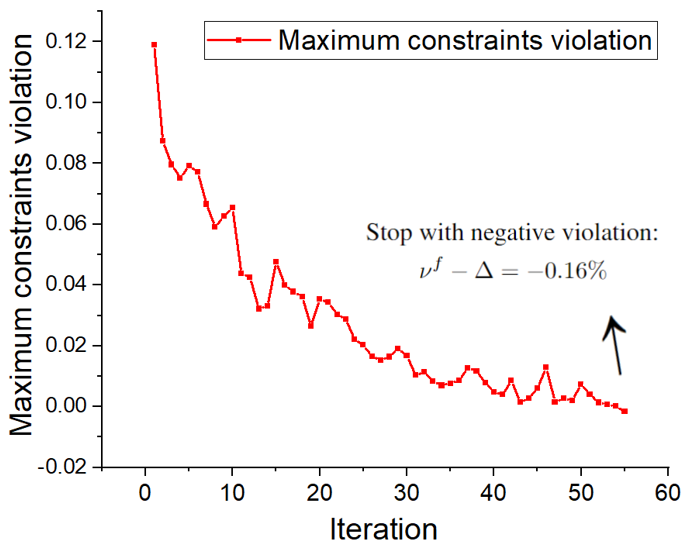}
  }
  \subfigure[Case300.]{
    \includegraphics[width = 0.3\textwidth]{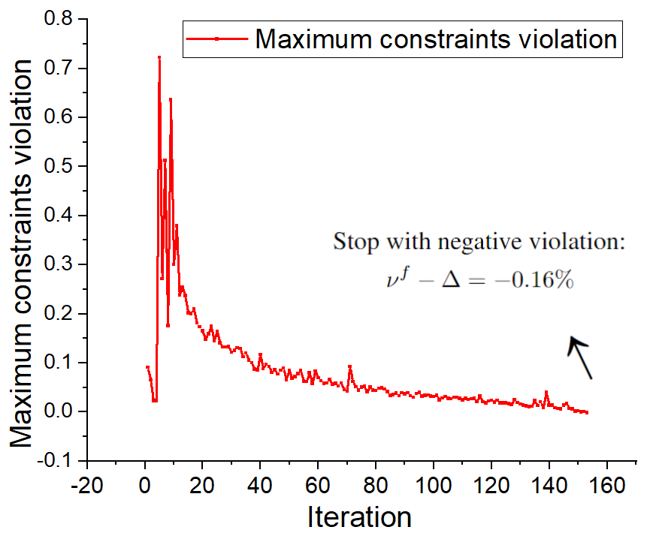}
  }\vspace{-0.1in}
  \caption{Worst-case violation of \textit{Adversarial-Sample Aware} algorithm at each iteration for IEEE Case30/118/300 in light-load regime with 3\% calibration rate.}\label{fig.asa-3-light}\vspace{-0.1in}
\end{figure*}

\begin{figure*} [!t]
  \centering
  \subfigure[Case30.]{
    \includegraphics[width = 0.28\textwidth]{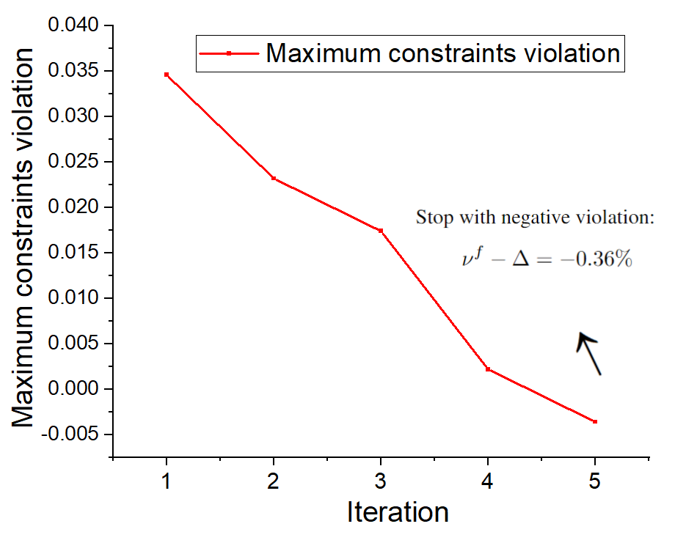}
  }
  \subfigure[Case118.]{
    \includegraphics[width = 0.28\textwidth]{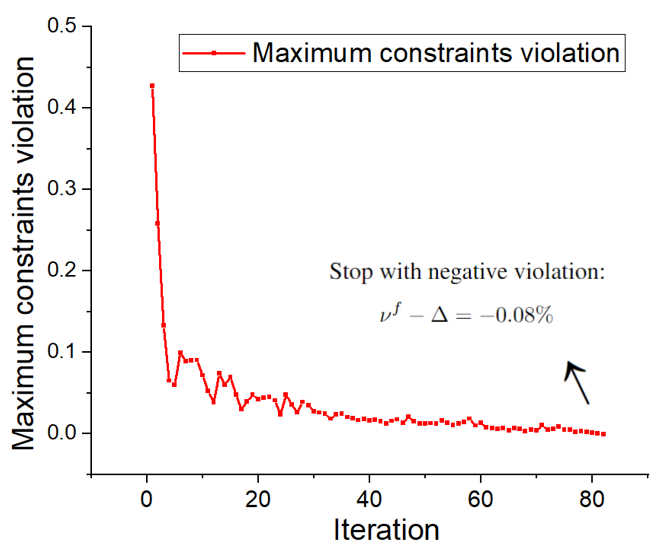}
  }
  \subfigure[Case300.]{
    \includegraphics[width = 0.28\textwidth]{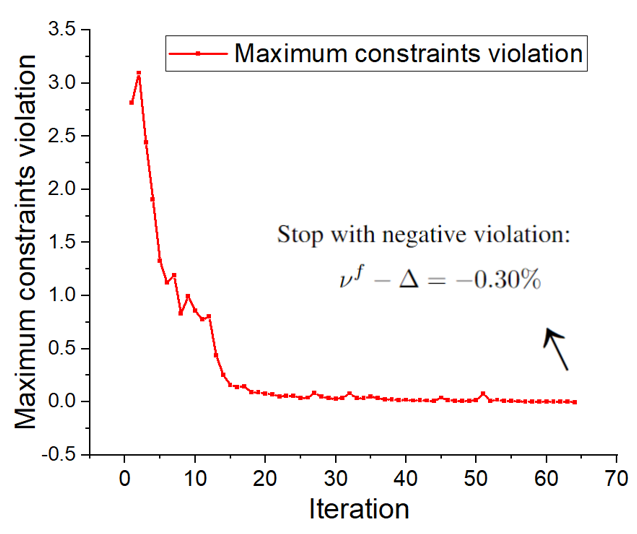}
  }\vspace{-0.1in}
  \caption{Worst-case violation of \textit{Adversarial-Sample Aware} algorithm at each iteration for IEEE Case30/118/300 in heavy-load regime with 3\% calibration rate.}\label{fig.asa-3-heavy}\vspace{-0.1in}
\end{figure*}

Third, the DNN models trained with the \textit{Adversarial-Sample Aware} algorithm achieve lower optimality loss (up to 0.19\%) while preserving universal feasibility. The observation justifies the effectiveness of \textit{Adversarial-Sample Aware} algorithm. We further present the relative violation ($\nu^f-\Delta$) on IEEE 30-/118/300-bus test cases at each iteration in both light-load and heavy-load regimes for illustration in Fig.~\ref{fig.asaalight} and Fig.~\ref{fig.asaaheavy} with a 7\% calibration rate. The above observations show that the \textit{Adversarial-Sample Aware} can efficiently achieve universal feasibility guarantee within both light-load and heavy-load regimes for IEEE 118-/300-bus test cases with at most 52 iterations. We remark that for Case30, the initial worst-case violation of the trained DNN with 7\% calibration rate is less than zero (-9.28\% and -2.93\% in light-load and heavy-load regimes respectively) and hence without the need for adversarial training. {The results under the 3\% calibration rate are presented in Fig.~\ref{fig.asa-3-light} and Fig.~\ref{fig.asa-3-heavy}, for which we observe that the ASA would take a longer number of iterations to achieve the universal feasibility guarantee due to the smaller room for prediction errors, e.g., at most 152 iterations. For Case30 under light-load regime with 3\% calibration rate, its initial worst-case violation is less than zero (-7.53\%) and hence without the need of ASA iterations.}


\rev{Furthermore, we present the parameters of three IEEE test cases and the settings of two \textsf{DeepOPF+} schemes in Table~\ref{tab.testcase} and Table~\ref{tab:deepopf+setting} respectively. The detailed runtime and cost and the time to configure the framework are listed in Table~\ref{tab:deepopf+time.complexity.light}-Table~\ref{table.detailed.cost.time.light} for each test case. Note that though a single DC-OPF may be efficient solved by the existing solver, due to increasing
uncertainty from renewable generation and flexible load, grid operators now need to solve DC-OPF problems under many scenarios in a short interval, e.g., 1000 scenarios in 1 minutes, to obtain a stochastically optimized solution, e.g., $\sim$2 minutes for the iterative solvers to solve a large number
of DC-OPF problems for Case118, resulting the fail of real-time operation. In contrast, the developed DNN scheme can return the solution with $\times$228 speedups, i.e., less than 0.6 seconds in total.  In addition, though our method takes additional training efforts, 1) it is conducted offline, once the DNN is configured, it can be continuously applied to many test instances such that the complexity is amortized, e.g., $<0.5$ ms for DC-OPF problems if the system operator needs to solve DC-OPF per 5 minutes over 1000 scenarios over a year; 2) as illustrated, the obtained DNN outperforms the existing approaching in avoiding any post-processing and resulting in a lower real-time runtime complexity, showing its advantage; 3) our theoretical analysis shows that the design can always provide the corresponding useful upper/lower bounds in each step of the framework in polynomial time, which can still be utilized for constraints calibration and DNN performance analysis; 4) the process can be further accelerated by applying advanced computation parallel techniques. Finally, we remark that if an impractically large DNN size is required, it would introduce an additional computational challenge, which can require more configuration efforts of the approach and it can be a potential limitation. It is also an interesting direction 
for solving the constrained program w.r.t. the DNN parameters and determining the sufficient DNN size more efficiently. We would like to leave how to set up the DNNs more efficiently and accelerate the corresponding steps as future work, which is non-trivial and still an open problem in DNN scheme design.}





\rev{
\section{Non-convex optimization example}\label{appen.sec.nonconvex}
We further consider solving a non-convex linearly constrained program with a non-convex objective function and linear constraints adapted from~\cite{donti2021dc3}. We examine this task for illustration:
\begin{align}\label{appen.equ.nonconvex.obj}
    \min_{y\in\mathcal{R}^n} \quad \frac{1}{2}y^TQy+p^T\sin(y), \text{s.t.} \quad Ay=x, -h\leq Gy\leq h,
\end{align}
for constants problem parameter $Q\in\mathcal{R}^{n\times n}, p\in\mathcal{R}^n, A\in\mathcal{R}^{n_{\text{eq}}\times n}, G\in\mathcal{R}^{n_{\text{ineq}}\times n}, h\in\mathcal{R}^{n_{\text{ineq}}}$. Here $x\in\mathcal{R}^{n_{\text{eq}}}$ is the problem input and $y\in\mathcal{R}^n$ denotes the decision variable. $n_{\text{ineq}}$ and $n_{\text{eq}}$ are the number of inequality and equality constraints. Here we focus on the non-degenerate case such that $n_{\text{eq}}\leq n$. Therefore, the DNN task aims to learn the mapping between $x$ to the optimal $y$. Similar to~\cite{donti2021dc3}, $Q$ is set to be a diagonal matrix whose diagonal entries are drawn i.i.d. from
the uniform distribution on $[0,1]$. The entries of $A,G$ are drawn i.i.d. from the unit normal distribution. The problem input region of $x$ is set to be $[-1,1]$ for each dimension. To ensure the problem feasibility, we set $h_i=\sum_j|(GA^+)_{ij}|$, where $A^+$ is the Moore-Penrose pseudoinverse of $A$. The feasibility of the problem can be seen that the point $y=A^+x$ is feasible. However, such a point can be generally non-optimal with large optimality loss. In our simulation, we set $n=50$, $n_{\text{eq}}=25$, and $n_{\text{ineq}}=25$. Therefore, the considered optimization has $50$ variables, $25$ equality constraints, and $100$ inequality constraints. 

We follow the procedures in the \textit{preventive learning} framework to generate the DNN with universal feasibility guarantee and achieve strong optimality performance.

\subsection{Reformatting the problem with only inequality constraints}
We reformulate the non-convex optimization with only $n-n_{\text{eq}}$ independent variables of $y_2$. Note that the equality constraints can be reformulated as
\begin{align}
    [A_1,A_2]\left[\begin{array}{cccc} 
    y_1  \\ 
    y_2  
\end{array}\right]=x
\end{align}
Here $A_1\in\mathcal{R}^{n_{\text{eq}}\times n_{\text{eq}}}, y_1\in\mathcal{R}^{n_{\text{eq}}}$ and $A_2\in\mathcal{R}^{n_{\text{eq}}\times (n-n_{\text{eq}})}, y_2\in\mathcal{R}^{n-n_{\text{eq}}}$. Therefore, given $x$ and $y_2$, the corresponding $y_1$ can be uniquely recovered, i.e., $y_1=A^{-1}_1(x-A_2y_2)$. Based on the above reformulation, the inequality constraints are given as
\begin{align}
    [G_1,G_2]\left[\begin{array}{cccc} 
    y_1  \\ 
    y_2  
\end{array}\right]\leq h,  \quad   -[G_1,G_2]\left[\begin{array}{cccc} 
    y_1  \\ 
    y_2  
\end{array}\right]\leq h
\end{align}
and hence
\begin{align}\label{appen:equ.reformualted.inequality}
    G_1A^{-1}_1x+(G_2-G_1A^{-1}_1A_2)y_2\leq h, \quad G_1A^{-1}_1x+(G_2-G_1A^{-1}_1A_2)y_2\geq -h
\end{align}
The objective can be equivalent modified by replacing the terms w.r.t. $y_1$ to be $y_2$ from $y_1=A^{-1}_1(x-A_2y_2)$. This completes the pre-reformulation of the above non-convex optimization.

\subsection{Determine the maximum allowable calibration rate}
We first examine that all inequality constraints are critical, i.e., exist a $y$ such that the constraint is binding. We then further determine the maximum calibration rate. From the description in Sec.~\ref{ssec:calibrationrange}, the program to determine the maximum calibration rate is given as
\begin{align}
    \min_{x\in[-1,1]}  &\max_{y,\nu^{c}} \;\;\;   \nu^{c} \\
    \mathrm{s.t.}\; &(\ref{appen:equ.reformualted.inequality}) \nonumber \\ 
    & \nu^c\leq (h_i-(G_1A^{-1}_1x+(G_2-G_1A^{-1}_1A_2)y_2)_i)/h_i,\;  i=1,...,n_{\text{ineq}},\\
    & \nu^c\leq (h_i+(G_1A^{-1}_1x+(G_2-G_1A^{-1}_1A_2)y_2)_i)/h_i,\;  i=1,...,n_{\text{ineq}}.
\end{align}
Note that given $x$, the inner problem is an LP and can be equivalently expressed by its sufficient and necessary KKT conditions. Following the MILP steps in Sec.~\ref{ssec:calibrationrange}, we solve the above program  to determine the maximum allowable calibration rate, we observe that the Gurobi solver with the branch-and-bound provides its optimal solution with zero optimality gap within $42$ms. The corresponding optimal $\nu^{c*}=100\%$, implying we can set $h=0$ such that problem is still feasible for each problem input $x\in[-1,1]$.

\subsection{Determine the sufficient DNN size to guarantee universal feasibility}

In our simulation, we consider a DNN with 3 hidden layers and each layer has $50$ neurons. Following the steps in Sec.~\ref{ssec:feasibilityDNN}, we observe that such a DNN size is sufficient to guarantee universal feasibility. The corresponding program is given as\vspace{-0.1in}
\begin{align}
    \min_{\mathbf{W}, \bold{b}} \max_{x\in[-1,1]} \;\; & {\nu}^f \\
    \mathrm{s.t.} \;\; & (\ref{equ.DNNinteger-1})-(\ref{equ.DNNinteger-2}),  1\leq i\leq N_{{\text{hid}}}, 1\leq k\leq N_{\text{neu}}, \nonumber\\
    &{\nu}^f=\max_{i=1,...,n_{\text{ineq}}}\left\{\begin{array}{cccc} 
    (G_1A^{-1}_1x+(G_2-G_1A^{-1}_1A_2)\hat{y}_2)_i/ h_i  \\ 
    -(G_1A^{-1}_1x+(G_2-G_1A^{-1}_1A_2)\hat{y}_2)_i/ h_i  
\end{array}\right\}.
\end{align}
Here $\hat{y}_2$ is the prediction of the DNN. We observe that the tested DNN size is sufficient to guarantee universal feasibility by achieve an upper bound of the relative violation of $\rho-\nu^f$ as $-9.3\%$ within $\sim6$ minutes. It implies that the tested DNN size is sufficient to guarantee universal feasibility. Recall that the obtained \textsf{DNN-FG} achieves unsatisfactory optimality performance (71.38\% optimality loss) as it only focuses on feasibility.

\subsection{Application of Adversarial-Sample Aware training algorithm}

We hence implement the proposed \textsf{ASA} training algorithm to further improve the optimality performance of the DNN with 5\% and 10\% calibration rates respectively. The time to obtain the corresponding (\textsf{Pre-DNN-5}, \textsf{Pre-DNN-10}) with 
5\% and 10\% calibration rate are $<52$ minutes and $<44$ minutes respectively.

We compare our approach against the classical non-convex optimization solver IPOPT and the other DNN schemes \textsf{DNN-P}, \textsf{DNN-D},\textsf{DNN-W}, and \textsf{DNN-G}. The number of training data is 15,000, and the number of test data is 3,000. The DNN size is set as 3 hidden layers and each layer has $50$ neurons. The results are listed in Table~\ref{table.non-convex}, and the worst-case violation at each iteration in the ASA training algorithm are given in Fig.~\ref{fig.non-convex}. Here the optimality \textit{Loss} metric is calculated as the average of $(\text{DNN objective}-\text{Optimal objective})/|\text{Optimal objective}|$. The negativity of \textit{Scheme} and \textit{Ref} simply means that the obtained DNN objective and Optimal objective of optimization (\ref{appen.equ.nonconvex.obj}) is negative.
\vspace{-0.1in}

\begin{table*}[!ht]
	\centering
	\caption{Simulation results of different DNN schemes for the non-convex optimization example.}
	\resizebox{\textwidth}{18mm}{
	\begin{threeparttable}
			\begin{tabular}{c|c|c|c|c|c|c|c|c}
			\toprule
			\hline
			\multirow{2}{*}{\tabincell{c}{Scheme}} &
			\multicolumn{3}{c|}{\tabincell{c}{Average objective}} &
			\multicolumn{3}{c|}{\tabincell{c}{Average running time (ms)}}& \multirow{2}{*}{\tabincell{c}{Feasibility\\ rate (\%)}} & \multirow{2}{*}{\tabincell{c}{Worst-case\\ violation (\%)}}\\
			\cline{2-7}
			&\tabincell{c}{ Scheme}&Ref.&\tabincell{c}{Loss (\%)}&\tabincell{c}{ Scheme}&Ref.&\tabincell{c}{Speedup}&&\\ 
			\hline
 			
 			\tabincell{c}{DNN-P} & -5.44 &\multicolumn{1}{|c|}{\multirow{6}{*}{-5.47}} & 0.40  &1.36& \multicolumn{1}{|c|}{\multirow{6}{*}{86.6}}&85.7&39.8&68.3\\
  			 \tabincell{c}{DNN-D}  & -5.44 &  &0.42 &0.79 & &117.0&39.8&41.5\\

  			 \tabincell{c}{DNN-W}  & -5.47 &  &0 &86.6 & &1.02&100&0\\
  			 \tabincell{c}{DNN-G}  & 53.69 &   &1076.0&1.00 & &87.0&100&0\\
   			 \tabincell{c}{Pre-DNN-5}  & -5.45 & &0.34  &0.60 & &144.9&100&0\\
  			 \tabincell{c}{Pre-DNN-10}  & -5.43 & &0.67&0.60 & &145.3&100&0\\
 			\hline
			\bottomrule
		\end{tabular}
		\begin{tablenotes}
			\scriptsize
			\item[*] Feasibility rate and Worst-case violation are the results \textit{before} post-processing. {Feasibility rates (resp Worst-case violation) after post-processing \\are 100\% (resp 0) for all DNN schemes. We hence report the results before post-processing to better show the advantage of our design. Speedup\\ and Optimality loss are the results \textit{after} post-processing of the final obtained feasible solutions.}
			\item[*] The \textit{correction} step in \textsf{DNN-D} (with $10^{-4}$ rate) is faster compared with $l_1$-projection in \textsf{DNN-P}, resulting in higher speedups.
		\end{tablenotes}
	\end{threeparttable}}
	\label{table.non-convex}\vspace{-0.05in}
\end{table*}

\begin{figure*} [!t]
  \centering
  \subfigure[$5\%$ calibration rate.]{
    \includegraphics[width = 0.32\textwidth]{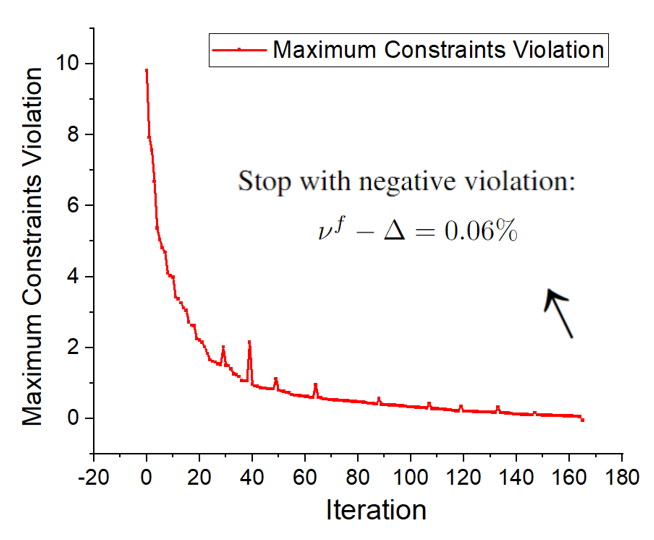}
  }
  \subfigure[$10\%$ calibration rate.]{
    \includegraphics[width = 0.3\textwidth]{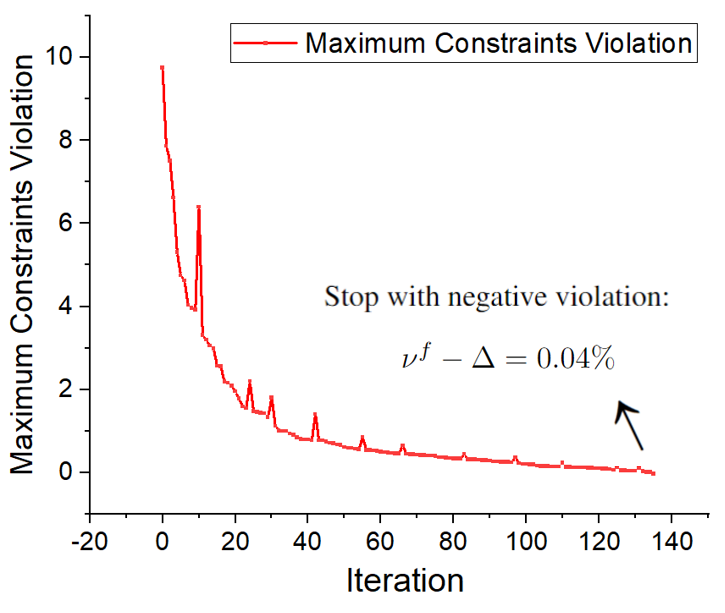}
  }\vspace{-0.1in}
  \caption{Worst-case violation of \textit{Adversarial-Sample Aware} algorithm at each iteration for the non-convex optimization example with 5\% and 10\% calibration rate.}\label{fig.non-convex}\vspace{-0.1in}
\end{figure*}

We remark that our obtained DNN schemes (\textsf{Pre-DNN-5}, \textsf{Pre-DNN-10}) with 
5\% and 10\% calibration rates outperform the existing DNN scheme in ensuring universal feasibility and maintaining minor optimality loss. The speedups of our scheme are also significantly larger than the other methods as post-processing steps to recover solution feasibility are avoided.
}
\end{appendices}
\end{document}